\documentclass[10pt]{article} 
\usepackage[accepted]{tmlr}

\usepackage{floatrow}
\usepackage{enumerate}

\usepackage{amsthm}

\usepackage{thmtools}
\usepackage{thm-restate}

\newtheorem{mythm}{Theorem}

\newtheorem{mydef}[mythm]{Definition}
\newtheorem{myRemark}[mythm]{Remark}

\makeatletter
\def\cleartheorem#1{%
    \expandafter\let\csname#1\endcsname\relax
    \expandafter\let\csname c@#1\endcsname\relax
}
\makeatother

\newcommand{\secref}[1]{\text{Sec.}~\ref{#1}}
\newcommand{\appref}[1]{\text{Appendix}~\ref{#1}}
\newcommand{\figref}[1]{\text{Fig.}~\ref{#1}}
\newcommand{\algoref}[1]{\text{Algorithm}~\ref{#1}}

\newcommand{\thmref}[1]{\text{Theorem}~\ref{#1}}

\newcommand{\defref}[1]{\text{Definition}~\ref{#1}}

\newcommand{\remref}[1]{\text{Remark}~\ref{#1}}

\usepackage{dsfont}
\usepackage[utf8]{inputenc} 
\usepackage[german=quotes]{csquotes}
\usepackage[T1]{fontenc}    
\PassOptionsToPackage{hyphens}{url}\usepackage{hyperref}   
\usepackage[shortcuts]{extdash}
\usepackage{booktabs}       
\usepackage{amsfonts}       
\usepackage{amsmath}
\usepackage{amssymb}
\usepackage{bm}
\usepackage{bbm}
\usepackage{nicefrac}       
\usepackage{stmaryrd}
\usepackage{relsize}
\usepackage{scalerel}
\usepackage{microtype}      
\usepackage{breakcites}
\usepackage{xspace}

\newcommand{\myParagraph}[1]{\noindent{\textbf{#1}\;\;}}
\usepackage{color}
\usepackage{algorithm}
\usepackage[noend]{algpseudocode}
\usepackage{dashrule}
\usepackage{multirow}
\newcommand\Algphase[1]{%
\vspace*{-.7\baselineskip}\Statex\hspace*{\dimexpr-\algorithmicindent-2pt\relax}\hdashrule[-0.4ex]{1.\textwidth}{0.4pt}{4pt 3pt}%
\Statex\hspace*{-\algorithmicindent}\textbf{#1}%
\vspace*{-.7\baselineskip}\Statex\hspace*{\dimexpr-\algorithmicindent-2pt\relax}\hdashrule{1.\textwidth}{0.4pt}{4pt 3pt}%
}

\algnewcommand{\LineComment}[1]{\State \(\triangleright\) #1}
\usepackage{graphicx}
\usepackage[percent]{overpic}
\usepackage[section]{placeins}

\newcommand{\fig}[5]{\begin{figure}[tb]
  \centering
  \includegraphics[width=#2\linewidth]{#1}
  \vspace*{-#4}
  \caption{#3}
  \vspace*{-#5}
  \label{fig:#1}
\end{figure}
}

\newcommand{\figSideCaption}[5]{\begin{figure}[tb]
  \floatbox[{\capbeside\thisfloatsetup{capbesideposition={right,center},capbesidewidth=#4}}]{figure}[\FBwidth]
{\caption{#3}\label{fig:#1}}
{\includegraphics[width=#2\linewidth]{#1}}
\vspace*{-#5}
\end{figure}
}

\usepackage{subfigure}

\usepackage{soul}
\DeclareRobustCommand*{\showDifferenceDanny}[2]{\ifthenelse{\equal{#1}{}}{}{\textcolor{blue}{\st{#1}}}\xspace \ifthenelse{\equal{#2}{}}{}{\textcolor{blue}{#2}}\xspace}
\DeclareRobustCommand*{\showDifferenceShin}[2]{\ifthenelse{\equal{#1}{}}{}{\textcolor{magenta}{\st{#1}}} \ifthenelse{\equal{#2}{}}{}{\textcolor{magenta}{#2}}\xspace}
\DeclareRobustCommand*{\showDifferenceKlaus}[2]{\ifthenelse{\equal{#1}{}}{}{\textcolor{green}{\st{#1}}} \ifthenelse{\equal{#2}{}}{}{\textcolor{green}{#2}}\xspace}
\DeclareRobustCommand*{\showDifferenceStefan}[2]{\ifthenelse{\equal{#1}{}}{}{\textcolor{cyan}{\st{#1}}} \ifthenelse{\equal{#2}{}}{}{\textcolor{cyan}{#2}}\xspace}

\newcommand{\mySum}[2]{\sideset{}{_{#1}^{#2}}\sum}

\DeclareMathOperator*{\bigInt}{\ensuremath{\mathlarger{\int}}\xspace}
\newcommand{\myInt}[1]{\ensuremath{\sideset{}{_{\hspace*{-5pt}#1}}\bigInt}\xspace}
\DeclareMathOperator*{\Prob}{\mathds{P}}
\DeclareMathOperator*{\R}{\mathds{R}}
\newcommand{\reals}[1]{\ensuremath{\R^{#1}_{}}\xspace}
\DeclareMathOperator*{\N}{\mathds{N}}
\DeclareMathOperator*{\Nzero}{\ensuremath{\mathds{N}^{}_{\raisebox{0pt}{\tiny\ensuremath{0}}}}\xspace}
\DeclareMathOperator*{\E}{\mathds{E}}

\DeclareMathOperator*{\Var}{\mathds{V}}
\DeclareMathOperator*{\indicator}{\scalerel*{\mathbbmss{1}}{\textstyle\sum}}
\newcommand{\indicatorFunction}[2]{\ensuremath{\sideset{}{_{#1}^{}}\indicator(#2)}}
\newcommand{\idMatrix}[1]{\ensuremath{\mathcal{I}_{#1}}}
\newcommand{\ones}[1]{\ensuremath{\mathbbmss{1}_{#1}^{}}}

\newcommand{\abs}[1]{\left|#1\right|}
\newcommand{\pnorm}[2]{\|#1\|_{\raisebox{-2pt}{\tiny\ensuremath{#2}}}}

\DeclareMathOperator*{\argmin}{\textbf{arg\hspace{0.1em}min}}
\DeclareMathOperator*{\argmax}{\textbf{arg\hspace{0.1em}max}}
\DeclareMathOperator*{\softmax}{\textbf{soft\hspace{0.1em}max}}
\newcommand{\sparsity}{\ensuremath{\kappa}\xspace}

\newcommand{\gateNoise}{\ensuremath{\mathfrak{s}}\xspace}

\newcommand{\diag}{\textbf{\textit{diag}}\xspace}

\newcommand{\trace}{\textbf{\textit{trace}}\xspace}
\newcommand{\vol}{\textbf{\textit{Vol}}\xspace}

\newcommand{\nonnegativeReal}[1]{\ensuremath{\mathds{R}^{#1}_{\raisebox{0pt}{\tiny\ensuremath{+}}}}\xspace}
\newcommand{\posDefSet}[1]{\ensuremath{\mathbb{S}^{#1}_{\raisebox{0pt}{\tiny\ensuremath{++}}}}\xspace}
\newcommand{\positiveReal}[1]{\ensuremath{\mathds{R}^{#1}_{\raisebox{0pt}{\tiny\ensuremath{++}}}}\xspace}
\newcommand{\inputSpace}{\ensuremath{\mathcal{X}}\xspace}
\newcommand{\outputSpace}{\ensuremath{\mathcal{Y}}\xspace}
\newcommand{\intrinsicDimension}{\ensuremath{\delta}\xspace}
\newcommand{\inputDimension}{\ensuremath{d}\xspace}

\newcommand{\trainingInputs}[1]{\ensuremath{\bm{X}_{#1}^{}}\xspace}
\newcommand{\trainingLabels}[1]{\ensuremath{\bm{Y}_{#1}^{}}\xspace}
\newcommand{\optTrainingInputs}[1]{\ensuremath{\bm{X}_{#1}'}\xspace}
\newcommand{\optTrainingLabels}[1]{\ensuremath{\bm{Y}_{#1}'}\xspace}
\newcommand{\validationInputs}{\ensuremath{\bm{X}_{\text{val}}^{}}\xspace}
\newcommand{\validationLabels}{\ensuremath{\bm{Y}_{\text{val}}^{}}\xspace}
\newcommand{\trueTestInputs}{\ensuremath{\bm{X}_{\text{T}}^{}}\xspace}

\newcommand{\evalInputs}{\ensuremath{\bm{X}_{*}}\xspace}
\newcommand{\evalLabels}{\ensuremath{\bm{Y}_{*}}\xspace}
\newcommand{\ipSymbol}{\ensuremath{\dagger}\xspace}
\newcommand{\ipInputs}{\ensuremath{\bm{X}_\ipSymbol}\xspace}

\newcommand{\poolInputs}{\ensuremath{\bm{X}_{pool}^{}}\xspace}
\newcommand{\poolLabels}{\ensuremath{\bm{Y}_{pool}^{}}\xspace}

\newcommand{\poolDensity}{\ensuremath{p_{\inputSpace}^{}}\xspace}

\newcommand{\approxPoolDensity}[1]{\ensuremath{\widehat{p}_{\inputSpace}^{#1}}\xspace}

\newcommand{\expertInducingPoints}[1]{\ensuremath{\bm{X}^{#1}_\ipSymbol}\xspace}
\newcommand{\expertInducingValues}[1]{\ensuremath{\bm{Y}^{#1}_\ipSymbol}\xspace}
\newcommand{\expertInducingValueMean}[1]{\ensuremath{\bm{\mu}^{#1}_\ipSymbol}\xspace}
\newcommand{\expertInducingValueCov}[1]{\ensuremath{\bm{S}^{#1}_\ipSymbol}\xspace}
\newcommand{\expertInducingValueWhitenedMean}[1]{\ensuremath{\widetilde{\bm{\mu}}^{#1}_\ipSymbol}\xspace}
\newcommand{\expertInducingValueWhitenedCov}[1]{\ensuremath{\widetilde{\bm{S}}^{#1}_\ipSymbol}\xspace}
\newcommand{\gateInducingPoints}{\ensuremath{\bm{X}^G_\ipSymbol}\xspace}
\newcommand{\gateInducingValueMean}{\ensuremath{\bm{\mu}^G_\ipSymbol}\xspace}

\newcommand{\kernelMatrix}[1]{\ensuremath{\bm{K}^{#1}}\xspace}
\newcommand{\GP}[1]{\ensuremath{\mathcal{GP}(#1)}\xspace}
\newcommand{\SGP}[1]{\ensuremath{\mathcal{SGP}(#1)}\xspace}
\newcommand{\SVGP}[1]{\ensuremath{\mathcal{SVGP}(#1)}\xspace}
\newcommand{\fGP}{\ensuremath{\widehat{f}_\text{\tiny{GP}}}\xspace}
\newcommand{\gpNoiseVariance}{\ensuremath{\widehat{v}}\xspace}
\newcommand{\fLPS}[1]{\ensuremath{\widehat{f}_\text{\tiny{LPS}}^{#1}}\xspace}
\newcommand{\uniformDist}[1]{\ensuremath{\mathcal{U}(#1)}\xspace}
\newcommand{\Gauss}[3]{\ensuremath{\mathcal{N}(#1; #2, #3)}\xspace}
\newcommand{\fctn}[3]{\ensuremath{#1\!:#2\,\rightarrow\,#3}\xspace}
\newcommand{\mySet}[2]{\ensuremath{\{#1,\dots,#2\}}\xspace}

\newcommand{\condset}[2]{\ensuremath{\left\{#1\;\middle|\;#2\right\}}\xspace}
\newcommand{\diffableFunctions}[2]{\ensuremath{\mathcal{C}^{#2}\left(#1\right)}\xspace}
\newcommand{\intableFunctions}[2]{\ensuremath{\mathcal{L}^{#2}\left(#1\right)}\xspace}
\newcommand{\relSampleSize}[2]{\raisebox{1pt}{\ensuremath{\varrho}}(#1, #2)\xspace}
\newcommand{\relSampleSizeSymbol}{\raisebox{1pt}{\ensuremath{\varrho}}\xspace}

\newcommand{\kullbackLeiblerDistance}[2]{\ensuremath{\mathcal{KL}\left[#1 \| #2\right]}\xspace}
\newcommand{\LPSorder}{\ensuremath{Q}\xspace}
\newcommand{\predictorLPS}[2]{\ensuremath{m^{#2}_{#1}}\xspace}

\newcommand{\biasLPS}[3]{\ensuremath{\text{b}_{#3}^{}\left[#2,#1\right]}\xspace}

\newcommand{\bandwidth}{\ensuremath{\Sigma}\xspace}
\newcommand{\SigmaSpace}{\ensuremath{\mathcal{S}}\xspace}
\newcommand{\SigmaStep}[1]{\ensuremath{\bm{\mathcal{S}}_{#1}}\xspace}

\newcommand{\convergenceInProbability}[2]{\ifthenelse{\equal{#1}{}}{\ensuremath{o_p^{}\left[#2\right]}}{\ensuremath{o_p^{}\begingroup\edef\x{\endgroup#1[}\x #2 \begingroup\edef\x{\endgroup#1]}\x}}\xspace}
\newcommand{\upperBoundedInProbability}[2]{\ifthenelse{\equal{#1}{}}{\ensuremath{O_p^{}\left[#2\right]}}{\ensuremath{O_p^{}\begingroup\edef\x{\endgroup#1[}\x #2 \begingroup\edef\x{\endgroup#1]}\x}}\xspace}
\newcommand{\exactRateconvergenceInProbability}[2]{\ifthenelse{\equal{#1}{}}{\ensuremath{\theta_p^{}\left[#2\right]}}{\ensuremath{\theta_p^{}\begingroup\edef\x{\endgroup#1[}\x #2 \begingroup\edef\x{\endgroup#1]}\x}}\xspace}
\newcommand{\divergenceInProbability}[2]{\ifthenelse{\equal{#1}{}}{\ensuremath{\omega_p^{}\left[#2\right]}}{\ensuremath{\omega_p^{}\begingroup\edef\x{\endgroup#1[}\x #2 \begingroup\edef\x{\endgroup#1]}\x}}\xspace}
\newcommand{\lowerBoundedInProbability}[2]{\ifthenelse{\equal{#1}{}}{\ensuremath{\Omega_p^{}\left[#2\right]}}{\ensuremath{\Omega_p^{}\begingroup\edef\x{\endgroup#1[}\x #2 \begingroup\edef\x{\endgroup#1]}\x}}\xspace}

\newcommand{\LOBfunctionOfLPS}[2]{\ensuremath{\bandwidth^{#2}_{#1}}\xspace}
\newcommand{\lobfunctionOfLPS}[2]{\ensuremath{\sigma^{#2}_{#1}}\xspace}
\newcommand{\approxLOBfunctionOfLPS}[2]{\ensuremath{\widehat{\bandwidth}^{#2}_{#1}}\xspace}
\newcommand{\approxlobfunctionOfLPS}[2]{\ensuremath{\widehat{\sigma}^{#2}_{#1}}\xspace}

\newcommand{\fMoE}{\ensuremath{\widehat{f}^{}_\text{\tiny{MoE}}}\xspace}
\newcommand{\thetaMoE}{\ensuremath{\Theta}\xspace}

\newcommand{\fctnComplexityLPS}[2]{\ensuremath{\mathfrak{C}^{#2}_{#1}}\xspace}
\newcommand{\approxFctnComplexityLPS}[2]{\ensuremath{\widehat{\mathfrak{C}}^{#2}_{#1}}\xspace}

\newcommand{\pOptLPS}[2]{\ensuremath{p^{#1,#2}_\text{\tiny{Opt}}}\xspace}
\newcommand{\pSup}[2]{\ensuremath{p^{#1,#2}_\text{\tiny{Sup}}}\xspace}

\newcommand{\approxPSup}[2]{\ensuremath{\widehat{p}^{#1,#2}_\text{\tiny{Sup}}}\xspace}

\newcommand{\AL}{\text{AL}\xspace}
\newcommand{\ML}{\text{ML}\xspace}
\newcommand{\SOTA}{\text{state-of-the-art}\xspace}
\newcommand{\LPS}{\text{LPS}\xspace}

\newcommand{\MSE}{\text{MSE}\xspace}
\newcommand{\MLL}{\text{Obj}\xspace}
\newcommand{\RMSE}{\text{RMSE}\xspace}
\newcommand{\maxAE}{\text{max\! AE}\xspace}
\newcommand{\MISE}{\text{MISE}\xspace}

\newcommand{\LOB}{\text{LOB}\xspace}
\newcommand{\LFC}{\text{LFC}\xspace}
\newcommand{\RBF}{\text{RBF}\xspace}
\newcommand{\GPR}{\text{GPR}\xspace}
\newcommand{\GPU}{\text{GP uncertainty sampling}\xspace}
\newcommand{\DGP}{\text{DGP}\xspace}
\newcommand{\MoGPU}{\text{MoGPU}\xspace}
\newcommand{\sGDML}{\text{sGDML}\xspace}
\newcommand{\LOO}{\text{LOO}\xspace}

\newcommand{\MoE}{\text{MoE}\xspace}
\newcommand{\inducingPoint}{IP\xspace}
\newcommand{\inducingPoints}{IPs\xspace}
\newcommand{\GaussianProcess}{GP\xspace}
\newcommand{\GaussianProcesses}{GPs\xspace}
\newcommand{\GFF}{\text{GFF}\xspace}
\newcommand{\kMeans}{\text{k-means}\xspace}
\newcommand{\kMeansPlusPlus}{\text{k-means\hspace{1pt}\raisebox{1pt}{\tiny\ensuremath{++}}}\xspace}
\newcommand{\randomTestSampling}{\emph{random test sampling}\xspace}
\newcommand{\equidistantSampling}{\emph{equidistant sampling}\xspace}
\newcommand{\optimalTrainingDensity}{optimal training density\xspace}
\newcommand{\superiorTrainingDensity}{\emph{superior training density}\xspace}
\newcommand{\superiorSamplingScheme}{\emph{superior sampling scheme}\xspace}
\newcommand{\locallyAdaptiveModels}{\emph{locally adaptive models}\xspace}
\newcommand{\robust}{\emph{robust}\xspace}
\newcommand{\robustProperty}{\emph{robustness}\xspace}
\newcommand{\optimal}{\emph{optimal}\xspace}
\newcommand{\optimalProperty}{\emph{optimality}\xspace}
\newcommand{\modelagnostic}{\emph{model-agnostic}\xspace}
\newcommand{\modelagnosticProperty}{\emph{model-agnosticity}\xspace}

\newcommand{\SC}{\text{S.~Chmiela}\xspace}
\newcommand{\DP}{\text{D.~Panknin}\xspace}
\newcommand{\SN}{\text{S.~Nakajima}\xspace}
\newcommand{\KRM}{\text{K.-R.~M\"uller}\xspace}
\soulregister{\KRM}{7}
\soulregister{\SN}{7}
\soulregister{\eqref}{7}
\soulregister{\secref}{7}
\soulregister{\appref}{7}
\soulregister{\remref}{7}
\soulregister{\cite}{7}
\soulregister{\citep}{7}
\soulregister{\modelagnostic}{7}
\soulregister{\modelagnosticProperty}{7}
\soulregister{\robust}{7}
\soulregister{\robustProperty}{7}
\soulregister{\optimal}{7}
\soulregister{\optimalProperty}{7}
\soulregister{\LOB}{7}
\soulregister{\LFC}{7}
\soulregister{\MoE}{7}
\soulregister{\GPR}{7}
\soulregister{\AL}{7}
\soulregister{\LPS}{7}
\soulregister{\superiorTrainingDensity}{7}
\soulregister{\LPSorder}{7}
\soulregister{\nonumber}{7}
\soulregister{\label}{7}

\title{Local Function Complexity for Active Learning\\via Mixture of Gaussian Processes}

\author{\name Danny Panknin \email danny.panknin@tu-berlin.de \\
      \addr Uncertainty, Inverse Modeling and Machine Learning Group, Berlin Institute of Technology, 10587 Berlin, Germany\\
      \addr Physikalisch-Technische Bundesanstalt, 10587 Berlin, Germany
      \AND
      \name Stefan Chmiela \email stefan@chmiela.com \\
      \addr Machine Learning Department,
      Berlin Institute of Technology, 10587 Berlin, Germany\\
      \addr BIFOLD-Berlin Institute for the Foundations of Learning and Data, Germany
      \AND
      \name Klaus-Robert M\"uller \email klaus-robert.mueller@tu-berlin.de\\
      \addr Machine Learning Department,
      Berlin Institute of Technology, 10587 Berlin, Germany\\
      \addr BIFOLD-Berlin Institute for the Foundations of Learning and Data, Germany\\
      \addr Department of Artificial Intelligence, Korea University, Seoul 136-713, South Korea\\
      \addr Max Planck Institute for Informatics, 66123 Saarbrücken, Germany
      \AND
      \name Shinichi Nakajima \email nakajima@tu-berlin.de \\
      \addr Machine Learning Department,
      Berlin Institute of Technology, 10587 Berlin, Germany\\
      \addr BIFOLD-Berlin Institute for the Foundations of Learning and Data, Germany\\
      \addr RIKEN AIP, 1-4-1 Nihonbashi, Chuo-ku, Tokyo, Japan
}

\def\month{12}  
\def\year{2023} 
\def\openreview{\url{https://openreview.net/forum?id=w4MoQ39zmc}} 

\begin{document}
\newif\ifshowComments
\showCommentstrue

\maketitle

\begin{abstract}
Inhomogeneities in real-world data, e.g., due to changes in the observation noise level or variations in the structural complexity of the source function, pose a unique set of challenges for statistical inference.
Accounting for them can greatly improve predictive power when physical resources or computation time is limited. In this paper, we draw on recent theoretical results on the estimation of \emph{local function complexity} (\LFC), derived from the domain of \emph{local polynomial smoothing} (\LPS), to establish a notion of local structural complexity, which is used to develop a model-agnostic \emph{active learning} (\AL) framework.
Due to its reliance on pointwise estimates, the \LPS model class is not robust and scalable concerning large input space dimensions that typically come along with real-world problems.
Here, we derive and estimate the \emph{Gaussian process regression} (\GPR)-based analog of the \LPS-based \LFC and use it as a substitute in the above framework to make it robust and scalable.
We assess the effectiveness of our \LFC estimate in an \AL application on a prototypical low-dimensional synthetic dataset, before taking on the challenging real-world task of reconstructing a quantum chemical force field for a small organic molecule and demonstrating state-of-the-art performance with a significantly reduced training demand.
\end{abstract}

\section{Introduction}
\label{sec:intro}

Inference problems from real-world data often exhibit inhomogeneities, e.g., the noise level, the density of the data distribution, or the complexity of the target function may change over the input space.
There exist different approaches from various domains that treat specific kinds of inhomogeneities. For example, \cite{Kersting07mostlikely,cawley06} deal with heteroscedasticity by reconstructing a \emph{local noise variance} function that is used to adapt the regularization of the model locally.
Some approaches adjust bandwidths locally with respect to the input density \citep{wang2007,mackenzie2004asymmetric,moody1989fast,width_Optimization_benoudjit2002}.
Inhomogeneous complexity can also be captured using a combination of several kernel-linear models with different bandwidths, either learned jointly \citep{multiscale_SVR_zheng2006, kbp_Guigue2005} or hierarchically \citep{multiscale_SVR_ferrari2010, hierarchical_SVR_Bellocchio2012}. 
The most widely applicable models treat all types of aforementioned inhomogeneities in a unified way \citep{tresp2001mixtures,panknin2021optimal}. Namely, they locally adapt bandwidths or regularization according to the inhomogeneities in noise, complexity, and data density.
%
%
While this is the path we will pursue, the focus in this work will be on inhomogeneous complexity under the assumption of homoscedastic noise. In addition, we will investigate our proposed estimates in a heteroscedastic setting to demonstrate negligible practical limitations.

Exposing inhomogeneities sheds light on the informativeness of certain locations of the input space, which subsequently can be used to guide the sampling process during training---also known as \emph{active learning} (\AL). \AL \citep{kiefer1959optimum,mackay1992information,seung1992query,seo2000gaussian} is a powerful tool to enhance the training process of a model when the acquisition of labeled training data is expensive.
It has been successfully implemented in various regression applications like reinforcement learning \citep{teytaud2007active}, wind speed forecasting \citep{douak2013}, and optimal control \citep{wu2020active}. 

Nowadays, \emph{machine learning} (\ML) methods are increasingly deployed in physical modeling applications across various disciplines. In that setting, the labels that are necessary for model training are typically expensive as they stem, e.g., from computationally expensive first-principles calculations \citep{chmiela2017} or even laboratory experiments.
Due to the need for effective training datasets, \AL has become an integral part of ever-growing importance in real-world applications, e.g., in the domains of pharmaceutics \citep{warmuth2003active} and quantum chemistry \citep{gubaev2018machine,tang2019prediction,huang2020quantum}---which raises the demand for \AL solutions and the importance of \AL research in general.

Through the advance of \ML in scientific fields that hold the potential for significant impact, new regression problems emerge, for which there is initially only scarce domain knowledge while they simultaneously require thousands to tens of thousands of training samples for \ML models to operate at an acceptable performance level.
Regarding \AL, these two characteristics of regression problems are hard to reconcile:
\vspace*{-1mm}
\par\noindent
Due to insufficient domain knowledge on the one hand, a suitable \AL approach shall be robust, since unjustified assumptions may result in a training performance that is even worse than \randomTestSampling.
By \randomTestSampling, we refer to the naive training data construction that draws samples i.i.d.~according to the test distribution.
Additionally, the \AL approach shall be model-agnostic since the \SOTA is ever-evolving for this particular kind of regression problem.
For these reasons, practitioners prefer model-free \AL approaches for regression \citep{wu2019pool} over sophisticated, model-based \AL approaches with strong assumptions as the former are inherently
robust and model-agnostic by ignoring label information.

On the other hand, it is preferable that an \AL approach outperforms \randomTestSampling even at large training sizes.
In the following, we will measure the \AL performance by the \emph{relative required sample size} $\relSampleSizeSymbol > 0$, which asymptotically equates the performance of $n\cdot\relSampleSizeSymbol$ active training samples to $n$ random test samples (see \defref{def:relSampleSize}). Accordingly, we call an \AL approach asymptotically superior to \randomTestSampling, if $\relSampleSizeSymbol < 1$.
Unfortunately, the performance gain of model-free \AL approaches that we observe at small training sizes over \randomTestSampling eventually diminishes completely ($\relSampleSizeSymbol = 1$) with growing training size.

For the described learning task, we therefore require an \AL approach that is model-based but comes with mild regularity assumptions at the same time to feature robustness and model-agnosticity to a certain extent.

Recently, \cite{panknin2021optimal} addressed the outlined \AL scenario, where the fundamental idea is to analyze the distribution of the optimal training set of a model in the asymptotic limit of the sample size.

Assuming that this limiting distribution exists, they then propose to sample training data in a \emph{top-down} manner from this very distribution, knowing that with growing sample size the training set will eventually become optimal.
By a \emph{top-down} \AL approach, we mean an (infinite) training data refinement process $x_1', x_2', \ldots$ such that---when optimizing an \AL criterion with respect to $\{x_1,\ldots,x_n\}$---$\{x_1',\ldots,x_n'\}$ asymptotically becomes a respective optimizer as $n\rightarrow\infty$.
They have shown for the \emph{local polynomial smoothing} (\LPS) model class \citep{cleveland1988locally} that the asymptotically optimal distribution exists, whose density furthermore factorizes into contributions of the test density, heteroscedastic noise, and \emph{local function complexity} (\LFC)---a measure of the local structural complexity of the regression function.
Intuitively, \LFC scales with the local amount of variation of the regression function. It is essentially estimated as the reciprocal determinant of the \emph{locally optimal kernel bandwidth} (\LOB) of the \LPS model, calibrated for the local effects of the training input density and noise level.
Given a small but sufficient training set, these factors can be estimated, allowing the construction of the \optimalTrainingDensity and subsequently enabling the refinement of the training data towards asymptotic optimality.

While the previous work by \cite{panknin2021optimal} provides a theoretically sound solution to our considered \AL scenario, the required pointwise estimates that are inherent to the \LPS model class prevent scalability with regard to the input space dimension \inputDimension.
The goal of our work is to extend the above approach in a scalable way.
The key idea is to build the required estimate of \LFC based on the \LOB of the related \emph{Gaussian process} (\GaussianProcess) model class that can naturally deal with high input space dimensions.
Subsequently, we plug our scalable \LFC estimate into the \AL framework of the existing method, whose functioning is justified by the method's model-agnostic nature.

It is particularly the almost assumption-free nature of \LPS that made the results of \cite{panknin2021optimal} model-agnostic.
To that effect, we base our results on the nonparametric, adaptive bandwidth \emph{Gaussian process regression} (\GPR) model to preserve this property.
While we lose the strict asymptotic sampling optimality this way,
we expect it to be reasonably close due to the model-agnosticity nevertheless.
We refer to the resulting training density as the
\superiorTrainingDensity for \locallyAdaptiveModels, by which we refer to models that adapt to the considered inhomogeneities, namely heteroscedasticity and inhomogeneous complexity.
Note that it is first and foremost superior for our adaptive bandwidth \GPR model.

\begin{figure}[htb]
  \centering
  \begin{overpic}[width=1.0\linewidth]
  {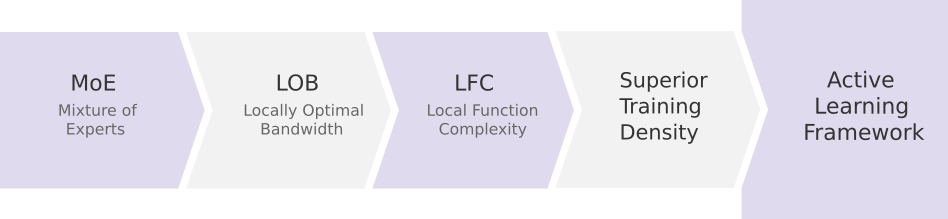}
   \put (1,4.3) {{\bf\fontfamily{cmss}\selectfont \footnotesize\textcolor{black!50}{\secref{subsec:modelArchitecture}}}}
   \put (41,4.3) {{\bf\fontfamily{cmss}\selectfont\footnotesize\textcolor{black!50}{\secref{subsec:GPRlfcEstimate}}}}
   \put (79.5,1) {{\bf\fontfamily{cmss}\selectfont\footnotesize\textcolor{black!50}{\secref{subsec:ALprocedure}}}}
  \end{overpic}
  \vspace*{-0.7cm}
  \caption{An overview of the steps of our contribution and how they are interlinked. We will elaborate on the main steps in the specified sections.}
  \vspace*{-0.0cm}
  \label{fig:summary_figure_new}
\end{figure}

\figref{fig:summary_figure_new} summarizes all steps of our contribution and shows how they are interlinked and in which sections they will be discussed.
Specifically, we contribute in two ways:
\vspace*{-1mm}
\par\noindent
\myParagraph{Theoretical contribution} Assuming homoscedastic data, we propose a \GPR-based \LFC estimate which is inspired by the design of the \LPS-based \LFC estimate. Here, we need to respect the scaling behavior of \LOB of \GPR that differs from the \LPS case, where we use asymptotic results on the scaling of optimal bandwidths for \GPR, as described in \secref{subsec:GPRbandwidthScaling}.
Making use of the model-agnosticity of the \optimalTrainingDensity of \LPS, we replace the herein contained \LPS-based \LFC estimate for our \GPR-based \LFC estimate to obtain a \superiorTrainingDensity for \locallyAdaptiveModels. 
From this point, we can implement the \AL framework by \cite{panknin2021optimal}, for which we propose a novel pool-based formulation.
Both, our \LFC and density estimate will inherit the scalability of the deployed \GPR-based \LOB estimate.

\myParagraph{Methodological contribution} We propose a scalable \LOB estimate for \GPR as the weighted average of bandwidth candidates, where the weights are given by the gate function of a \emph{sparse mixture of \GaussianProcesses} model---a special case of a \emph{mixture of experts} (\MoE) model \citep{jacobs1991adaptive, jordan1994hierarchical, pawelzik1996annealed}.
Here, each expert of the \MoE is a \GPR model that holds an individual, fixed bandwidth candidate.
We construct the \MoE in \emph{PyTorch} \citep{paszke2019pytorch} out of well-established components from the related work and design a training objective that is regularized with respect to small bandwidth choices to obtain a robust and reasonable \LOB estimate in the end.
In addition, we provide an implementation\footnote{\url{https://github.com/DPanknin/modelagnostic_superior_training}} of both, the \AL framework and our model.
Finally, we propose a novel model-agnostic way of choosing \emph{inducing points} (\inducingPoints) of sparse \GPR models: Respecting \LFC, we place additional basis functions of a kernel method in more complex regions while removing basis functions in simpler regions of the input space.

\vspace*{-10pt}

To show the capabilities of our approach, we consider two inhomogeneously complex regression problems:
\vspace*{-1mm}
\par\noindent
\myParagraph{The Doppler function}
In a controlled setting of 1-dimensional synthetic data, we will first analyze our \MoE model and the proposed estimates of \LFC and the \superiorTrainingDensity, where we will demonstrate the asymptotic superior performance of our \superiorSamplingScheme and compare to related work. By the \superiorSamplingScheme, we refer to i.i.d.~sampling from the \superiorTrainingDensity.

\myParagraph{Force field reconstruction}
Quantum interactions exhibit multi-scale behavior due to the complex electronic interactions that give rise to any observable property of interest, like the total energy or atomic forces of a system \citep{bereau2018non, yao2018tensormol, grisafi2019incorporating, ko2021fourth, unke2021spookynet}.
To demonstrate the scalability of our approach, we consider a force field reconstruction problem of a molecule with 27 dimensions, where the application of the \LPS-based \AL framework by \cite{panknin2021optimal} is intractable.
Besides the asymptotic superior \AL performance, we gain insights into the local structural complexity of this high-dimensional molecular configuration space through visualizations of the scalar-valued \LFC function.

We begin by discussing our work in the context of related work in \secref{sec:relatedWork}.
Next, we give a formal definition of the considered regression problem and the asymptotic \AL task, and review asymptotic results for \LPS and \GPR in \secref{sec:theoryPreliminaries}.
In \secref{sec:theoryIsotropicOptimalSampling}, we describe our \MoE model and derive the \GPR-based \LFC and \superiorTrainingDensity estimates. In \secref{sec:experiments}, we then describe our experiments and results,
which will be further discussed in \secref{sec:discussion}.
We finally conclude in \secref{sec:conclusion}.

\section{Related work}
\label{sec:relatedWork}

\myParagraph{Choice of \MoE, experts and the gate}
The common assumption of \MoE approaches is that
the overall problem to infer is too complex for a single, comparably simple expert. This is the case, for example in regression of nonstationary or piecewise continuous data, and naturally in classification where each cluster shape may follow its own pattern.
In such a scenario each expert of the \MoE model can specialize in modeling an individual, (through the lens of a single expert) incompatible subset of the data, where the gate learns a soft assignment of data to the experts.
Under these assumptions, the hyperparameters of each expert can be tuned individually on the respective assigned data subset.
In the light of this paradigm, there exist several instances of mixture of \GaussianProcesses, for example, \cite{tresp2001mixtures,meeds2006alternative,yuan2009variational, yang2011efficient,chen2014precise}.

In our work, we aim to infer a single regression problem, where there is no such segmentation as described above: Each individual (reasonably specified) expert of our mixture model is eventually capable of modeling the whole problem on its own.
Yet, if the problem possesses an inhomogeneous structure, the prediction performance can be increased by allowing for a local individual bandwidth choice.
Therefore, we deviate from the common \MoE paradigm, sharing all those parameters across the experts that describe the regression function.
This less common assumption was also made by \cite{pawelzik1996annealed}, where---locally dependent---some experts are expected to perform superior compared to the others.

For the expert and gate components of our \MoE model, we focus on the sparse, variational \GPR model (see, e.g.,~\cite{hensman2015scalable}) trained by stochastic gradient descent.
However, there exist other (sparse) \GPR approaches that could be considered for the gate or the experts of our \MoE model. Some are computationally appealing as they solve for the inducing value distribution analytically \citep{seeger2003fast,snelson2005sparse,titsias2009variational} or do not require inducing points in the first place in the case of a full \GPR model (see, e.g.~\cite{williams1996gaussian}).

Particularly the expert models of our \MoE model can be exchanged for arbitrary sparse and full formulations of \GPR, as long as we can access the posterior predictive distribution. We give a short summary of these model alternatives in \appref{subsec:analyticGP}~and~\ref{subsec:analyticSparseGP}, which are also included in our provided implementation.

For the gate, however, analytic approaches come with complications as they require labels. Such labels do not exist for the gate, and so we would need to train the \MoE in an \emph{expectation maximization} loop, where the likelihood of an expert to have produced a training label functions as a pseudo-label to the gate.

\myParagraph{Alternative nonstationary \GaussianProcesses}
As opposed to a standard \GaussianProcess that features a stationary covariance structure,
our \MoE model with \GaussianProcess experts of individual bandwidths can be interpreted as a nonstationary \GaussianProcess.
Apart from \MoE model architectures, there exist other approaches to construct a nonstationary \GaussianProcess:
\vspace*{-1mm}
\par\noindent
Closely related to the \MoE model, \citep{rulliere2018nested} proposed to aggregate individual expert model outputs through a nested \GaussianProcess. Note that this approach is not well-suited for our purpose as the aggregation weights lack the interpretation of a hidden classifier, leaving the subsequent \LOB estimation (as in \eqref{eq:gprLOB}) open.
\citep{gramacy2008bayesian} deploy individual stationary \GaussianProcesses on local patches of the input space that are given by an input space partitioning of a tree.

\citep{gramacy2015local} identify subsets of the training data that are necessary to resemble the \GaussianProcess covariance structure at each individual evaluation point. Careful bandwidth choice for each subset then yields a nonstationary \GaussianProcess as well as local bandwidths.
\cite{roininen2019hyperpriors} obtain local bandwidths by imposing a hyperprior on the bandwidth of a \GaussianProcess.
Note that our work intends to elaborate the estimation of \LFC and model-agnostic superior training, given any estimate of \LOB of \GPR. We deployed an \MoE approach as a simple means to obtain these estimates. The \MoE component in our \LFC and superior training density estimates may be readily replaced by the approaches of \cite{gramacy2015local} or \cite{roininen2019hyperpriors}.

In \emph{deep Gaussian process} (\DGP) regression \citep{damianou2013deep}, the inputs are mapped through one or more hidden (stationary) \GaussianProcesses.
This warping of the input space yields a nonstationary covariance structure of \DGP.
An approximate \DGP model through random feature expansions was exercised by \cite{roininen2019hyperpriors}.
\cite{sauer2023active} discuss active learning for \DGP regression.
We will implement the \DGP model as well as the \AL scheme of \cite{sauer2023active} and compare the \AL performance of our superior \AL scheme on this very model to demonstrate the model-agnosticity of our work in \secref{subsec:doppler}.

\myParagraph{\inducingPoint selection}
In our work (see~\secref{subsec:inducingPointInit}), we choose the \inducingPoint locations of the gate and the experts of our \MoE model in a diverse and representative way but also in alignment with the structural complexity of the target function, interpreting this choice as a nested \AL problem. There are a variety of \inducingPoint selection approaches in the literature.
\cite{zhang2008improved} interpreted the choice of \inducingPoint locations from a geometric view that is similar to ours: They derived a bound on the reconstruction error of a full kernel matrix by a Nyström low-rank approximation in terms of the sum of distances of all training points to their nearest \inducingPoint. This exposes a local minimum by letting the \inducingPoint locations be the result of \emph{k-Means clustering}. This choice of \inducingPoint locations is representative and diverse, while it solely considers input space information. In this sense, our approach extends their work by additionally considering label information. This and our approach draw a fixed number of \inducingPoints at once.
There are also a lot of Nyström method based \inducingPoint selection approaches that select columns of the full kernel matrix according to a fixed distribution \citep{drineas2005nystrom} or one-by-one in a greedy, adaptive way \citep{smola2000sparse,fine2001efficient,seeger2003fast}. An intensive overview of Nyström method based \inducingPoint selection methods was given by \cite{kumar2012sampling}, where they also analyzed ensembles of low-rank approximations.
We compare our proposed \inducingPoint choice \eqref{eq:optimizedIPdistribution} to the greedy fast forward \inducingPoint selection approach
by \cite{seeger2003fast} in \secref{subsec:doppler}.

\cite{moss2023inducing} incorporate a \emph{quality function} into a diverse \inducingPoint selection process that can be specified flexibly. They consider Bayesian optimization rather than regression, they exercise a quality function proportional to the label. However, other measures of informativeness that are better suited for regression could be deployed. Note that \LFC would be a possible candidate for this purpose.

\myParagraph{The \AL scenario}
In this work, we consider model-agnostic \AL with persistent performance at large (or even asymptotic) training size as opposed to the common \AL paradigm that is concerned with small sample sizes. In this sense, we delimit ourselves from \AL approaches that are tied to a model, e.g., when they are based on a parametric model \citep{kiefer1959optimum,mackay1992information,he2010laplacian,sugiyama2009pool,gubaev2018machine}, or which refine training data \emph{bottom-up} in a greedy way to maximize its information content at small sample size, where the information is either based on the inputs only \citep{seo2000gaussian,teytaud2007active,yu2010passive,wu2019pool,liu2021pool} or also incorporates the labels \citep{burbidge2007active,cai2013maximizing}.
By a \emph{bottom-up} \AL approach, we mean a training data refinement process that is constructed by choosing the n\textsuperscript{th} input $x_n$ as the optimizer of an \AL criterion with respect to $\{x_1,\ldots,x_n\}$, when keeping the previously drawn inputs $\{x_1,\ldots,x_{n-1}\}$ (with labels $\{y_1,\ldots,y_{n-1}\}$) fixed.

Our work is therefore complementary to the latter kind of approaches which can be better suited in another \AL scenario. For example, if there is enough domain knowledge such that we can deduce a reasonable parametric model without the need for a model change in hindsight, an active sampling scheme based on this model will be best. Our category of interest is for the other case, when domain knowledge is scarce, where we have no idea about the regularity or structure of the problem to decide on a terminal model.
Here, for small training sizes (and particularly from scratch), input space geometric arguments \citep{teytaud2007active,yu2010passive,wu2019pool,liu2021pool} are applied in practice. However, as already noted in the introduction, their benefit is limited to this small sample size regime, which we will demonstrate on our synthetic dataset. They serve reasonably for the initialization of supervised \AL approaches, including ours, nevertheless.

Regarding our considered \AL scenario, \cite{panknin2021optimal} have recently proposed an \AL framework based on the \LPS model class, where training samples are added so as to minimize the \emph{mean integrated squared error} (\MISE) in the asymptotic limit. This approach is therefore provably asymptotically superior to \randomTestSampling.
Additionally, it is robust since the \LPS model is almost free of regularity assumptions.
Finally, their \LPS-based solution then showed to be model-agnostic: On the one hand, this is indicated theoretically by the fact that the \LPS model has only indirect influence on the asymptotic form of \LFC and the \optimalTrainingDensity since the predictor is asymptotically not involved (see, e.g., Eq.~\eqref{eq:isotropicAsymptoticComplexityLPS}); On the other hand, this is validated empirically by assessing the performance of their \LPS-based training dataset construction under \emph{reasonable} model change in hindsight.
This model change is restricted to \locallyAdaptiveModels. Here, \cite{panknin2021optimal} observed a consistent performance superior to \randomTestSampling when training a random forest model and a \emph{radial basis function} (\RBF)-network \citep{moody1989fast}, using their proposed training dataset.

\myParagraph{\AL for classification}
Note that the outlined \AL scenario can be solved more easily for classification:
\vspace*{-1mm}
\par\noindent Here, \AL is intuitively about the identification and rendering of the decision boundaries, which is inherently a model-agnostic task.
In addition, since the decision boundaries are a submanifold of the input space \inputSpace, a substantial part of \inputSpace can be spared when selecting training samples. Therefore, \AL for classification leverages the decay of the generalization error from a polynomial to an exponential law \citep{seung1992query} over \randomTestSampling.
For the above reasons, \AL for classification has been applied successfully in practice \citep{lewis1994sequential,roy01,goudjil2018novel,warmuth2003active,pasolli2010,saito2015robust,bressan2019breast,sener2018active,beluch2018power,haut2018active,tong2001support,he2010laplacian}.
In contrast, the performance gain of \AL for regression is more limited in the sense that, under weak assumptions, we are tied to the decay law of the generalization error of \randomTestSampling \citep{gyorfi2002distribution,willett2005faster}.

\myParagraph{GP uncertainty sampling} 
There exists a lot of research on \AL for \GPR \citep{seo2000gaussian,pasolli2011gaussian,schreiter2015safe,yue2020active}, which is typically based on minimizing prediction uncertainties of the model.
With our proposed \AL approach being based on \GPR models, this research area is the most related competitor to our work.

For a standard \GPR model, the prediction uncertainty is the higher the farther away we move from training inputs. In this way, \emph{GP uncertainty sampling} samples (pseudo\=/)uniformly from the input space
which makes up for a low-dispersion sequence \citep{niederreiter1988low} (see~\defref{def:dispersion}). Note that standard \GPU is an input space geometric argument since it does not depend on the regression function to infer.
As already indicated in the introduction and as we will show in \secref{subsec:doppler}, input space geometric arguments feature no benefit regarding asymptotic \AL performance.

Since our model is a mixture of \GPR experts, it is straightforward to derive its uncertainty as a \emph{mixture of Gaussian process uncertainties} (\MoGPU) by simply weighting the predictive variances of all experts with respect to the gate output (see \eqref{eq:MoGPU} for a definition). As opposed to \GPU, \MoGPU can cope with structural inhomogeneities. Therefore, we consider \MoGPU as a fair baseline competitor to our \superiorSamplingScheme and compare both in \secref{subsec:doppler}.

\section{Preliminaries}
\label{sec:theoryPreliminaries}
We will now give a formal definition of the regression task, the \AL objective and \LOB, and a short review of the asymptotic results on \LFC and the optimal training distribution of the \LPS model in \secref{subsec:formalIntroduction} and \ref{subsec:LOB_LFC_pOpt}. Then we recap asymptotic results on the optimal bandwidth of \GPR in \secref{subsec:GPRbandwidthScaling} and known models in \secref{subsec:modelPreliminaries} that will serve as building blocks of our proposed adaptive bandwidth \MoE model later on.

In the following, we denote by $\diag(z)\in\reals{d\times d}$ the diagonal matrix with the entries of the vector $z\in\reals{\inputDimension}$ on its diagonal and by $\idMatrix{\inputDimension} = \diag(\ones{\inputDimension})$ the identity matrix, where $\ones{\inputDimension}$ is the vector of ones in $\reals{\inputDimension}$.

\subsection{Formal definition of the regression task and \AL objective}
\label{subsec:formalIntroduction}
Let $f$ be the target regression function defined on an input space $\inputSpace\subset\reals{\inputDimension}$ that we want to infer from noisy observations $y_i^{} = f(x_i^{}) + \varepsilon_i^{}$, 
where $x_i^{} \in \inputSpace$ are the training inputs and $\varepsilon_i^{}$ is independently drawn noise from a distribution with mean $\E[\varepsilon_i^{}] = 0$ and local noise variance $\Var[\varepsilon_i^{}] = v(x_i^{})$.
We denote a training set by $(\trainingInputs{n}, \trainingLabels{n})$, where $\trainingInputs{n} = (x_1,\ldots,x_n)\in\inputSpace^n$ and $\trainingLabels{n} = (y_1,\ldots,y_n) \in \reals{n}$.
For a given model class $\widehat{f}$ that returns a predictor $\widehat{f}_{\trainingInputs{n}, \trainingLabels{n}}$ for a training set $(\trainingInputs{n}, \trainingLabels{n})$, we can define the pointwise \emph{conditional mean squared error} of $\widehat{f}$ in $x\in\inputSpace$, given $\trainingInputs{n}$, by
\begin{align}
\label{eq:conditionalMSE}
\MSE\left(x, \widehat{f} | \trainingInputs{n}\right) = \sideset{}{_{\trainingLabels{n}}}\E \left[(\widehat{f}_{\trainingInputs{n}, \trainingLabels{n}}(x) - f(x))^2\right] = \sideset{}{_{\bm{\varepsilon}_n}}\E \left[(\widehat{f}_{\trainingInputs{n}, f(\trainingInputs{n})+\bm{\varepsilon}_n}(x) - f(x))^2\right].
\end{align}
Note that via marginalization the conditional mean squared error is no function of the training labels $\trainingLabels{n}$.
Given a test probability density $q \in \diffableFunctions{\inputSpace, \nonnegativeReal{}}{0}$ such that
$\mathop{\mathlarger{\int}}_{\hspace*{-5pt}\inputSpace}q(x)dx = 1$,
the \emph{conditional mean integrated squared error} of the model under the given training set is then defined as
\begin{align}
\label{eq:conditionalMISE}
\MISE\left(q, \widehat{f} | \trainingInputs{n}\right) = \textstyle \mathop{\mathlarger{\int}}_{\hspace*{-5pt}\inputSpace} \MSE\left(x, \widehat{f} | \trainingInputs{n}\right) q(x) dx.
\end{align}
With these preparations, the \AL task is to construct a training dataset $(\optTrainingInputs{n},\optTrainingLabels{n})$ such that
\begin{align}
 \label{eq:ALtask}
 \optTrainingInputs{n} \approx \textstyle \argmin_{\trainingInputs{n} \in \inputSpace^n_{} }\MISE\left(q, \widehat{f} | \trainingInputs{n}\right).
\end{align}

\subsection{Locally optimal bandwidths, function complexity, and optimal training}
\label{subsec:LOB_LFC_pOpt}
Let $\widehat{f}^\bandwidth$ be a family of \emph{kernel machines}
which is characterized by a positive definite bandwidth matrix parameter $\bandwidth \in \posDefSet{\inputDimension}$ of an RBF kernel
$k^\bandwidth(x,x') := \abs{\bandwidth}^{-1} \phi(\pnorm{\bandwidth^{-1}(x-x')}{})$ for a monotonically decreasing function $\fctn{\phi}{\nonnegativeReal{}}{\nonnegativeReal{}}$. The well known Gaussian kernel is for example implemented by $\phi(z) = \exp\{-\frac{1}{2}z^2\}$.

Given a bandwidth space $\SigmaSpace \subseteq \posDefSet{\inputDimension}$ we define the \LOB function of $\widehat{f}$ by
\begin{align}
\label{eq:LOBDefinition}
 \LOBfunctionOfLPS{\widehat{f}}{n}(x)
  = \textstyle \argmin_{\bandwidth \in \SigmaSpace}\MSE\left(x, \widehat{f}^\bandwidth | \trainingInputs{n}\right),
\end{align}
assuming that this minimizer uniquely exists for all $x\in\inputSpace$.

Denote by $\predictorLPS{\LPSorder}{\bandwidth}$ the predictor of the \LPS model of order \LPSorder under bandwidth \bandwidth
and by $\displaystyle\LOBfunctionOfLPS{\LPSorder}{n}$ $\displaystyle := \LOBfunctionOfLPS{\predictorLPS{\LPSorder}{}}{n}$ the \LOB function \eqref{eq:LOBDefinition} of \LPS, if it is well-defined. This is the case, e.g., for the \emph{isotropic} bandwidths space $\SigmaSpace = \condset{\sigma\idMatrix{\inputDimension}}{\sigma > 0}$ under mild assumptions\footnote{For \LOB being well-defined in the isotropic case, we generally require a non-vanishing bias and variance in terms of a bias-variance-decomposition of the \MSE of the predictor in x, for all $x \in \inputSpace$. See, e.g., Eq.~\eqref{eq:lpsLOBDefinition} for the \LPS predictor $\predictorLPS{\LPSorder}{}$, or \cite{silverman1986density,wand1994kernel} in more general.}, where we particularly can write $\LOBfunctionOfLPS{\LPSorder}{n}(x) = \lobfunctionOfLPS{\LPSorder}{n}(x)\idMatrix{\inputDimension}$.
We refer to \appref{sec:optimalSamplingSupplement} for details on the \LPS model and asymptotic results.
For the optimal predictor
\begin{align}
 \label{eq:LPSoptimalPredictor}
    \fLPS{\LPSorder} := \predictorLPS{\LPSorder}{\LOBfunctionOfLPS{\LPSorder}{n}(x)}(x)
\end{align}
of \LPS,
letting $\widehat{f} = \fLPS{\LPSorder}$ in Eq.~\eqref{eq:ALtask},
\cite{panknin2021optimal} have shown that there exists an \optimalTrainingDensity $\pOptLPS{\LPSorder}{n}$ that allows the optimal training inputs in Eq.~\eqref{eq:ALtask} to be asymptotically obtained by independently and identically sampling
$\optTrainingInputs{n} \sim \pOptLPS{\LPSorder}{n}$.
They have also shown that this density exhibits a closed-form
\begin{align}
    \label{eq:optimalSamplingFiniteLPS}
    \pOptLPS{\LPSorder}{n}(x) \propto \textstyle \left[\fctnComplexityLPS{\LPSorder}{n}(x) q(x)\right]^{\frac{2(\LPSorder+1)+\inputDimension}{4(\LPSorder+1)+\inputDimension}}v(x)^{\frac{2(\LPSorder+1)}{4(\LPSorder+1)+\inputDimension}}(1 + o(1)),
\end{align}
where for an arbitrary training dataset $(\trainingInputs{n}, \trainingLabels{n})$ with $\trainingInputs{n} \sim p$, the \LFC of \LPS is defined by
\begin{align}
    \label{eq:isotropicComplexityLPS}
    \fctnComplexityLPS{\LPSorder}{n}(x) := \left[\frac{v(x)}{p(x)n}\right]^{\frac{\inputDimension}{2(\LPSorder+1)+\inputDimension}}\abs{\LOBfunctionOfLPS{\LPSorder}{n}(x)}^{-1} = \left[\frac{v(x)}{p(x)n}\right]^{\frac{\inputDimension}{2(\LPSorder+1)+\inputDimension}}\lobfunctionOfLPS{\LPSorder}{n}(x)^{-d}.
\end{align}
The \LFC in \eqref{eq:isotropicComplexityLPS} asymptotically solely depends on the behavior of $f$ as opposed to $p,v$, and $n$: It scales with the local variation of $f$ in the vicinity of $x$. 
For example, 
\begin{align}
\label{eq:isotropicAsymptoticComplexityLPS}
\fctnComplexityLPS{1}{n}(x) \propto \trace(D^2_f(x))^\frac{2\inputDimension}{2(\LPSorder+1)+\inputDimension} (1 + o(1))
\end{align}
is a function of the trace of the Hessian of $f$ \citep{fan1997local}.

The optimal density $\pOptLPS{\LPSorder}{n}$ in \eqref{eq:optimalSamplingFiniteLPS} implies that we require more training data where the problem is locally more complex (large $\fctnComplexityLPS{\LPSorder}{n}$) or noisy (large $v$), or where test instances are more likely (large $q$).
As already noted in the introduction, the results to \LFC and the \optimalTrainingDensity of \LPS indicate their problem intrinsic nature, as they reflect no direct dependence on the \LPS model
except for the order \LPSorder. Note that for $f \in \diffableFunctions{\inputSpace,\R}{\alpha}$, there is a canonical choice $\LPSorder = \lceil\alpha\rceil - 1$ of the \LPS model order. When deriving \LFC under this canonical-order model, we consider the dependence of the associated \LFC and the \optimalTrainingDensity on \LPSorder negligible, as its choice is driven by the problem intrinsic regularity.

In practice, we obtain $\optTrainingInputs{n} \sim \pOptLPS{\LPSorder}{n}$ by estimating Eq.~\eqref{eq:optimalSamplingFiniteLPS}~and~\eqref{eq:isotropicComplexityLPS} from $(\trainingInputs{n'}, \trainingLabels{n'})$ with $\trainingInputs{n'} \sim p$ for an arbitrary training density $p$, where $n' < n$, followed by adding the remaining $n-n'$ inputs appropriately (see \secref{subsec:ALprocedure}).

The construction of $\pOptLPS{\LPSorder}{n}$ crucially depends on reliable estimates of \LOB as the key ingredient for the estimation of \LFC. While \cite{panknin2021optimal} provide such an estimate based on Lepski's method
\citep{lepski1991problem,lepski1997optimal},
it does not scale well with increasing input space dimension \inputDimension.
This is because pointwise estimates suffer from the \emph{curse of dimensionality} regarding robustness and computational feasibility.
The goal of this work is to implement the above \AL framework but based on a functional \LOB estimate in the domain of \GPR instead of \LPS, since the \GPR model class can naturally deal with high input space dimensions \citep{williams1996gaussian}.
Relying on the model-agnosticity, we expect that \LOB estimates based on \LPS can be exchanged for \LOB estimates based on \GPR in the formulation of \LFC and $\pOptLPS{\LPSorder}{n}$ when matching the degree \LPSorder to the smoothness of the regression function appropriately.

\subsection{On the scaling of \GPR bandwidths}
\label{subsec:GPRbandwidthScaling}

The major difference between \LPS and \GPR is that we keep a fixed model complexity---in the sense of the number of basis functions---in the former while there is varying model complexity in the latter as we add further training instances. E.g., under the Gaussian kernel the model complexity of \GPR grows infinitely.
When the regularity of the kernel and the target function $f$ match, then, as soon as the training size $n$ becomes large enough, there is no need for further shrinkage of the bandwidth to reproduce $f$ with \GPR in the asymptotic limit. In particular, given enough samples, there is no need for local bandwidth adaption.

However, there is a mismatch if $f \in \diffableFunctions{\inputSpace,\R}{\alpha}$ is $\alpha$-times continuously differentiable since the Gaussian kernel is infinitely often continuously differentiable.
As shown by \cite{van2007bayesian, van2009adaptive}, in order to obtain optimal \emph{minimax}-convergence of the predictor (except for logarithmic factors), the associated (global) bandwidth has to follow the asymptotic law
\begin{align}
\label{eq:GPRbandwidthSamplesizeProportionality}
\LOBfunctionOfLPS{\GPR}{n} \propto n^{-\frac{1}{2\alpha + d}}.
\end{align}
Note that for $f \in \diffableFunctions{\inputSpace,\R}{\alpha}$, where the theoretical results of \LPS apply, the scaling factor $n^{-\frac{1}{2\alpha + d}}$ of \LOB in sample size matches exactly for both classes, \LPS and \GPR.
In our work, we will use \eqref{eq:GPRbandwidthSamplesizeProportionality} to deduce a \GPR-based \LFC estimate in analogy to the \LPS-based \LFC estimate \eqref{eq:isotropicComplexityLPS} by \cite{panknin2021optimal}.

\subsection{Preliminaries on the applied models}
\label{subsec:modelPreliminaries}
We will now introduce the models that we implement in this work. 
For the \RBF-kernel $k$, we define the kernel matrix between $X\in\inputSpace^n$ and $X'\in\inputSpace^m$ as $\kernelMatrix{\bandwidth}(X, X') =  \left[k^\bandwidth(x,x')\right]_{x\in X, x'\in X'}$.
As a shorthand notation we furthermore define $\kernelMatrix{\bandwidth}(X) := \kernelMatrix{\bandwidth}(X, X)$.

\subsubsection{Sparse variational Gaussian processes}
\label{subsubsec:sparseVarGP}
We define the sparse \GPR model $\widehat{y} \sim \SVGP{\theta}$ (see, e.g.~\cite{williams1996gaussian,hensman2015scalable}) as follows:
The sparse \GaussianProcess is described by the (hyper-) parameters $\theta = (\mu, \lambda, \gpNoiseVariance, \bandwidth, \expertInducingPoints{}, \expertInducingValueMean{}, \expertInducingValueCov{})$, which are the global constant prior mean $\mu$, the regularization parameter $\lambda$, the label noise variance function $\gpNoiseVariance$, the bandwidth matrix $\bandwidth$ of the kernel and the prior distribution, given by the \inducingPoint locations $\expertInducingPoints{} \in \inputSpace^m$ as well as their inducing value distribution, characterized by the moments $\expertInducingValueMean{}$ and $\expertInducingValueCov{}$. That is, for the inducing values $\expertInducingValues{}$ of $\expertInducingPoints{}$ we assume $\expertInducingValues{} = \widehat{y}(\expertInducingPoints{}) \sim \Gauss{\cdot}{\expertInducingValueMean{}}{\expertInducingValueCov{}}$.
Here, the degree of sparsity is described by $m$ \inducingPoints: This number can be fixed in advance or gradually increased with training size $n$, where the increase $m_n = o[n]$ is typically much slower than $n$.
If we can assume homoscedastic noise, we let $\gpNoiseVariance(x) \equiv \sigma_\varepsilon^2$.

The sparse \GaussianProcess then outputs
\begin{align}
 \label{eq:predictiveDistribution}
 \widehat{y}(\evalInputs) \sim \Gauss{\cdot}{\bm{\mu}^*(\evalInputs)}{\bm{C}^*(\evalInputs) | \theta_e}
\end{align}
for the mean function
\begin{align}
 \label{eq:predictiveMean}
 \bm{\mu}^*(\evalInputs) = \bm{K}_{*\ipSymbol} \bm{K}_{\ipSymbol}^{-1/2} (\expertInducingValueWhitenedMean{} - \bm{K}_{\ipSymbol}^{-1/2} \bm{\mu}(\expertInducingPoints{})) + \bm{\mu}(\evalInputs),
\end{align}
and the covariance function
\begin{align}
 \label{eq:predictiveVar}
 \bm{C}^*(\evalInputs) = \lambda\left[\bm{K}_* + \bm{K}_{*\ipSymbol} \bm{K}_{\ipSymbol}^{-1/2} (\expertInducingValueWhitenedCov{} - \idMatrix{m}) \bm{K}_{\ipSymbol}^{-1/2} \bm{K}_{*\ipSymbol}^\top\right] + \diag(\gpNoiseVariance(\evalInputs)),
\end{align}
where $\expertInducingValueWhitenedMean{} = \bm{K}_{\ipSymbol}^{-1/2}\expertInducingValueMean{}$ and $\expertInducingValueWhitenedCov{} = \bm{K}_{\ipSymbol}^{-1/2}\expertInducingValueCov{}\bm{K}_{\ipSymbol}^{-1/2}$ are the whitened moments of the inducing value distribution (\cite{pleiss2020fast}, Sec.~5.1), and we have defined $\bm{K}_* = \kernelMatrix{\bandwidth}(\evalInputs)$, $\bm{K}_{\ipSymbol} = \kernelMatrix{\bandwidth}(\expertInducingPoints{})$ and $\bm{K}_{*\ipSymbol} = \kernelMatrix{\bandwidth}(\evalInputs,\expertInducingPoints{})$.

We choose $\bm{\mu}$ to be the constant mean function, i.e., $\bm{\mu}(X) = \mu\ones{n}$ for $X \in \inputSpace^n$, noting that other mean functions are possible. Note that test predictions $\fGP(x) = \bm{\mu}^*(x)$ are given by Eq.~\eqref{eq:predictiveMean}.

\myParagraph{The training objective}
Let $P$ denote the prior distribution of the inducing function values $\expertInducingValues{}$ of the \inducingPoints $\expertInducingPoints{}$ and let $Q$ denote a tractable \emph{variational distribution} intended to approximate $P(\cdot) \approx Q(\cdot|\trainingInputs{n},\trainingLabels{n})$.
In \emph{variational inference}, we want to minimize the \emph{Kullback-Leibler divergence} $\kullbackLeiblerDistance{Q}{P}$ between $Q$ and $P$, which is equivalent to maximizing the data log-evidence $\log(P(\trainingInputs{n},\trainingLabels{n}))$.
As a tractable approximation, we maximize the \emph{evidence lower bound} (ELBO), given by
\begin{align}
\nonumber
&\sideset{}{_{u\sim Q(\cdot|\trainingInputs{n},\trainingLabels{n})}}\E \log(P(\trainingInputs{n},\trainingLabels{n}|u)) - \kullbackLeiblerDistance{Q(\cdot|\trainingInputs{n},\trainingLabels{n})}{P}\\
 \label{eq:elbo}
\approx &\frac{1}{n}\mySum{b=1}{n} P_b - \kullbackLeiblerDistance{Q(\cdot|\trainingInputs{n},\trainingLabels{n})}{P},
\end{align}
where $P_{b}$ is the predictive log-likelihood in $x_b$, marginalized over the variational distribution $Q$, that is,
\begin{align}
\label{eq:predLogLik}
P_{b} :=
\sideset{}{_{u\sim Q(\cdot|\trainingInputs{n},\trainingLabels{n})}}\E \log\mathop{\mathlarger{\int}}P(y_b|f)P(f|u,x_b)df.
\end{align}

\subsubsection{Sparse mixture of experts}
\label{subsubsec:sparseMoE}
Given a finite set of expert models $\widehat{y}_l$ that are parameterized by $\theta_{e_l}$, the \MoE model is given by
\begin{align}
 \label{eq:fMoE}
 \fMoE(x) = \mySum{l=1}{L} G(x)_l\widehat{y}_l(x),
\end{align}
where the \emph{gate} $\fctn{G}{\inputSpace}{[0,1]^L}$ is a probability assignment of an input $x$ to the experts. In particular, it holds $\mySum{l=1}{L} G(x)_l \equiv 1$ and $G(x)_i \geq 0, \forall x\in\inputSpace$ and $1\leq i\leq L$.

We implement the approach of \cite{shazeer2017outrageously} to model the gate $G$ as follows:
For the \emph{softmax} function
\begin{align}
\label{eq:softmax}
\softmax(\bm{a})_i := \exp\{\bm{a}_i\} \Big/ \mySum{l=1}{L} \exp\{\bm{a}_l\},
\end{align}
where $\bm{a} \in \reals{L}$,
\cite{shazeer2017outrageously} propose to set
\begin{align}
\label{eq:gateSparse}
G(x) = \softmax(\widetilde{h}_1(x), \ldots, \widetilde{h}_L(x)),\quad\text{where}\quad\quad\widetilde{h}_{i_{l}}(x) = \begin{cases} h_{i_{l}}(x) &, l < \sparsity \\-\infty &, l\geq\sparsity+1 \end{cases}
\end{align}
for an adequate permutation $(i_1,\ldots,i_L)$ of $\mySet{1}{L}$ such that $h_{i_{l}}(x) > h_{i_{l+1}}(x)$ are ordered decreasingly.
Here, $h_l(x) = g_l(x) + \mathcal{N}(0, \gateNoise_l^2)$ is a noisy version of single-channel gating models $g_l$ with parameters $\theta_{g_l}$. Note that these models can be chosen freely and may also deviate from the choice of expert models $\widehat{y}_l$.

The cutoff value $1 \leq \sparsity \leq L$ controls the sparsity of the \MoE, as it enforces the minor mixture weights to strictly equal zero.
For stability reasons, during the training, we give each expert a chance to become an element of the top-\sparsity components by adding independent Gaussian noise $\mathcal{N}(0, \gateNoise_l^2)$ before thresholding, where $\gateNoise\in\positiveReal{L}$ is another hyperparameter to set or learn.
This noisy gating prevents a premature discarding of initially underperforming experts.

The overall \MoE hyperparameter set is thus given by
\begin{align}
 \label{eq:MoEHyperpars}
 \thetaMoE = (\{\theta_{e_l}\}_{l=1}^L, \{\theta_{g_l}\}_{l=1}^L, \sparsity, \gateNoise).
\end{align}

\section{Estimating locally optimal bandwidths via mixture of Gaussian processes}
\label{sec:theoryIsotropicOptimalSampling}
In this section, we derive our main contribution, namely the \GPR-based \AL framework, which we summarized in \figref{fig:ALframework}.
We first derive our \GPR-based estimates of \LFC and the \superiorTrainingDensity in \secref{subsec:GPRlfcEstimate} (\figref{fig:ALframework},~B).
Combining this estimate with the \AL framework from \cite{panknin2021optimal}, we obtain a \GPR-based, model-agnostic \superiorSamplingScheme in \secref{subsec:ALprocedure} (\figref{fig:ALframework},~C).
Next, we describe our \GPR-based \MoE model in
\secref{subsec:modelArchitecture} (\figref{fig:ALframework},~A)
of which we obtain the required \LOB estimate of \GPR for the estimation of \superiorTrainingDensity. The scalability of this estimate enables the application of our \superiorSamplingScheme to problems of high input space dimensions.
We then give details on the training of the \MoE in \secref{subsec:modelTraining} and finally propose an \LFC-based \inducingPoint selection method in \secref{subsec:inducingPointInit}.
We summarize the pseudo-code of our \superiorSamplingScheme in \algoref{alg:superiorSamplingScheme}.

\fig{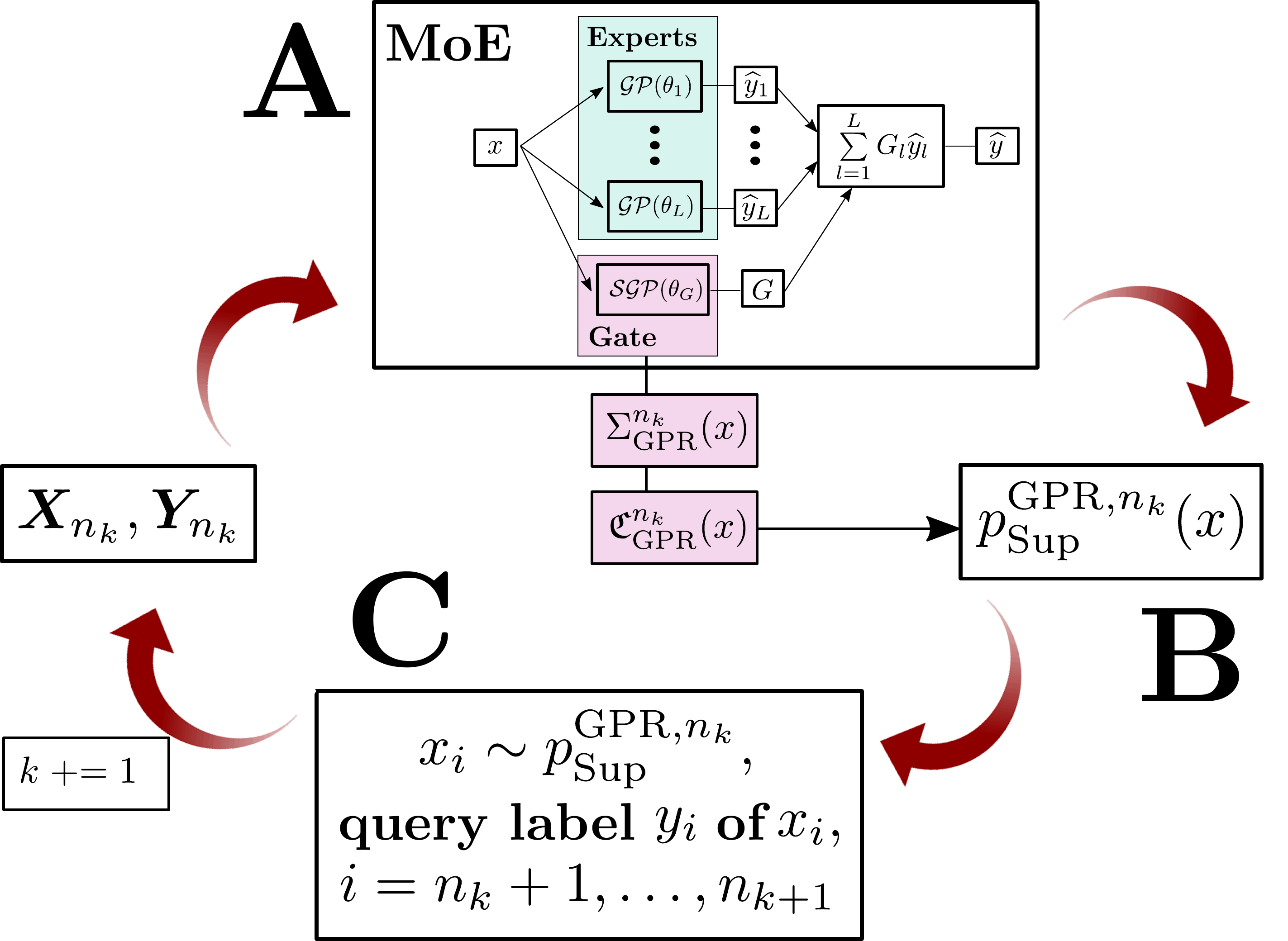}{0.79}{The proposed \AL framework.}{0.1cm}{0.0cm}
\begin{algorithm}[tb]
  \caption{Superior training data process $(\trainingInputs{n}, \trainingLabels{n})_{n\in\N}$ with labels $\trainingLabels{n}$ of training inputs $\trainingInputs{n} \overset{d}{\rightarrow} \pSup{\GPR}{n}$}
  \label{alg:superiorSamplingScheme}
  \renewcommand\thealgorithm{}
  \color{black}
  \relsize{-1}
\vspace{1.5mm}
  \begin{algorithmic}[1]
  \Algphase{Input}
  \State Intermediate training sizes $(n_k)_{k\in\Nzero}$ with $n_k < n_{k+1}, \forall k\in\Nzero$ for reestimation
  \State A labeled validation set $\bm{X}_\text{val}, \bm{Y}_\text{val}$
  \State An input generating process $\poolInputs\in\inputSpace^{N}$ with $\poolInputs \sim \poolDensity$
  \State The label oracle $\fctn{\bm{y}}{\inputSpace}{\R}$
  \State (optional) The test density $q$ 
  \State (optional) The intrinsic dimension $\intrinsicDimension \leq d$ of the input space $\inputSpace$
  \State (optional) The regularity $\alpha$ of the target function $f \in \diffableFunctions{\inputSpace,\R}{\alpha}$
  \Algphase{Output}
  \State (Infinite) training data process $(\trainingInputs{n}, \trainingLabels{n})$ with labels $\trainingLabels{n}$ of training inputs $\trainingInputs{n} \sim \pSup{\GPR}{n}$
  \Algphase{Procedure}
  \LineComment{Initialization}
  \State Estimate pool density $\approxPoolDensity{}$ based on $\poolInputs$ \Comment{e.g., using kernel density estimation}
  \If{$q$ is not specified}
  \State Set $q \leftarrow \approxPoolDensity{}$
  \EndIf
  \If{$\intrinsicDimension$ is not specified}
  \State Estimate $\intrinsicDimension$ based on \poolInputs \Comment{e.g., following the work of \cite{facco2017estimating}}
  \EndIf
  \If{$\alpha$ is not specified}
  \State Set $\alpha \leftarrow \infty$ \Comment{as discussed in \secref{sec:discussion}}
  \EndIf
  \State Set $p_0 \leftarrow q$
  \State Draw initial training inputs $\trainingInputs{n_0} \sim p_0$
  \State Query labels $\trainingLabels{n_0} \leftarrow \bm{y}(\trainingInputs{n_0})$ from the oracle
  \State Set $(\Theta_H, \bandwidth_E) \leftarrow \text{hyper\_init}(\trainingInputs{n_0}, \trainingLabels{n_0},p_0, \bm{X}_\text{val}, \bm{Y}_\text{val})$ \Comment{see \algoref{alg:hyperparameterInit} in \appref{sec:algo}}
  \LineComment{Sample Process}
  \For{$k\in\Nzero$}
  \If{$k > 0$}
  \State Update \inducingPoint locations $\expertInducingPoints{E},\gateInducingPoints \in \Theta_H$, where  $\expertInducingPoints{E},\gateInducingPoints \sim \sqrt{p_k\cdot\approxFctnComplexityLPS{\GPR}{n_{k-1}^{}}}$ \Comment{see \eqref{eq:optimizedIPdistribution} in \secref{subsec:inducingPointInit}}
  \If{$k == 1$}
  \State Gradually decrease $m_E = \abs{\expertInducingPoints{E}}$ and $m_G = \abs{\gateInducingPoints}$ as long as the validation performance of $\fMoE$ does not degrade as discussed in \secref{subsubsec:hyperpars}
  \EndIf
  \EndIf
  \State Train the model \fMoE from \secref{subsec:modelArchitecture} with hyperparameters $\Theta_H$ on $(\trainingInputs{n_k}, \trainingLabels{n_k})$ as described in \secref{subsec:modelTraining} 
  \State Estimate the \LOB $\approxLOBfunctionOfLPS{\GPR}{n_k}$ of \GPR according to \eqref{eq:gprLOB}
  \State Estimate the \LFC $\approxFctnComplexityLPS{\GPR}{n_k^{}} \leftarrow \left[1\big/p_k^{}(x)\right]^{\frac{1}{2\alpha+\inputDimension}}\abs{\approxLOBfunctionOfLPS{\GPR}{n_k^{}}(x)}^{-1}$ according to \eqref{eq:approxLFC}
  \State Estimate the superior training density $\approxPSup{\GPR}{n_k^{}} \leftarrow \left[\approxFctnComplexityLPS{\GPR}{n_k^{}}(x) q(x)\right]^{\frac{2\alpha+\inputDimension}{4\alpha+\inputDimension}}\gpNoiseVariance(x)^{\frac{2\alpha}{4\alpha+\inputDimension}}$ according to \eqref{eq:approxPOpt}
  \State Set $\gamma_1^{} = \max_{x\in\inputSpace}\left. p_k(x) \middle/ \approxPSup{\GPR}{n_k^{}}(x) \right.$ and $\gamma_2^{} = \max\left\{0, \left.(0.5-\gamma_1^{-1})\middle/(1-\gamma_1^{-1})\right.\right\}$ \Comment{see \secref{subsec:ALprocedure}}
  \State Set $p_{k+1} \leftarrow \gamma_2^{} p_k + (1 - \gamma_2^{}) \approxPSup{\GPR}{n_k^{}}$
  \State Set $\widetilde{p}_{k+1} \leftarrow 2 p_{k+1} - p_k$ \Comment{see \secref{subsec:ALprocedure}}
  \State Draw $X_{n_k+1},\ldots,X_{n_{k+1}} \sim \widetilde{p}_{k+1}$ via importance sampling from \poolInputs as described in \secref{subsec:ALprocedure}
  \State Query labels $y_{n_k+1},\ldots,y_{n_{k+1}} \leftarrow \bm{y}(X_{n_k+1},\ldots,X_{n_{k+1}})$ from the oracle
  \State Set $\trainingInputs{n_{k+1}} \leftarrow \trainingInputs{n_{k}} \cup \{X_{n_k+1},\ldots,X_{n_{k+1}}\}$ and $\trainingLabels{n_{k+1}} \leftarrow \trainingLabels{n_{k}} \cup \{y_{n_k+1},\ldots,y_{n_{k+1}}\}$ \Comment{Then $\trainingInputs{n_{k+1}} \sim p_{k+1}$}
  \EndFor
  \end{algorithmic}
\end{algorithm}

\subsection{\GPR-based \LFC and the \superiorTrainingDensity}
\label{subsec:GPRlfcEstimate}

Let $\LOBfunctionOfLPS{\GPR}{n}(x)$ denote the \LOB function \eqref{eq:LOBDefinition} of \GPR.
Inspired by the results to \LFC and the \superiorTrainingDensity of \LPS in Eq.~\eqref{eq:isotropicComplexityLPS}~and~\eqref{eq:optimalSamplingFiniteLPS}, we are able to deduce their \GPR-based analog.
Here, we need to take into account that \GPR adapts universally\footnote{That is, the \MISE decays at the minimax-rate $n^{-\frac{2\alpha}{2\alpha+\inputDimension}}$ of nonparametric models.} to functions $f \in \diffableFunctions{\inputSpace,\R}{\alpha}$, as opposed to \LPS, whose decay rate is determined by the specified polynomial order $\LPSorder$.
The idea of the \LFC estimate was to adjust \LOB appropriately so that it becomes invariant under the influence of the training density, heteroscedasticity, and its global decay with respect to the training size $n$.

Combining the local effective sample size $p(x)n$ with the scaling result of the global $\LOBfunctionOfLPS{\GPR}{n}$ in \eqref{eq:GPRbandwidthSamplesizeProportionality} from \secref{subsec:GPRbandwidthScaling}, we propose an \LFC estimate for \GPR as follows (see \appref{sec:proofLFCGPR} for proof details).
\begin{restatable}[\LFC of \GPR]{mythm}{isotropicComplexityForGPR}
\label{thm:isotropicComplexityForGPR}
For $f \in \diffableFunctions{\inputSpace,\R}{\alpha}$, $\trainingInputs{n} \sim p$ and homoscedastic noise, the \GPR-based \LFC estimate of $f$ in $x\in\inputSpace$ is asymptotically given by
 \begin{align}
 \label{eq:isotropicComplexityGPR}
 \fctnComplexityLPS{\GPR}{n}(x) := \left[\frac{1}{p(x)n}\right]^{\frac{\inputDimension}{2\alpha+\inputDimension}}\abs{\LOBfunctionOfLPS{\GPR}{n}(x)}^{-1}.
\end{align}
\end{restatable}
In analogy to Eq.~\eqref{eq:isotropicComplexityLPS}, $\fctnComplexityLPS{\GPR}{n}$ measures the structural complexity of $f$, as it asymptotically does not depend on $p$, $v$ and $n$.
Note that the \LPS model provides no explicit way to adapt to the local noise variance $v(x)$, such that the \LOB of \LPS scales with respect to $v$ to address heteroscedasticity (see \eqref{eq:lpsLOBDefinition} in \appref{sec:optimalSamplingSupplement}).
For \GPR, we have made the restriction of homoscedastic noise in the definition of $\fctnComplexityLPS{\GPR}{n}$ in \thmref{thm:isotropicComplexityForGPR}, since we are not aware of a theory on the scaling of \GPR-based \LOB with respect to heteroscedasticity. However, as opposed to \LPS, a heteroscedastic \GPR model provides an explicit way to adapt to the local noise variance $v(x)$ via regularization.
As a result, we observe only very little influence of heteroscedasticity on \LOB function, which we will demonstrate in \secref{subsec:doppler}.
Thus, $\fctnComplexityLPS{\GPR}{n}$ will be sufficiently calibrated in a heteroscedastic scenario, making it a reasonable estimate of \LFC in practice without further restrictions.



Now, when putting $\fctnComplexityLPS{\GPR}{n}$ into Eq.~\eqref{eq:optimalSamplingFiniteLPS} with $\LPSorder = \alpha - 1$, we obtain the \superiorTrainingDensity
\begin{align}
 \label{eq:GeneralizedOptimalSampling}
 \pSup{\GPR}{n}(x) \propto \textstyle \left[\fctnComplexityLPS{\GPR}{n}(x) q(x)\right]^{\frac{2\alpha+\inputDimension}{4\alpha+\inputDimension}}v(x)^{\frac{2\alpha}{4\alpha+\inputDimension}}(1 + o(1)).
\end{align}
For even $\alpha\in\N$ with $\LPSorder = \alpha-1$ and $\fctnComplexityLPS{\GPR}{n} \equiv \fctnComplexityLPS{\LPSorder}{n}$, $\pSup{\GPR}{n}$ and $\pOptLPS{\LPSorder}{n}$ coincide, which proved to be optimal for \LPS. In this sense, \eqref{eq:GeneralizedOptimalSampling} generalizes \eqref{eq:optimalSamplingFiniteLPS} to the general case of $\alpha\in\nonnegativeReal{}$, where we expect that the true \optimalTrainingDensity for $f \in \diffableFunctions{\inputSpace}{\alpha}$ will not deviate by a lot from $\pSup{\GPR}{n}$.
Since \LPS and \GPR are related models, we furthermore expect $\fctnComplexityLPS{\GPR}{n}$ to be similar to $\fctnComplexityLPS{\LPSorder}{n}$ for the appropriate order $\LPSorder$.

Note that for $f \in \diffableFunctions{\inputSpace}{\infty}$, we let 
$\alpha \rightarrow \infty$ in Eq.~\eqref{eq:isotropicComplexityGPR}~and~\eqref{eq:GeneralizedOptimalSampling} to obtain
\begin{align}
\label{eq:GeneralizedOptimalSamplingInfAlpha}
\textstyle \fctnComplexityLPS{\GPR}{n}(x) = \abs{\LOBfunctionOfLPS{\GPR}{n}(x)}^{-1}
\quad\text{and}\quad\quad
\pSup{\GPR}{n}(x) \propto \left[\fctnComplexityLPS{\GPR}{n}(x) q(x)v(x)\right]^{\frac{1}{2}}.
\end{align}

While $\trainingInputs{n}\sim \pSup{\GPR}{n}$ will not be optimal for our model, we expect it to be asymptotically superior to the naive \randomTestSampling, i.e., $\trainingInputs{n}\sim q$, due to the model-agnosticity of the \LPS-based result.
To assess the asymptotic performance of a training density $p$ (such as $\pSup{\GPR}{n}$), let us first observe the following:
\vspace*{-1mm}
\par\noindent For regression problems and under weak assumptions the law of the \MISE does not change with respect to $p$, except for a constant multiple \citep{gyorfi2002distribution,willett2005faster}.
Accordingly, the number of actively selected training samples ($\sim p$) that are required to achieve the same level of accuracy of \randomTestSampling is given by a constant $\relSampleSizeSymbol > 0$.
Formally, we can define $\relSampleSizeSymbol$ as follows.
\begin{mydef}
\label{def:relSampleSize}
Over the space of square-integrable functions $f\in\intableFunctions{\inputSpace}{2}$, for a nonparametric regression model $\widehat{f}$ and a training density $p$,
we define by $\relSampleSize{\widehat{f}}{p} > 0$ the relative required sample size such that for $n' = \relSampleSize{\widehat{f}}{p} n$, $\bm{X}_{n'}' \sim p$ and $\trainingInputs{n} \sim q$ it holds that
\[\textstyle\MISE\left(q,\widehat{f}| \trainingInputs{n}\right) = \MISE\left(q,\widehat{f}| \bm{X}_{n'}'\right) (1 + o(1)).\]
\end{mydef}
Thus, a training density $p$ is asymptotically superior to \randomTestSampling, if $\relSampleSize{\widehat{f}}{p} < 1$, since we achieve the same performance as \randomTestSampling with only a fraction of the number of training samples.
In \secref{sec:experiments} we will demonstrate the superiority of the training density $\pSup{\GPR}{n}$ for our \GPR-based \MoE model.

\myParagraph{Respecting the intrinsic dimension in high-dimensional input spaces}
In Eq.~\eqref{eq:isotropicComplexityGPR}, \eqref{eq:GeneralizedOptimalSampling} and \eqref{eq:GeneralizedOptimalSamplingInfAlpha} we assume the input space \inputSpace to have full degrees of freedom \inputDimension, which in practice is particularly not the case in high-dimensional feature spaces.
For an intrinsic dimension $\intrinsicDimension < \inputDimension$ of \inputSpace, we adjust as follows:
For the space $\SigmaSpace = \condset{\sigma\bandwidth}{\sigma > 0}$ of bandwidth candidates that are essentially isotropic up to a fixed, shared positive definite factor $\bandwidth \in \posDefSet{d}$ that is, e.g., calculated in a pre-processing step, let
$\LOBfunctionOfLPS{\GPR}{n}(x) = \lobfunctionOfLPS{\GPR}{n}(x)\bandwidth$ be the \LOB function of \GPR with respect to $\SigmaSpace$.
Then we replace all occurrences of \inputDimension for $\intrinsicDimension$ and $\abs{\LOBfunctionOfLPS{\GPR}{n}(x)}$ for $\lobfunctionOfLPS{\GPR}{n}(x)^{\intrinsicDimension}$ in Eq.~\eqref{eq:isotropicComplexityGPR}, \eqref{eq:GeneralizedOptimalSampling} and \eqref{eq:GeneralizedOptimalSamplingInfAlpha}.

Besides being an ingredient to \AL, \LFC can also be used to reduce the required model complexity. For example, in an RBF-network or a sparse \GPR model, we can coarsen or refine the model resolution by placing an adequate amount of basis functions or \inducingPoints, respecting \LFC.
We will discuss this choice in \secref{subsec:inducingPointInit} and demonstrate its ability to reduce the overall model complexity in \secref{subsec:doppler}.
Finally, \LFC can be inspected to obtain deeper insights into the research field of the regression problem, which is particularly hard for high-dimensional data (see \secref{subsec:MDMalon}).

\subsection{The active learning framework}
\label{subsec:ALprocedure}
Starting with an initial training set $\trainingInputs{n_0}, \trainingLabels{n_0}$ of size $n_0$ with $\trainingInputs{n_0} \sim p_0^{}$ for some initial training distribution such as $p_0^{} \equiv q$, we implement the online sampling procedure as described in \cite{panknin2021optimal}, such that $\trainingInputs{n}\sim \pSup{\GPR}{n}$ as $n\rightarrow\infty$.
We grow the training set as follows:

Given the current training set $\trainingInputs{n_k}, \trainingLabels{n_k}$ we estimate $\approxLOBfunctionOfLPS{\GPR}{n_{k}}$ as described in \secref{subsec:modelArchitecture}.
Using \eqref{eq:isotropicComplexityGPR}, \eqref{eq:GeneralizedOptimalSampling}, it is
\begin{align}
\label{eq:approxLFC}
&\textstyle\approxFctnComplexityLPS{\GPR}{n_k^{}}(x) \propto \left[1\big/p_k^{}(x)\right]^{\frac{1}{2\alpha+\inputDimension}}\abs{\approxLOBfunctionOfLPS{\GPR}{n_k^{}}(x)}^{-1},\;\;\text{and}\\
\label{eq:approxPOpt}
&\textstyle\approxPSup{\GPR}{n_k^{}}(x) \propto \left[\approxFctnComplexityLPS{\GPR}{n_k^{}}(x) q(x)\right]^{\frac{2\alpha+\inputDimension}{4\alpha+\inputDimension}} \gpNoiseVariance(x)^{\frac{2\alpha}{4\alpha+\inputDimension}}.
\end{align}

Letting the next sample size be $n_{k+1}^{} = 2n_k^{}$, we have already drawn half the samples of $n_{k+1}^{}$ according to a potentially different distribution $p_k^{}$ than the new proposed $\approxPSup{\GPR}{n_k^{}}$. The closest we can get in distribution to $\approxPSup{\GPR}{n_k^{}}$ is given by
$\displaystyle\trainingInputs{n_{k+1}} \!\!\!\sim p_{k+1}$, where $p_{k+1} := \gamma_2^{} p_k+(1-\gamma_2^{})\approxPSup{\GPR}{n_k^{}}$, for $\displaystyle\gamma_2^{} = \max\left\{0, \frac{0.5-\gamma_1^{-1}}{1-\gamma_1^{-1}}\right\} \in [0,0.5)$ and $\displaystyle\gamma_1^{} = \max_{x\in\inputSpace} \frac{p_k(x)}{\approxPSup{\GPR}{n_k^{}}(x)}$.
This is achieved by sampling $x_{n_k+1},\ldots,x_{n_{k+1}} \sim \widetilde{p}_{k+1}$ for $\widetilde{p}_{k+1} = 2p_{k+1} - p_k$, which is a valid probability density \citep{panknin2021optimal}.

\myParagraph{Adaptions in the pool-based active learning scenario}
In the \AL framework described above, we deal with properly normalized probability densities. But in the \emph{pool-based} \AL scenario such normalization is usually impossible since our information about the input space \inputSpace is restricted to a large, unlabeled \emph{pool} of samples $\poolInputs\in\inputSpace^{N}$. This pool follows a distribution $\poolInputs \sim \poolDensity$, for which it is common to assume an (unnormalized) density estimate $\approxPoolDensity{}$ to be given: Unlabeled inputs are considered cheaply accessible, whereas querying labels is expensive.

For our \AL framework to be applicable, it suffices to keep all considered densities such as $\approxPSup{\GPR}{n_k^{}}$ at equal norm, which we can enforce via normalizing a density $p$ by $\bar p = p/\text{norm}(p)$, where
\[
\text{norm}(p) = \abs{\poolInputs}^{-1}\mySum{x\in \poolInputs}{} p(x) / \approxPoolDensity{}(x).
\]
To see this, note that first of all $\approxPoolDensity{}$ is an unnormalized estimate of $\poolDensity$ such that we can write $\approxPoolDensity{} \approx c \cdot \poolDensity$ for some unknown constant $c>0$.
On the one hand, it is $\displaystyle\int_\inputSpace \approxPoolDensity{}(x) dx = c$ by definition. On the other hand, it is
\[
\text{norm}(p) \approx \int_\inputSpace \frac{p(x)}{\approxPoolDensity{}(x)} \poolDensity(x) dx = \frac{1}{c}\int_\inputSpace p(x) dx,
\]
such that also $\displaystyle \int_\inputSpace \bar p(x) dx = \frac{1}{\text{norm}(p)}\int_\inputSpace p(x) dx \approx c$
holds for any unnormalized density $p$.

Subsequently, the required samples $x_{n_k+1},\ldots,x_{n_{k+1}} \sim \widetilde{p}_{k+1}$ are obtained via \emph{importance sampling} from the pool with \emph{importance weights} $\displaystyle \Prob(x_{i} = x) \propto \widetilde{p}_{k+1}(x) / [\text{norm}(\widetilde{p}_{k+1}) \approxPoolDensity{}(x)]$ for $x \in \poolInputs$ and $n_k+1 \leq i \leq n_{k+1}$.

\subsection{Sparse mixture of Gaussian processes}
\label{subsec:modelArchitecture}
Recall the sparse \MoE model \eqref{eq:fMoE} from \secref{subsubsec:sparseMoE}, given by
\begin{align*}
 \fMoE(x) = \mySum{l=1}{L} G(x)_l\widehat{y}_l(x),
\end{align*}
where the gate $\fctn{G}{\inputSpace}{[0,1]^L}$ is a probability assignment of an input $x$ to the expert models $\widehat{y}_l$. In particular, it holds $\mySum{l=1}{L} G(x)_l \equiv 1$ and $G(x)_i \geq 0, \forall x\in\inputSpace$ and $1\leq i\leq L$.
According to \eqref{eq:MoEHyperpars}, besides the expert and gate model parameters $\{\theta_{e_l}\}_{l=1}^L$ and $\{\theta_{g_l}\}_{l=1}^L$, this \MoE approach has two hyperparameters, \sparsity and \gateNoise, for controlling the sparsity of the gate and adding noise to the gate responses during the training to escape local optima.

We choose the expert models as well as the single channel gating models to be sparse variational \GaussianProcesses \citep{williams1996gaussian,hensman2015scalable}, that is, $\widehat{y}_l \sim \SVGP{\theta_{e_l}}$ and $g_l \sim \SVGP{\theta_{g_l}}$, which are parameterized by $\theta_{e_l}$ and $\theta_{g_l}$, as described in \secref{subsubsec:sparseVarGP}.
The overall \MoE hyperparameter set is thus given by
$\thetaMoE = (\{\theta_{e_l}\}_{l=1}^L, \{\theta_{g_l}\}_{l=1}^L, \sparsity, \gateNoise)$.

We will keep certain hyperparameters of $\thetaMoE$ constant after initialization, and share some hyperparameters across experts and the channels of the gate: 
While the covariances of the inducing value distributions $\expertInducingValueCov{} \in \theta_{}, \theta_{} \in \thetaMoE$ could be full positive definite matrices, we apply $\expertInducingValueCov{} = 0$ throughout, giving favorable stability and computational efficiency.
For the same reasons, we fix the \emph{inducing point} (IP) locations $\expertInducingPoints{}\in \theta_{}, \theta_{} \in \thetaMoE$ after initialization. Furthermore, we share the \inducingPoint locations among the experts, respectively the gate channels, such that for $\expertInducingPoints{} \in \theta_{e_l}$ we apply $\expertInducingPoints{} = \expertInducingPoints{E}$ and for $\expertInducingPoints{} \in \theta_{g_l}$ we apply $\expertInducingPoints{} = \gateInducingPoints$, for all $1 \leq l \leq L$.

In this work, our goal is to fit a single, coherent regression problem by a \MoE approach. Therefore, we propose to share all the parameters across the experts that characterize the regression function rather than the expert model. That is, we share the mean $\mu_E$, the regularization parameter $\lambda_E$, and the noise variance function $\gpNoiseVariance$, respectively the global noise variance $\sigma_\varepsilon^2$ with $\gpNoiseVariance(x) \equiv \sigma_\varepsilon^2$ in case of homoscedasticity.
Furthermore, we apply a fixed, logarithmically spaced set of individual expert bandwidth scaling factors $\sigma_1 < \ldots < \sigma_L$ that are multiplied by a fixed, shared bandwidth matrix $\bandwidth_E$. Our expert parameters therefore reduce to
\[\theta_{e_l} = (\mu_E, \lambda_E, \gpNoiseVariance, \sigma_l\bandwidth_E, \expertInducingPoints{E}, \expertInducingValueMean{e_l}, 0).\]

\begin{myRemark}
 Recall from \secref{sec:relatedWork} that it is possible to replace the variational \GPR expert models for full as well as sparse analytic \GPR formulations (see~\appref{sec:analyticGPR}). With slight abuse of notation, these cases are subsumed by setting $\expertInducingValueMean{e_l} = \emptyset$ or $\expertInducingPoints{E} = \expertInducingValueMean{e_l} = \emptyset$ for sparse, respectively full analytic \GPR.
\end{myRemark}

Since our objective does not incorporate any likelihood about the gate's output, there is no noise function to fit for the gate, such that we set $\gpNoiseVariance \equiv 0$ for $\gpNoiseVariance\in\theta_{g_l}$ and all $1\leq l\leq L$.
Each output channel of the gate poses its own classification problem, which is why we do not share the means. Yet, we share the regularization parameter and the bandwidth, as the individual channels should be structurally similar.
Our gate parameters therefore reduce to
\[\theta_{g_l} = (\mu_{g_l}, \lambda_G, 0, \sigma_G\idMatrix{\inputDimension}, \gateInducingPoints, \expertInducingValueMean{g_l}, 0).\]
After training as described in \secref{subsec:modelTraining}, this \MoE can cope with a varying structural complexity through the individual bandwidth scaling factors $\sigma_l$ of the experts and heteroscedastic noise through the adaptive regularization.
Additionally, we can now use the gate of our \MoE to propose an \LOB estimate of \GPR.

\myParagraph{A \GPR-based \LOB estimate} After training of the \MoE, we use the learned gate $G$ from \eqref{eq:fMoE} to predict $\LOBfunctionOfLPS{\GPR}{n}(x)$ as
\begin{align}
 \label{eq:gprLOB}
 \approxLOBfunctionOfLPS{\GPR}{n}(x) = \approxlobfunctionOfLPS{\GPR}{n}(x)\bandwidth_E,\quad\text{where}\quad\quad
 \approxlobfunctionOfLPS{\GPR}{n}(x) = \exp\left\{\mySum{l=1}{L}G(x)_l\log(\sigma_l)\right\}.
\end{align}
Due to the finite candidate set $\sigma_1,\ldots,\sigma_L$ we are limited to measure a quantization of $\LOBfunctionOfLPS{\GPR}{n}(x)$ through $G(x)_l = \Prob(\LOBfunctionOfLPS{\GPR}{n}(x) = \sigma_l\bandwidth_E)$.
If, in fact, $\LOBfunctionOfLPS{\GPR}{n} \in \mySet{\sigma_1\bandwidth_E}{\sigma_L\bandwidth_E}$ holds true, then there exists an index function $j(x) \in \mySet{1}{L}$ such that $\LOBfunctionOfLPS{\GPR}{n}(x) = \sigma_{j(x)}\bandwidth_E$.
In this case, we are able to exactly recover \LOB with $G(x)_l = \begin{cases} 1, &l = j(x) \\ 0, &\text{else}\end{cases}$.
In any other case, the estimate \eqref{eq:gprLOB} of \LOB is a reasonable interpolation, which deviation from $\LOBfunctionOfLPS{\GPR}{n}$ can be controlled by the number of bandwidth candidates of the \MoE.

\subsection{Model training}
\label{subsec:modelTraining}
This section is devoted to the training of the model described in \secref{subsec:modelArchitecture}.
We first set up the training objective in \secref{subsubsec:objective} and describe the training procedure of our model in \secref{subsubsec:trainingSetup}, where we identify hyperparameters of the approach.
Then, we discuss how to choose the essential hyperparameters systematically on the initial training dataset in \secref{subsubsec:hyperpars}.

\subsubsection{The training objective}
\label{subsubsec:objective}
First, we will set up the objective function for training our \MoE model in \emph{batch mode}.

\myParagraph{The main objective}
Denote by $\varnothing \subsetneq \mathcal{B} \subseteq \mySet{1}{n}$ the indices of a batch, and let $w_{\mathcal{B}} = \mySum{b\in\mathcal{B}}{}w(x_b)$ for the training importance weight function $w \propto q/p$.
Let $P_l$ be the prior distribution of the inducing function values of the l-th expert and $Q_l$ the corresponding variational distribution as defined in \secref{subsubsec:sparseVarGP}.
We choose the (through the gate G) weighted sum of the individual expert negative ELBO objectives \eqref{eq:elbo}, denoted by
\[\textstyle \MLL(\trainingInputs{n}, \trainingLabels{n}, \mathcal{B}, w, \thetaMoE) = -\mySum{l=1}{L}\left[ w_{\mathcal{B}}^{-1}\mySum{b\in\mathcal{B}}{}v_l(x_b) P_{b,l} - \frac{1}{n_l}\kullbackLeiblerDistance{Q_l(\cdot|\trainingInputs{n},\trainingLabels{n})}{P_l}\right],\]
as our main objective, where 
$n_l = n w_{\mathcal{B}} / \nu_{\mathcal{B},l}$ for $\nu_{\mathcal{B},l} = \mySum{b\in\mathcal{B}}{}v_l(x_b)$ with $v_l(x) = G(x)_l w(x)$ and
\[P_{b,l} = \sideset{}{_{u\sim Q_l(\cdot|\trainingInputs{n},\trainingLabels{n})}}\E\log \mathop{\mathlarger{\int}}P_l(y_b|f)P_l(f|u,x_b)df\]
is the predictive log-likelihood \eqref{eq:predLogLik} of the l-th expert in $x_b$, marginalized over its variational distribution $Q_l$.

\myParagraph{A penalty on small bandwidth choices}
In the spirit of Lepski's method \citep{lepski1991problem,lepski1997optimal}, we prefer the largest choice of bandwidth out of all candidates that perform comparably well.
In order to enforce this, we penalize smaller bandwidth choices by adding the following term:
\begin{align}
 \label{eq:smallBWpenalty}
 \text{pen}_\sigma(\trainingInputs{n}, \trainingLabels{n},\mathcal{B}, w, \thetaMoE) = \frac{2}{(L-1)}\mySum{l=1}{L}\nu_{\mathcal{B},l} (L-l) \big/ \mySum{l=1}{L}\nu_{\mathcal{B},l}.
\end{align}
Note that $\text{pen}_\sigma(\trainingInputs{n}, \trainingLabels{n},\mathcal{B}, w, \thetaMoE) = 1$ if $\nu_{\mathcal{B},1} = \ldots = \nu_{\mathcal{B},L}$. Our total objective then amounts to
\begin{align}
 \label{eq:fullObjective}
 \textstyle \text{Obj}(\trainingInputs{n}, \trainingLabels{n}, \mathcal{B}, w, \thetaMoE&) = \MLL(\trainingInputs{n}, \trainingLabels{n}, \mathcal{B}, w, \thetaMoE)
 + \vartheta_\sigma\text{pen}_\sigma(\trainingInputs{n}, \trainingLabels{n},\mathcal{B}, w, \thetaMoE).
\end{align}

\begin{myRemark}
If we assume our problem to be (almost) noise-free, we replace the \MLL in our objective \eqref{eq:fullObjective} for the \emph{mean squared error}
\[\MSE(\trainingInputs{n}, \trainingLabels{n}, \mathcal{B}, w, \thetaMoE) = w_{\mathcal{B}}^{-1}\mySum{b\in\mathcal{B}}{}w(x_b) \pnorm{y_b - \widehat{y}(x_b)}{}^2.\]
\end{myRemark}

\subsubsection{Training procedure}
\label{subsubsec:trainingSetup}
We implement our model in \emph{PyTorch} \citep{paszke2019pytorch}, using the \emph{GPyTorch}-package \citep{gardner2018gpytorch}.
Given the training set $(\trainingInputs{n}, \trainingLabels{n})$, we minimize the objective described in \secref{subsubsec:objective} via ADAM-optimization \citep{kingma2015ADAM}.
It remains to identify those variables of the \MoE that will be adapted as parameters during the training. Then, the remaining variables are hyperparameters that need to be specified or tuned through an external validation step.

Recall from \secref{subsec:modelArchitecture} that the \MoE has two further hyperparameters, $\sparsity$ for enforcing sparse gate responses and a noise term on the gate responses during the training, which is controlled by $\gateNoise$.
Instead of learning $\gateNoise$ as a parameter during the training---like proposed by \cite{shazeer2017outrageously}---we propose to shrink $\gateNoise \leftarrow \gateNoise \eta_{\gateNoise}$ after each training epoch, for a multiplicative factor $\eta_{\gateNoise} < 1$ and an initial value $\gateNoise := \gateNoise_0$ as hyperparameters.
We discuss this heuristic choice in \appref{sec:designChoices}.

We require appropriate learning rates for the optimization of the parameters and tunable hyperparameters of the model. Generally, we suggest applying an adaptive base learning rate $\eta$, where we shrink $\eta \leftarrow \eta_i := \frac{1}{2}\eta_{i-1}$ for an initial base learning rate $\eta_0$ during the training as soon as the validation performance gets stuck until $\eta_i$ crosses a lower threshold, e.g., $\eta_i < \eta_0/1000$.
Note that, relative to the base learning rate, good learning rates for the individual types of tunable parameters should be deployed: Within a \GaussianProcess component $\SVGP{\theta}$ with $\theta = (\mu, \lambda, \gpNoiseVariance, \bandwidth, \expertInducingPoints{}, \expertInducingValueMean{}, \expertInducingValueCov{})$, the hyperparameters $(\mu, \lambda, \gpNoiseVariance)$ must be updated on a smaller scale than the inducing value distribution, given by $(\expertInducingValueMean{}, \expertInducingValueCov{})$.
In this regard, let $\eta_H \leq 1$ be the factor such that, if we update $\expertInducingValueMean{}$ at rate $\eta$, then we update $(\mu, \lambda, \gpNoiseVariance)$ at rate $\eta_H\eta$.

Similarly, we need to update the gate parameters $\theta_{g_l}$ on a smaller scale than the expert parameters $\theta_{e_l}$.
In this regard, let $\eta_G \leq 1$ be the factor such that, if we update $\theta_{e_l}$ at rate $\eta$, then we update $\theta_{g_l}$ at rate $\eta_G\eta$.

The set of hyperparameters that require off-training selection (e.g., via cross-validation) is thus given by 
\begin{align}
 \label{eq:hyperPars}
 \Theta_H = (B, \sparsity, \{\sigma_l\}_{l=1}^L, \sigma_G, \lambda_G, \expertInducingPoints{E}, \gateInducingPoints, \gateNoise_0, \eta_{\gateNoise}, \vartheta_\sigma, \eta_0, \eta_H, \eta_G),
\end{align}
whereas the overall set of parameters that get tuned while training is given by
\begin{align}
 \label{eq:MoEpars}
 \Theta_T = (\mu_E, \lambda_E, \gpNoiseVariance, \bandwidth_E, \expertInducingValueMean{E}, \mu_G, \gateInducingValueMean).
\end{align}
We provide further details on the design choices for our \MoE model in \appref{sec:designChoices}.

\subsubsection{Choosing the hyperparameters}
\label{subsubsec:hyperpars}
Since our \MoE approach is based on known building blocks \citep{williams1996gaussian,hensman2015scalable,shazeer2017outrageously} we can train our model using well-established software libraries \citep{kingma2015ADAM,paszke2019pytorch,gardner2018gpytorch}, with the hyperparameters chosen by following best practice.
While the set of hyperparameters \eqref{eq:hyperPars} appears to be large, most of them can be tuned in advance on the initial training dataset of moderate size and held fixed in the subsequent training data refinement process.

Note that some hyperparameters impact the computational complexity rather than the model performance. Thus, as long as they are not underestimated, their tuning is optional and will therefore be postponed:
\vspace*{-2mm}
\begin{itemize}
    \item Since our \MoE is robust concerning unnecessarily large choices of the gate output sparsity \sparsity, we initialize $\sparsity \equiv L$ while choosing the remaining hyperparameters, followed by tuning $\sparsity$ as the last hyperparameter, where we successively reduce \sparsity until we observe a significant loss of performance of the \MoE.
    \item The numbers $m_E = \abs{\expertInducingPoints{E}}$, $m_G = \abs{\gateInducingPoints}$ of \inducingPoints of the expert and the gate are the main driver of the computational complexity of our \MoE. While unnecessarily large numbers will not hurt the model performance, they should therefore be set to the smallest value that leads to no significant loss of performance to keep the computational complexity of the model moderate at larger training sizes.
    In the initial iteration, we use $m_E=n_0$ for the experts and $m_G = \frac{n_0}{4}$, where the locations of the \inducingPoints $\expertInducingPoints{E}, \gateInducingPoints \sim p$ are chosen \emph{diverse} as described in \appref{sec:diverseIPs}. In the second iteration, where we have first estimates of \LFC, $\expertInducingPoints{E}$ and $\gateInducingPoints$ are drawn as discussed in \secref{subsec:inducingPointInit}.
    Here, we gradually decrease $m_E$ and $m_G$ until we observe a significant loss of validation performance of the \MoE. We hold $m_E$ and $m_G$ fixed in subsequent iterations.
    Note that the \inducingPoint locations are not subject to optimization.
\end{itemize}
The initial base learning rate and the expert's internal hyperparameters learning rate ($\eta_0, \eta_H$) and the batch size $B$ that are related to a single \GPR expert rather than the whole \MoE model:
\vspace*{-2mm}
\begin{itemize}
\item First, we hold $\eta_H = 0.2$, $B = n_0$ fixed at reasonable initial values and perform line search over $\eta_0$ according to the resulting validation performance of a single, isotropic, sparse \GPR expert.
Here, too small values of $\eta_0$ result in slow convergence of the objective, in which case we interrupt the training immediately and increase $\eta_0$ as long as the first objective updates are consistently decreasing.
\item Next, we choose $\eta_H$ according to the resulting validation performance of a single, isotropic, sparse \GPR expert, where we gradually decrease $\eta_H$, starting from $\eta_H = 1$.
Again, we interrupt the training for too large choices of $\eta_H$, where the training objective will diverge.
\item Finally, we choose $B$ according to the resulting validation performance of a single, isotropic, sparse \GPR expert, where we gradually decrease $B$, starting from $B = n_0$.
\end{itemize}
Next, we have to deal with the remaining hyperparameters that are related to the \MoE model. For this, we first initialize the less crucial hyperparameters at reasonable values, tuning them afterward:
We apply a set of experts with $\{\sigma_l\}_{l=1}^L$, where $L=7$, and $\sigma_l = 2^\frac{l-4}{\intrinsicDimension}$, which are logarithmically spaced around the global bandwidth estimate $\bandwidth_E$ of the best-performing model that we obtained from the above tuning of the hyperparameters related to a single \GPR expert.
The noise added to the gate ($\gateNoise_0, \eta_{\gateNoise}$) as well as the regularization $\vartheta_\sigma$ of the bandwidth function are about fine-tuning of the model. We set them to $\gateNoise_0 = 0.1, \eta_{\gateNoise} = 1/\sqrt{2}$ and $\vartheta_\sigma = 0.01$. We suggest to keep $\eta_{\gateNoise} = 1/\sqrt{2}$ throughout without further tuning.
\vspace*{-2mm}
\begin{itemize}
\item Via grid search, we choose $\sigma_G, \lambda_G$ according to the resulting validation performance of the \MoE model, where we gradually decrease the gate learning rate from $\eta_G = 1$.
\end{itemize}
As a last step, we choose the hyperparameters for fine-tuning of the \MoE model:
\vspace*{-2mm}
\begin{itemize}
    \item First, we perform line search over $\vartheta_\sigma$ according to the resulting validation performance of the \MoE.
    \item Next, we tune $\{\sigma_l\}_{l=1}^L$: We observe that unreasonable bandwidths will be automatically dropped during the training. Therefore, if the minimal or maximal candidate associated with $\sigma_1, \sigma_L$ is not chosen during the training, we remove the respective expert and retrain the \MoE. Vice versa, we expand the bandwidth candidate range beyond $\sigma_1, \sigma_L$ with a factor of $2^{\mp\frac{1}{\intrinsicDimension}}$ as long as the boundary candidates are not dropped during the training.
    \item Finally, we perform line search over $\gateNoise_0$ according to the resulting validation performance of the \MoE.
\end{itemize}

\subsection{Initializing the \inducingPoint locations}
\label{subsec:inducingPointInit}
Since the numbers of \inducingPoints of the gate $\gateInducingPoints$, as well as the experts $\expertInducingPoints{E}$ are the computational bottleneck of our model, they should be chosen advisedly. 
We can interpret the choice of \inducingPoints as a nested \AL task at small sample size.
In the small sample size regime, input space geometric arguments have proven to be robust and superior in comparison to naive approaches like random sub-sampling from the training inputs \citep{teytaud2007active,yu2010passive,wu2019pool,liu2021pool}. They are representative, respecting the training distribution, and diverse (with \emph{low-dispersion}) so that they achieve an acceptable representation of the dataset at the smallest possible number of \inducingPoints. By low-dispersion, we resort to the following definition:
\begin{mydef}
 \label{def:dispersion}
 The \emph{dispersion}, given by $\sup_{x\in\inputSpace}\min_{1\leq i\leq n} \pnorm{x-x_i}{}$ \citep{niederreiter1988low} is a measure of how well spread out the training sample is. We say that a sequence has low-dispersion if its dispersion is lower than the dispersion of random uniform sampling.
\end{mydef}
Indeed, by sampling the \inducingPoints in this manner, we can reduce the distance of an evaluation point $x$ to its closest neighbor in $\expertInducingPoints{E}$, which is known to reduce the reconstruction error of a full kernel matrix by a sparse representation \citep{zhang2008improved}.

In addition, recall that our derived \LFC measure of local structural complexity quantifies the local variation of the target function. Intuitively, we require more \inducingPoints to sense and reconstruct the target function where this local variation is higher.
In summary, we therefore propose to choose the \inducingPoints
\begin{align}
\label{eq:optimizedIPdistribution}
\expertInducingPoints{E},\gateInducingPoints \sim \sqrt{p\cdot\fctnComplexityLPS{\GPR}{n}}    
\end{align}
as the \emph{geometric mean} of \LFC and the training density $\trainingInputs{n}\sim p$ in a diverse way.
Here, we ensure diversity by implementing distribution preserving clustering or particle repulsion as described in \appref{sec:diverseIPs}.

\section{Experiments}
\label{sec:experiments}
In this section, we will first analyze our approach on toy-data, regarding the \MoE model, \LFC, and the \superiorTrainingDensity.
Then, we apply our approach to a high-dimensional MD simulation dataset from quantum chemistry, by which we can deduce deeper insights into this regression problem. 

We denote
the \emph{root mean squared error} (\RMSE) and the \emph{maximum absolute error} (\maxAE) of a model $\widehat{f}$ for a test set $\trueTestInputs\in\inputSpace^{N}$ with $\trueTestInputs\sim q$ by
\begin{align*}
\RMSE(\widehat{f},\trainingInputs{n},\trainingLabels{n}) = \left[\frac{1}{N}\!\!\sum\limits_{x\in\trueTestInputs}\!\!\abs{f(x)-\widehat{f}_{\trainingInputs{n}, \trainingLabels{n}}(x)}^2\right]^\frac{1}{2}
\;\text{and}\quad
\maxAE(\widehat{f},\trainingInputs{n},\trainingLabels{n}) = \max\limits_{x\in\trueTestInputs}\abs{f(x)-\widehat{f}_{\trainingInputs{n}, \trainingLabels{n}}(x)}.
\end{align*}

As already discussed in \secref{subsec:GPRlfcEstimate}, the learning rate is invariant under change of the training density $\trainingInputs{n}\sim p$ in the considered scenario. For our \MoE model and $f \in \diffableFunctions{\inputSpace,\R}{\alpha}$ we, thus, can write
\begin{align}
 \label{eq:RMSEasymptotics}
 \RMSE(\fMoE,\trainingInputs{n},\trainingLabels{n}) = C_p n^{-\tau}(1 + o(1))\;\;\text{, with}\quad\tau = \begin{cases} \frac{\alpha}{2\alpha+d}, &\alpha < \infty \\ 1/2, &\alpha = \infty
 \end{cases},
\end{align}
where $C_p > 0$ is a constant depending on the training density $p$.
Note that we can theoretically bound the asymptotic \RMSE from below by $C^* n^{-\tau}$, where we have defined $C^* := C_{p^*}$ with $\optTrainingInputs{n} \sim p^*$ being the optimal training set from \eqref{eq:ALtask}. Unfortunately, since $p^*$ is unknown---even when given the ground truth---we are not able to estimate $C^*$ and, thus, provide a lower bound of the \RMSE beyond the known learning rate $n^{-\tau}$.

As an \AL performance measure, we use the relative required sample size from \defref{def:relSampleSize} which can be estimated for a \GPR-based model such as our \MoE and $f \in \diffableFunctions{\inputSpace,\R}{\alpha}$ according to
\begin{align}
 \label{eq:relSampleSizeGPRApprox}
 \textstyle \relSampleSize{\fMoE}{p} \approx \left[\frac{\RMSE(\fMoE,\bm{X}_{n}',\bm{Y}_{n}')}{\RMSE(\fMoE,\trainingInputs{n},\trainingLabels{n})}\right]^\frac{1}{\tau}
\end{align}
where it is $\bm{X}_{n}' \sim p$ and $\trainingInputs{n} \sim q$ with respective labels $\bm{Y}_{n}'$ and $\trainingLabels{n}$.
Using \eqref{eq:relSampleSizeGPRApprox}, we can compare the asymptotic \AL performance of different \AL sampling schemes in the following experiments.
For example, we can quantify the \AL performance of our proposed \AL framework by sampling $\bm{X}_{n}' \sim \approxPSup{\GPR}{n}$.

\subsection{Doppler function}
\label{subsec:doppler}

We will first demonstrate our approach on the \emph{Doppler} function (see, for example, \cite{donoho1994ideal}), which was also discussed in related work that deals with inhomogeneous complexity \citep{panknin2021optimal,bull2013spatially}.
For $x \in \inputSpace = [0,1]$, let
\begin{align*}
&\Prob(y | x) = \Gauss{y}{f(x)}{1},
\quad f(x) = C\sqrt{x(1-x)}\sin\left(2\pi(1 + \epsilon)\middle/(x + \epsilon)\right),
\end{align*}
where $\epsilon = 0.05$, $C$ is chosen such that $\pnorm{f}{2} = 7$ and $\Gauss{\cdot}{\mu}{\sigma^2}$ denotes the Gaussian distribution with mean $\mu$ and variance $\sigma^2$. We assume a uniform test distribution $q \sim \uniformDist{\inputSpace}$ in all Doppler function experiments.

This one-dimensional, homoscedastic toy-example allows for an easy and intuitive visualization.
\figref{fig:dopplerExperiment_IPdistribution_test_vs_opt} shows an example dataset as blue dots and the true function $f$ to infer in black.
Due to the strong variation of structural complexity, a single-scale \GPR model
does not cope well with the Doppler function (see \appref{subsec:dopplerGlobalGPR} for a comparison of single-scale to multi-scale \GPR).

We implement our proposed \MoE model as described in \secref{subsec:modelArchitecture} with sparse \GaussianProcesses as the expert and gate models and using the Gaussian kernel $k$. We apply $512$, respectively $128$ \inducingPoints for the experts and the gate, which are chosen via SVGD (see \appref{sec:diverseIPs}).
Furthermore we apply $\sigma_j = 10^{(j-10)/3}, 1\leq j\leq 7$, as the expert bandwidths, $\lambda_E = 20$ as the initial expert regularization, and $\sigma_G = 0.05$ and $\lambda_G = 10$ for the gate. For the training, we apply a batch size of $B = 512$, a terminal expert sparsity $\sparsity = 2$, a penalty factor of $\vartheta_\sigma = 0.5$ for small bandwidth choices, gate noise parameters $\gateNoise_0 = 0.1$ and $\eta_{\gateNoise} = 1/\sqrt{2}$, and learning rate parameters $\eta = 0.01$, $\eta_H = 0.2$, $\eta_G = 1$.

\figref{fig:dopplerExperiment_gate_LFC_LOB_pOpt} shows the gate function after training of the \MoE model, as described in \secref{subsubsec:objective}, and the associated estimates of \LOB, \LFC and the \superiorTrainingDensity, calculated according to \eqref{eq:gprLOB} and \eqref{eq:GeneralizedOptimalSamplingInfAlpha}.

\fig{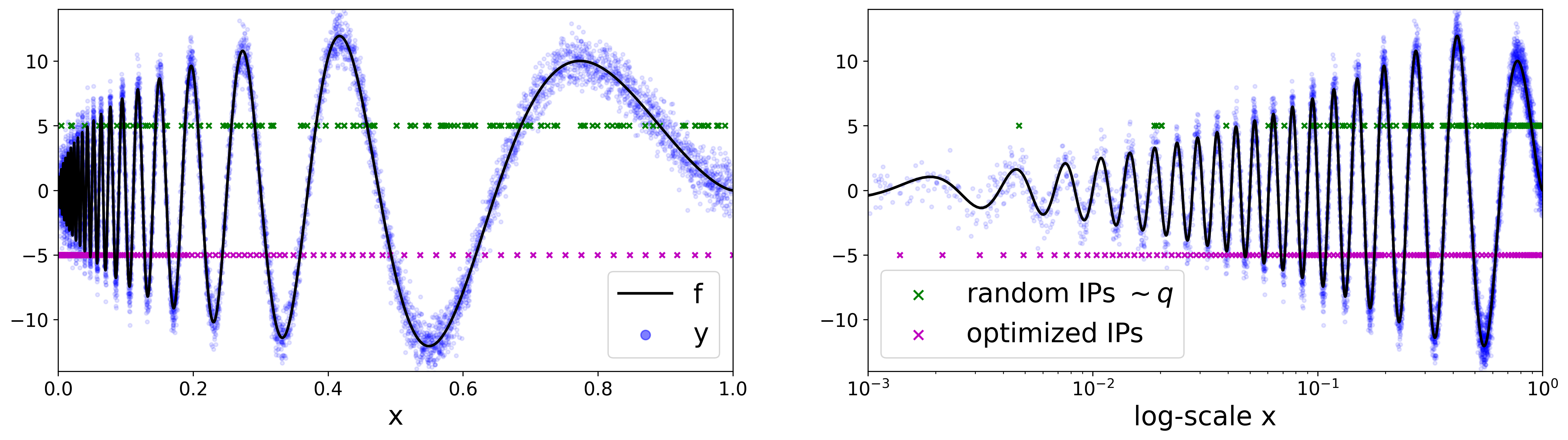}{1}{The Doppler experiment: An exemplary dataset and the locations of $128$ \inducingPoints, once sampled most naively---that is, random according to the test distribution---and once optimized regarding diversity as well as structural complexity and representativeness as described in \secref{subsec:inducingPointInit}, shown on natural x-scale (left) and on logarithmic x-scale (right).}{0.7cm}{0.0cm}

\fig{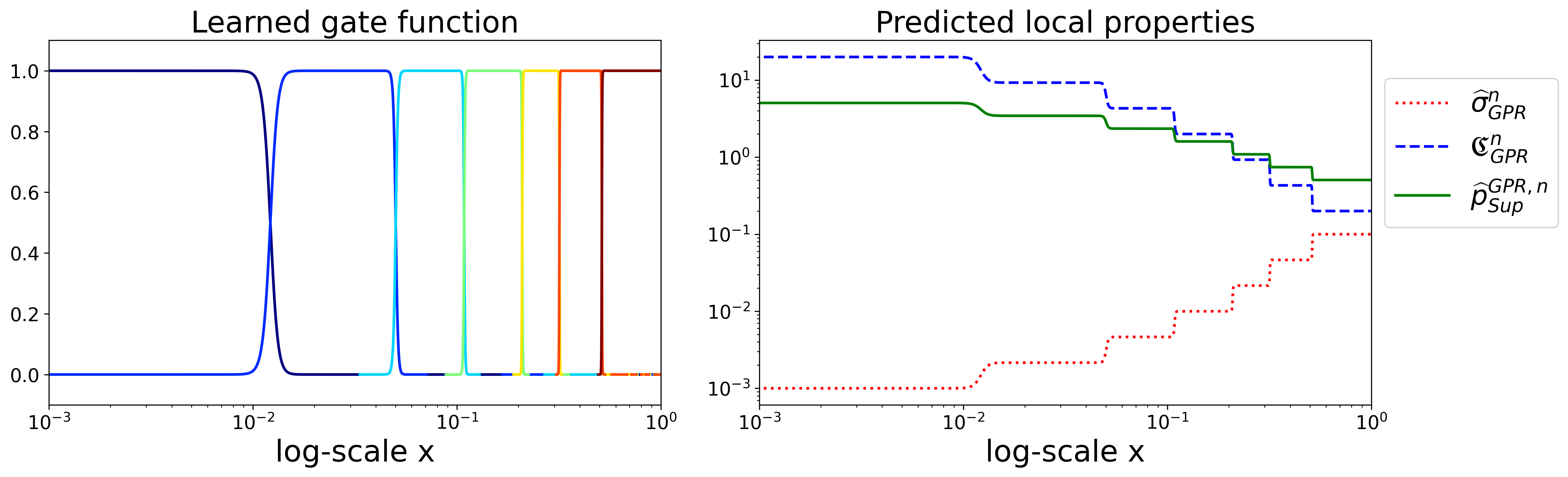}{1}{The Doppler experiment: The gate function (left) and the associated estimates of \LOB, \LFC and \superiorTrainingDensity (right) trained on the dataset from \figref{fig:dopplerExperiment_IPdistribution_test_vs_opt} and shown on a logarithmic x-scale.}{0.8cm}{0.0cm}


\myParagraph{Comparing the active learning framework in the \LPS and \GPR domain}
Since $f \in \diffableFunctions{\inputSpace,\R}{\infty}$, our deduced \superiorTrainingDensity estimate is given by Eq.~\eqref{eq:GeneralizedOptimalSamplingInfAlpha}.
In \figref{fig:dopplerExperiment_bandwidths_optDensities_LPSvsGPR} we plot our estimates of \LOB and the \superiorTrainingDensity in comparison to the \LPS-based results for polynomial degrees of order $Q =1, 3$, and with implementation and hyperparameters as described in \cite{panknin2021optimal}.
Here, we can observe the qualitative similarity of the \LPS- and \GPR-based estimates of \LOB.

\fig{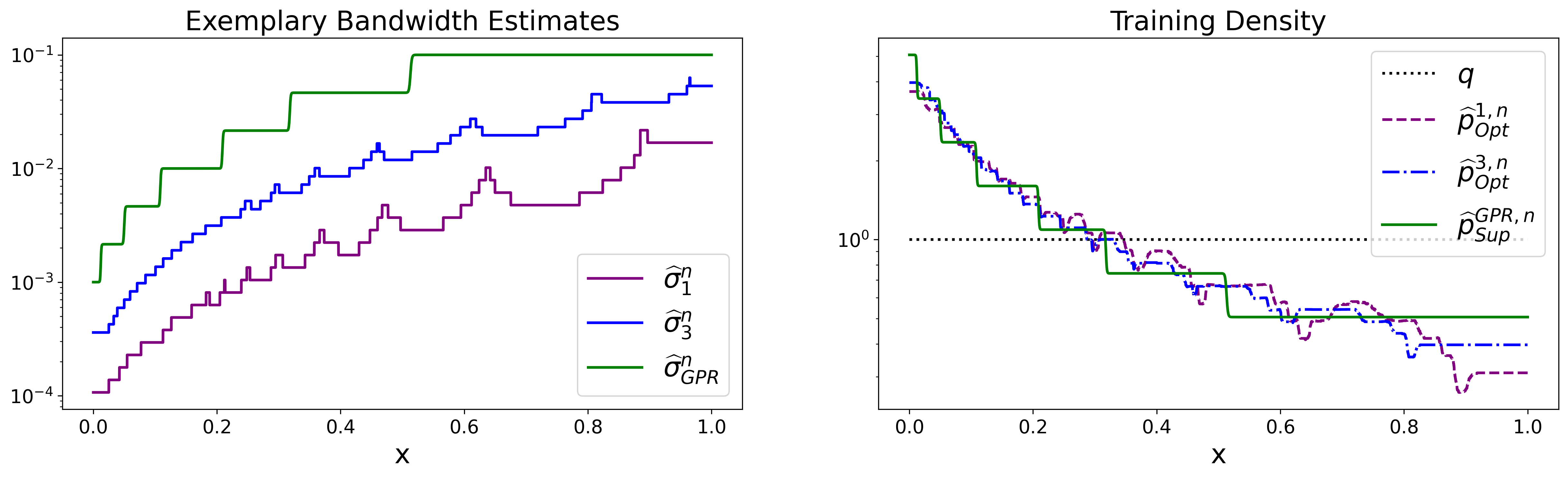}{1}{The Doppler experiment: The \LOB estimates (left) and the resulting \superiorTrainingDensity of our proposed \GPR-based approach in comparison to the \LPS-based approach of order $\LPSorder = 1, 3$ (right). The results are averaged over $20$ repetitions.}{0.8cm}{0.0cm}

\fig{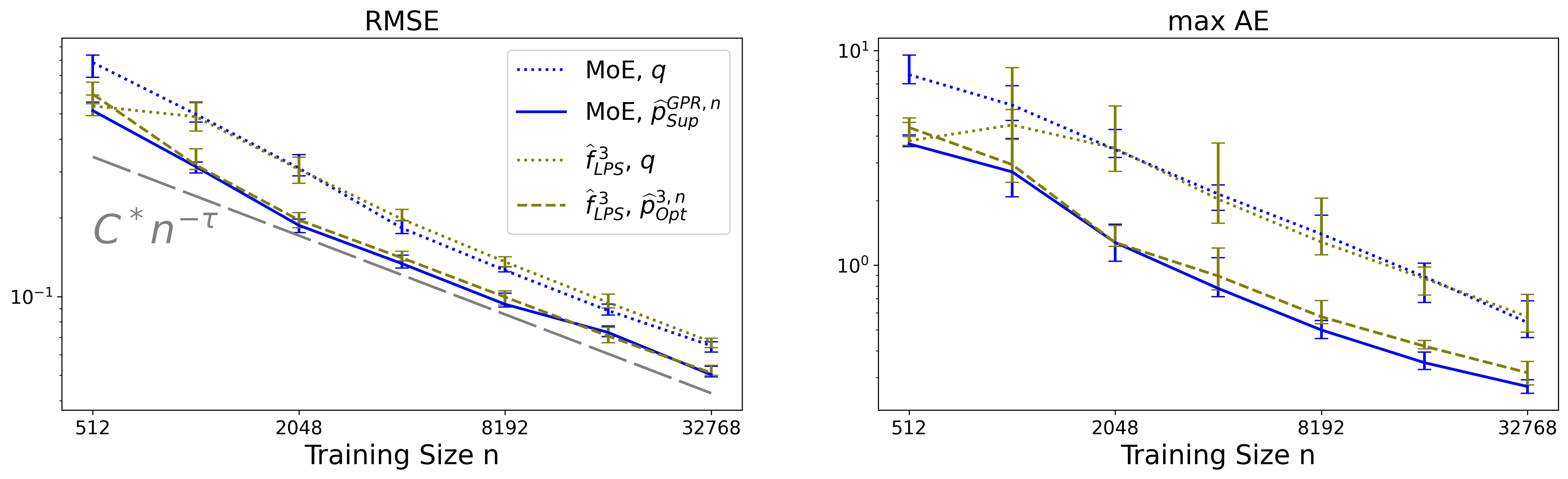}{1}{The Doppler experiment: The \RMSE (left) and the \maxAE (right) of our proposed \GPR-based approach in comparison to the \LPS-based approach of order $\LPSorder = 3$ (see Eq.~\eqref{eq:LPSoptimalPredictor}~and~\eqref{eq:optimalSamplingFiniteLPS}), once using the respective \AL scheme and once, applying \randomTestSampling. The results are averaged over $20$ repetitions. The long-dashed, gray line is for illustration of the optimal learning rate $\tau$ from \eqref{eq:RMSEasymptotics}, where the offset $C^*$ is imaginary. It shall therefore not be confused with a true lower bound.}{0.8cm}{0.0cm}
When conducting the proposed \GPR-based active sampling scheme as described in \secref{subsec:ALprocedure}, we additionally observe quantitative benefits in \figref{fig:dopplerExperiment_LPSOpt_Vs_GPROpt_alternative} over \randomTestSampling---quite similar to the \LPS-based result for $Q = 3$:
When estimating the relative sample size \eqref{eq:relSampleSizeGPRApprox} we require to achieve the same \RMSE via active sampling compared to \randomTestSampling, we obtain $\textstyle\relSampleSize{\fMoE}{\approxPSup{\GPR}{n}} = 0.58\pm0.04$. This means that we save about 42\% of samples via our active sampling scheme.

This provides evidence for the effectiveness of our \superiorSamplingScheme, combining the theoretical foundation of the \LPS domain with the efficient access to \LOB estimates in the \GPR domain.

\myParagraph{Comparing \randomTestSampling to \equidistantSampling}
In the introduction, we indicated that the advantage of the robust and model-agnostic input space geometric arguments
\citep{teytaud2007active,yu2010passive,wu2019pool,liu2021pool}
diminishes as the training size grows. We can substantiate this claim by comparing \randomTestSampling to \equidistantSampling on the Doppler dataset. By \equidistantSampling over $\inputSpace = [0,1]$ we mean the deterministic construction of the training inputs, where for $n$ given training samples the subsequent $n$ training inputs get placed halfway between all nearest neighbors of the former $n$ samples. In this way, the training input inter-distances are halved exactly with each iteration.
We regard this construction as the optimal input space geometric choice, which result will subsume all \AL competitors of this type.
We now observe in \figref{fig:dopplerExperiment_GPR_UnifVsEqui} that, indeed, \equidistantSampling is superior to \randomTestSampling at small training sizes. As claimed, however, with growing training size, this advantage gradually diminishes until it has vanished completely at $n=2^{15}$ training samples.

\fig{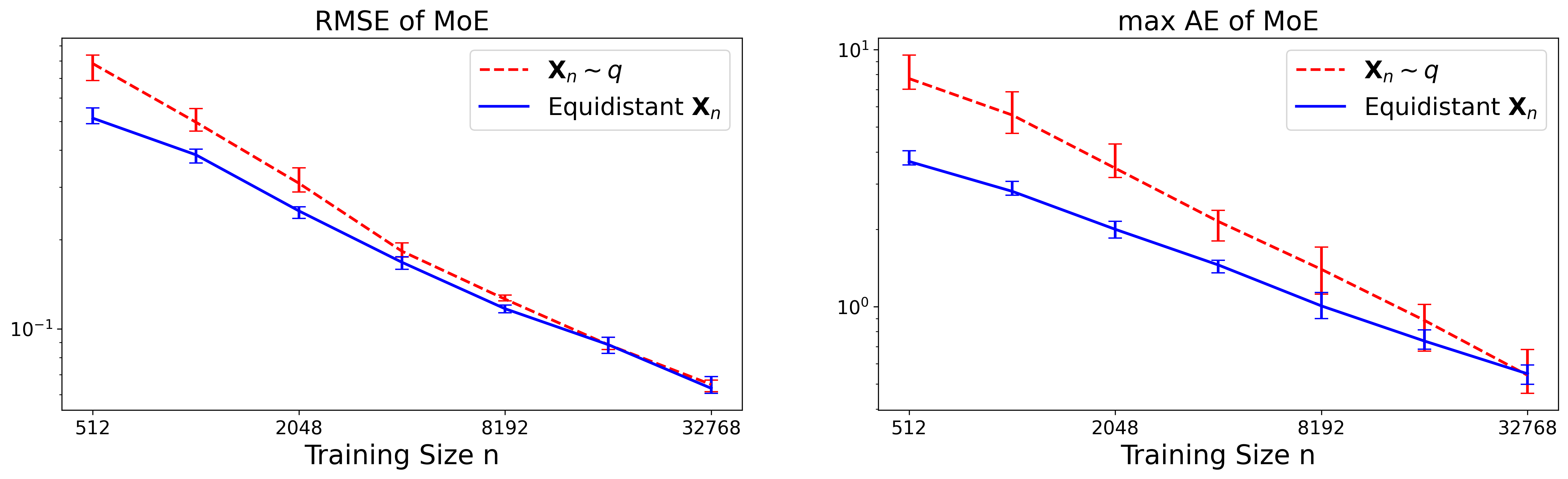}{1}{The Doppler experiment: The \RMSE (left) and the \maxAE (right) of our proposed \MoE model when comparing \randomTestSampling to \equidistantSampling. The results are averaged over $20$ repetitions.}{0.8cm}{0.0cm}

\myParagraph{On Gaussian process uncertainty}
In \secref{sec:relatedWork} we mentioned that \GPU is the most related approach to our \superiorSamplingScheme since both build on \GPR models. As also discussed therein, we can regard standard \GPU as an input space geometric argument, whose performance we can subsume by equidistant sampling in the Doppler experiment. Particularly this implies that standard \GPU provides no benefits regarding asymptotic \AL performance.

Instead---given the gate function of our \MoE from the previous part of this experiment, which was obtained for $2^{15}$ training samples and which we now keep fixed---we define the uncertainty estimate of our model \MoGPU as a straightforward extension of \GPU which takes the inhomogeneous complexity of data into account:
By simply weighting the predictive variances of all experts in some input $x$ with respect to the gate values $G(x)$ from \eqref{eq:fMoE}, we derive
\begin{align}
 \label{eq:MoGPU}
 \MoGPU(x) = \mySum{l=1}{L}G(x)_l \bm{C}_{\theta_l}^*(x),
\end{align}
where $\bm{C}_{\theta_l}^*$ is the predictive variance of the l-th expert (see~\eqref{eq:predictiveVar}).
Note that we consider \MoGPU as a baseline competitor to our \superiorSamplingScheme.

\fig{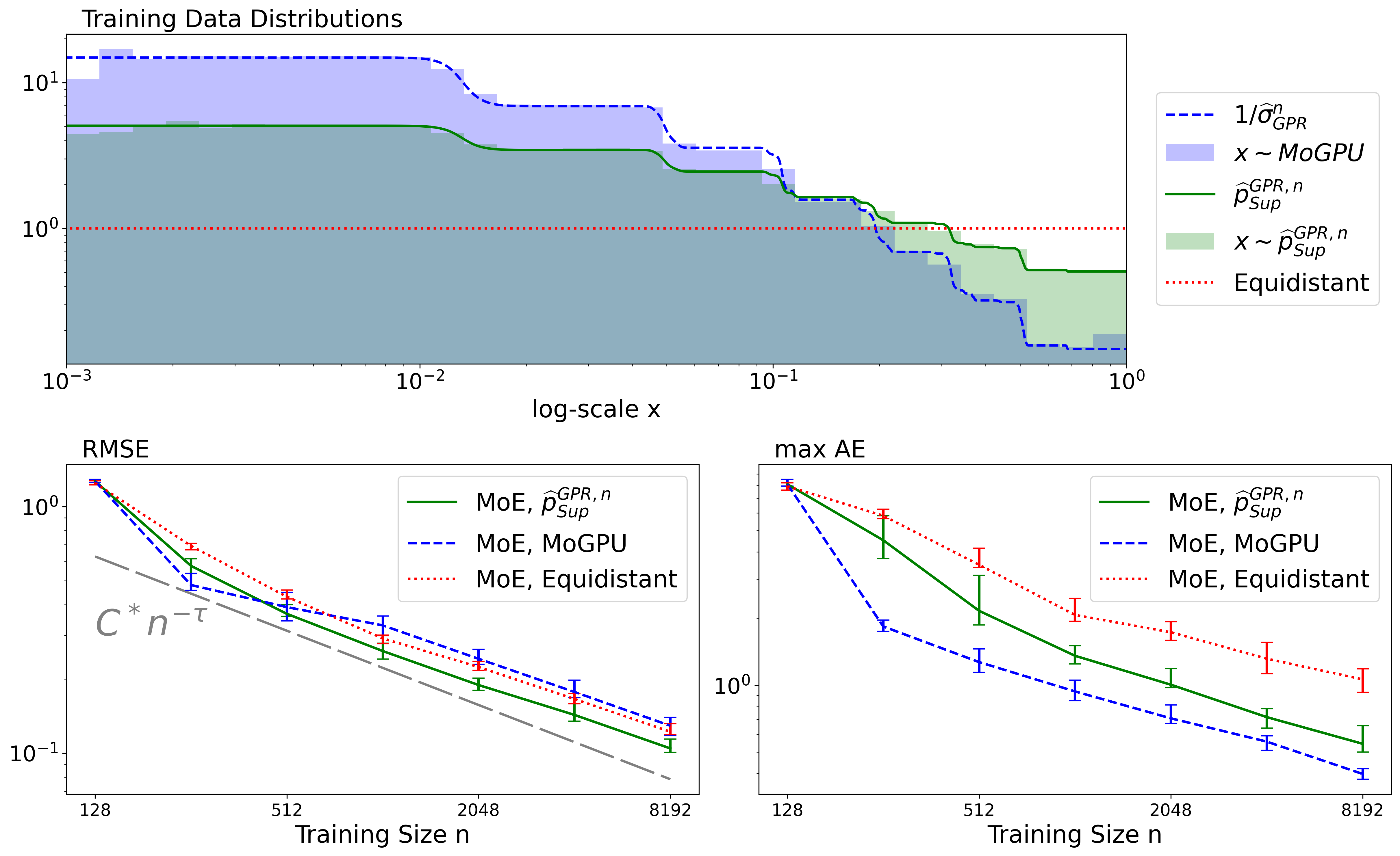}{1}{A comparison of the mixture of Gaussian process uncertainty and the \equidistantSampling baselines to our proposed active sampling scheme for the Doppler experiment: (Top) The training data histograms after $2^{13}$ samples, contrasted with functions of $\lobfunctionOfLPS{\GPR}{n}$, and the \RMSE (bottom left) and the \maxAE (bottom right) at several training sizes of the compared schemes. The results are averaged over $20$ repetitions. The long-dashed, gray line is for illustration of the optimal learning rate $\tau$ from \eqref{eq:RMSEasymptotics}, where the offset $C^*$ is imaginary. It shall therefore not be confused with a true lower bound.}{0.7cm}{0.0cm}

Intuitively, the uncertainty estimate in $x\in\inputSpace$ increases as the applied bandwidth $\lobfunctionOfLPS{\GPR}{n}(x)$ decreases. Now, in order to equalize uncertainty over the input space, \MoGPU will sample more in regions where $\lobfunctionOfLPS{\GPR}{n}$ is smaller.
For $\trainingInputs{n}$ drawn according to \MoGPU, we expect $\trainingInputs{n} \sim [\lobfunctionOfLPS{\GPR}{n}]^{-d}$. This expectation holds as can be seen at the top in \figref{fig:dopplerExperiment_SABER_EquiVsUncertaintyVsPOpt2}.

For evaluation, we combine the fixed gate function with full \GPR experts and compare our proposed sampling scheme with \MoGPU (both determined through the gate).
In all error measures the beneficial effect of the low-dispersion property of $\trainingInputs{n}$ drawn according to \MoGPU has already vanished for about $512$ training samples, from where the asymptotic law dominates. As expected, our approach is superior to \MoGPU when comparing \RMSE. Interestingly, \MoGPU is superior to our approach regarding the \maxAE, suggesting that $\trainingInputs{n} \sim [\lobfunctionOfLPS{\GPR}{n}]^{-d}$ is the preferable training distribution under the supremum-norm.

\myParagraph{\AL Performance on deep Gaussian processes}
To demonstrate the model-agnosticity of our \AL approach, we deploy the \DGP model of \cite{sauer2023active} using the CRAN package \emph{deepgp}\footnote{See \url{https://cran.r-project.org/web/packages/deepgp/index.html}}. This package also implements an aggregate variance-based \AL criterion \citep{cohn1994neural}, which they named \emph{active learning Cohn} (ALC) after the originator.
We deploy a 3-layer \DGP model, using the Gaussian kernel. For test evaluation, We train the model using \emph{Vecchia-approximation} \citep{sauer2023vecchia} with a total of 10,000 Gibbs-sampling steps, burning the initial 8000, and thinning the remaining steps to 1,000.

Beginning with 128 equidistant samples, we refine the training data of the \DGP model using the ALC criterion, \randomTestSampling, and our proposed superior training scheme. The resulting training data distributions and the performance of the \DGP model are plotted in \figref{fig:dopplerExperiment_DGP_EquiVsUncertaintyVsPOpt}.
As expected, our superior training scheme performs superior to \randomTestSampling. While the ALC criterion that is particularly designed for the \DGP model performs best, we observe only very little difference at 512 training samples.
Note that sampling according to the ALC criterion becomes computationally challenging already at this point since the \DGP model has to be re-trained after each new sample. In contrast, sampling $\trainingInputs{n}\sim\pSup{\GPR}{n}$ can be performed in batch mode. This result emphasizes the complementary nature of our asymptotic work to the classic \emph{bottom-up} \AL literature.

\fig{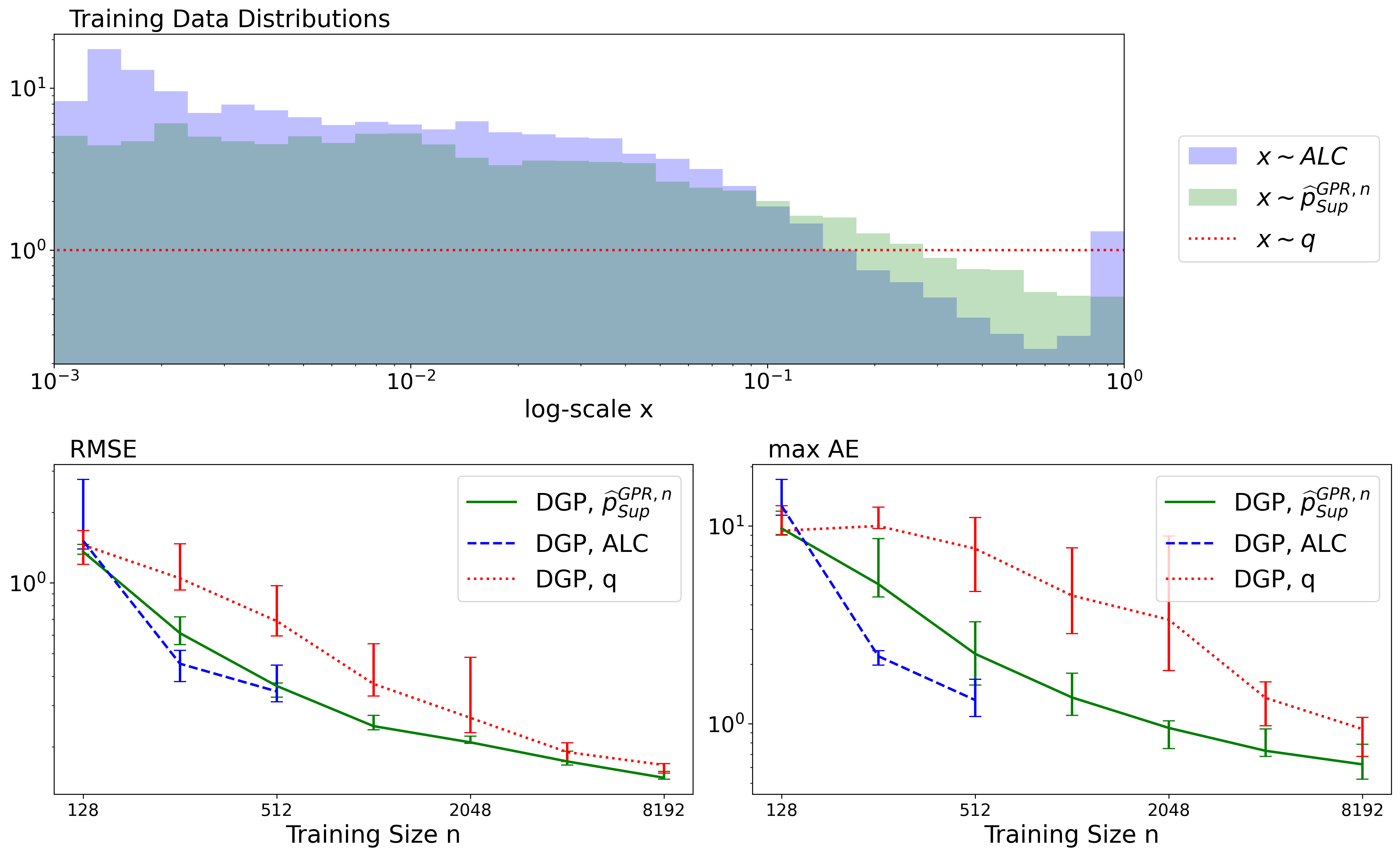}{1}{A comparison of the \DGP model performance, using the ALC criterion for \DGP, the \randomTestSampling baseline, and our proposed active sampling scheme for the Doppler experiment: (Top) The training data histograms after $512$ samples, and the \RMSE (bottom left) and the \maxAE (bottom right) at several training sizes of the compared schemes. The results are averaged over $5$ repetitions.}{0.6cm}{0.0cm}

\myParagraph{Necessity of the small bandwidth penalty}
We impose a penalty on small bandwidth choices through the factor $\vartheta_\sigma = 0.5$ to regularize the bandwidth function and prevent overfitting, as described in \appref{sec:designChoices}.
We demonstrate this overfitting issue in \appref{subsec:dopplerUnregularizedCase} that results from applying no regularization ($\vartheta_\sigma = 0$).

\myParagraph{Parsimonious modeling using \LFC}
In \secref{subsec:GPRlfcEstimate} we mentioned that \LFC can also be used to coarsen or refine the model resolution adequately to reduce the overall complexity of the model.
While we fixed the \inducingPoints to reasonable numbers in the other parts of the Doppler experiment, that is, $m=512$ and $m=1024$ \inducingPoints under active, respectively \randomTestSampling, we here investigate the influence of the number of \inducingPoints and their distribution on the capability to resemble the Doppler function.
Recall from \secref{subsec:inducingPointInit} that we interpret the choice of the \inducingPoints as a nested \AL task at small sample size $(m \ll n)$, where it is reasonable for them to be sampled in a diverse way, respecting the training distribution but also the structural complexity of the target function. In \figref{fig:dopplerExperiment_IPdistribution_test_vs_opt}, we show a naive choice and our optimized choice of \inducingPoints.

In \figref{fig:dopplerExperiment_varyingIPstrategies}, we compare the \RMSE for the fixed training size $n=2^{15}$ for both settings, active and passive, when sampling the \inducingPoints according to the training density $p$, the \LFC and their geometric mean \eqref{eq:optimizedIPdistribution}.
First of all, we observe that we generally require less \inducingPoints with active sampling compared to \randomTestSampling, which originates from the fact that the \superiorTrainingDensity $\approxPSup{\GPR}{n}$ already respects \LFC to some degree. Next, we observe that the geometric mean of the training density and \LFC performs best, provided that the number of \inducingPoints $m$ is large enough. Finally, we observe that, non-surprisingly, we can shrink $m$ the most under the \LFC distribution, namely to $m = 128$, before the performance of the model degrades substantially.

In summary, we are able to shrink the model complexity up to a factor of $8$ for the Doppler function without a significant loss of performance, when respecting \LFC in the model design.

\fig{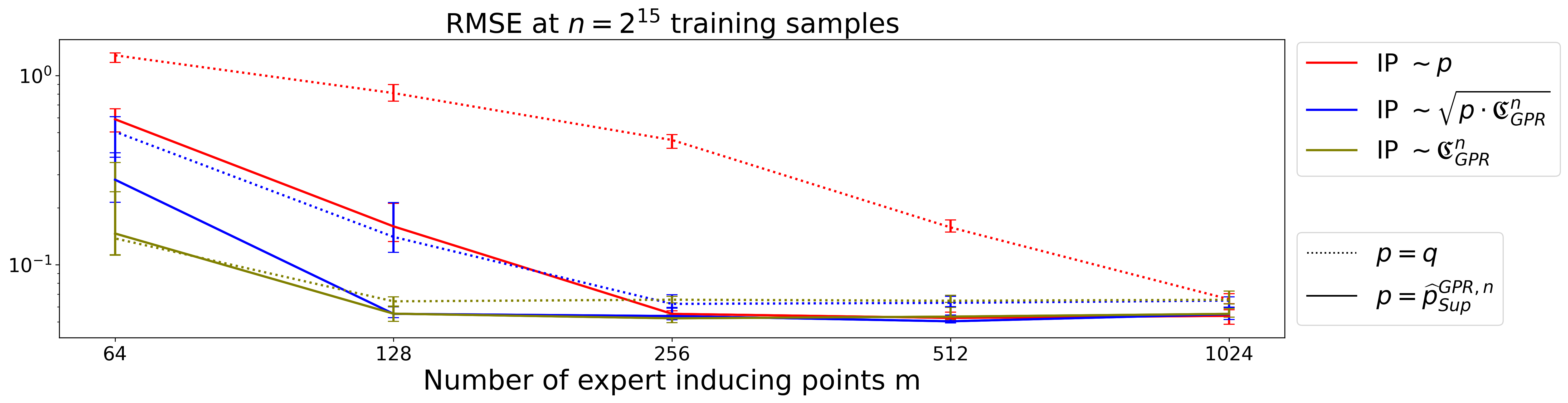}{1}{The Doppler experiment: The curves show the \RMSE at training size $n = 2^{15}$ for a varying number of expert \inducingPoints $m$. The colors correspond to different \inducingPoint distributions, whereas the line styles correspond to the underlying training distribution. The results are averaged over $20$ repetitions.}{0.7cm}{0.0cm}

\myParagraph{Comparing our proposed \inducingPoint selection method to a greedy fast forward selection}
In \secref{sec:relatedWork}, we discussed other \inducingPoint selection approaches. For comparison, we have implemented the \emph{greedy fast forward} (\GFF) \inducingPoint selection method of \cite{seeger2003fast}, in which, beginning from scratch, the most informative training inputs are gradually added to the set of \inducingPoints as a means to approximate the full $\GP{\theta}$ distribution.
Here, the information of an \inducingPoint candidate $x_i \in \trainingInputs{n} \setminus \expertInducingPoints{}$ is measures by
\[J(x_i) = \kullbackLeiblerDistance{Q_{\expertInducingPoints{} \cup \{x_i\}}}{Q_{\expertInducingPoints{}}},\]
which is the Kullback-Leibler divergence between the posterior distributions based on the \inducingPoints $\expertInducingPoints{} \cup \{x_i\}$ and $\expertInducingPoints{}$.
Accordingly, the updated set of \inducingPoints is given by $\expertInducingPoints{} \leftarrow \expertInducingPoints{} \cup \{x_i^*\}$, where
\[
x_i^* = \textstyle\argmax_{x_i \in \trainingInputs{n} \setminus \expertInducingPoints{}} J(x_i).
\]
The procedure converges, when the remaining \inducingPoint candidates carry no further information, that is, $J(x_i^*) < \varepsilon_\mathcal{J}$, up to a specified threshold $\varepsilon_\mathcal{J} \geq 0$.

At a given threshold $\varepsilon_\mathcal{J}$, we observe that the number of selected \inducingPoints is very small for the experts with a large bandwidth, while it increases drastically ($\propto \sigma_i^{-1}$) for experts with a small bandwidth. Now that the overall complexity of the \MoE is dominated by the expert with the most \inducingPoints, it is fair to compare the number of \inducingPoints of the expert at bandwidth $\sigma_1$ with our statically specified number $m$ of \inducingPoints in \figref{fig:dopplerExperiment_varyingIPstrategies}.
Here, we will vary the threshold $\varepsilon_\mathcal{J}$ to obtain a curve that maps the associated number of \inducingPoints to the achieved \RMSE. The results in \figref{fig:dopplerExperiment_ourIPsVsGFF} show that with \GFF almost no \inducingPoints can be saved for this inhomogeneously complex problem, as opposed to our proposed \inducingPoint selection method.

In any case, even under training according to $\pSup{\GPR}{n}$, the selected \inducingPoints by \cite{seeger2003fast} are uniformly distributed. In particular, the need for less \inducingPoints of the experts at smaller bandwidths to the right of \inputSpace is not recognized and, thus, we observe no \inducingPoint savings at an acceptable performance over a random \inducingPoint selection at all for this inhomogeneously complex problem.

\fig{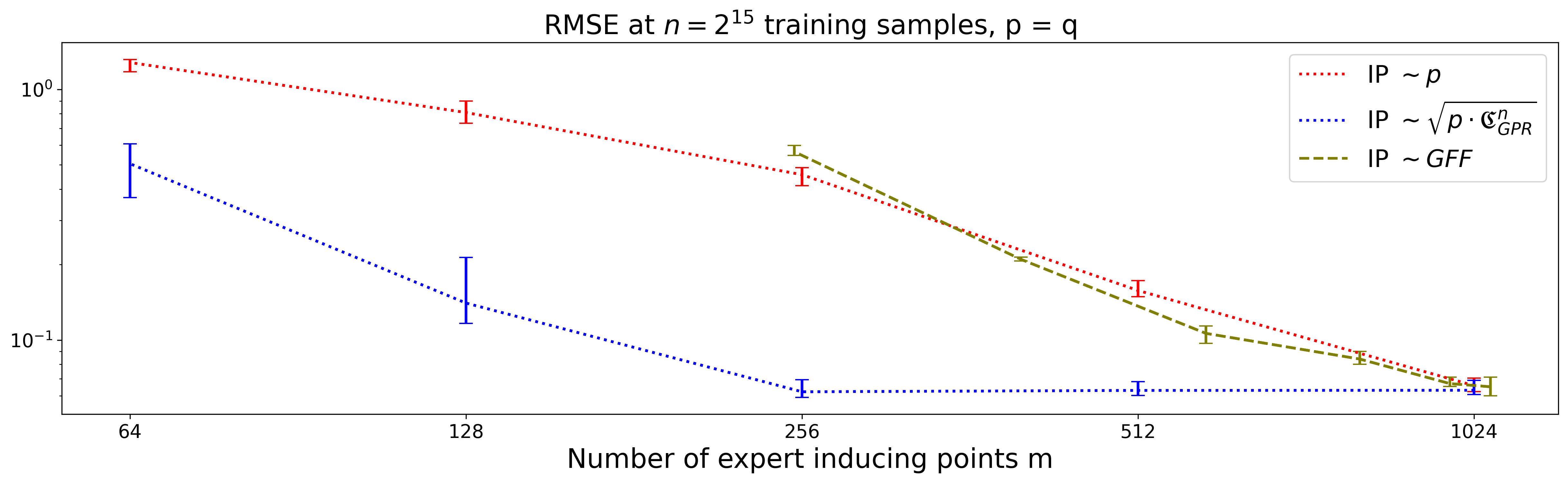}{1}{The Doppler experiment: The curves show the \RMSE at training size $n = 2^{15}$ at a varying number of expert \inducingPoints $m$, where the \inducingPoints are either chosen at random, according to our proposed selection method, and via \GFF. The results are averaged over $20$ repetitions.}{0.7cm}{0.0cm}

\fig{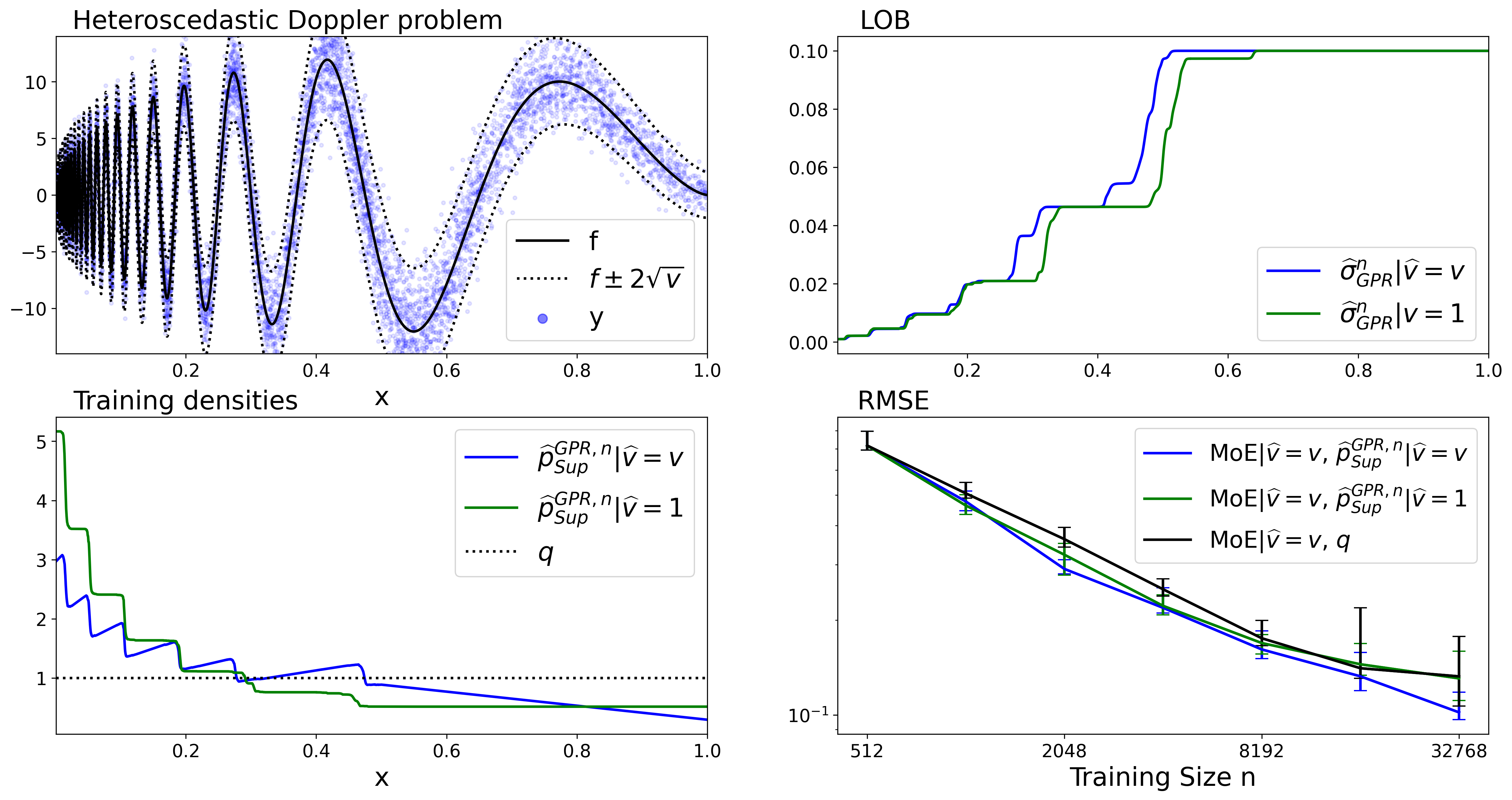}{1}{The Doppler experiment under heteroscedastic noise: (Top left) An exemplary dataset; (Top right) The \LOB estimates, when comparing the homoscedastic to the heteroscedastic Doppler experiment; (Bottom left) The training densities of \randomTestSampling, \pSup{\GPR}{n} and \pSup{\GPR}{n} when wrongly assuming homoscedasticity; (Bottom right) The \RMSE at several training sizes of the compared sampling schemes. The results are averaged over $20$ repetitions.}{0.8cm}{0.0cm}

\myParagraph{Heteroscedastic noise treatment}
While the treatment of heteroscedastic noise is not the main focus of this work, we will now demonstrate our approach on a heteroscedastic version of the Doppler experiment. For this, we let $v(x) = (3 - 4\abs{x-0.5}))^2 \in [1,9]$, which we plot in \figref{fig:dopplerHExperiment_data_pOpt_lob_rmse} (top left) together with the resulting dataset.
Here, we assume the local noise variance (or an estimate of it) to be provided externally, again, since its estimation is out of the scope of this work. However, note that the estimation of $v$ is well-studied in the literature, especially for \GaussianProcesses \citep{Kersting07mostlikely,cawley06,tresp2001mixtures}.

In \figref{fig:dopplerHExperiment_data_pOpt_lob_rmse} (top right) we compare the \LOB estimates obtained from the homoscedastic dataset and the heteroscedastic version.
As we have suggested in \secref{subsec:GPRlfcEstimate}, the influence of $v$ on the \LOB estimates obtained from heteroscedastic \GPR experts is relatively small. Likewise, we proceed with the evaluation of our proposed \AL scheme under heteroscedasticity.
Here, we compare to \randomTestSampling but also to $\pSup{\GPR}{n}$ under the wrong assumption of homoscedastic noise. Note that we use the heteroscedastic \MoE in all cases since the wrong assumption of homoscedastic noise in the experts makes the \MoE very volatile. The respective training densities and \RMSE learning curves can be seen in the bottom row in \figref{fig:dopplerHExperiment_data_pOpt_lob_rmse}. Due to the stronger noise (compared to the homoscedastic experiment), the asymptotic behavior begins to materialize later from $n=2^{13}$ training samples. Until this point, sampling only with respect to the structural complexity looks also promising. However, as soon as the training size becomes large enough to roughly resemble the target function, respecting the inhomogeneity in the noise level becomes crucial to achieve a homogeneous pointwise convergence and, thus, maintaining asymptotic superiority.

\subsection{Force field reconstruction}
\label{subsec:MDMalon}

We now turn our attention to a real-world example in which we predict the \emph{potential energy surface} (PES) and corresponding \emph{force field} (FF) of a molecule from first-principles calculations. The PES function links the geometry $x = \left[R_1,\ldots, R_{\bm{a}}\right]\in\reals{3\times \bm{a}}$ of a molecule to its potential energy $E\in\reals{}$, where $R_i$ are the Cartesian positions of the $\bm{a}$ atoms of the molecule.
In ab initio computations, this mapping is achieved by solving the time-independent Schr{\"o}dinger equation.
The PES encodes essential information on the properties of a molecule.
Due to thermal and quantum effects, molecules are never perfectly rigid but assume different configurations.
The distribution of these configurations is determined by the shape of the PES.
For example, the minima of the PES will be sampled more frequently than other regions and correspond to stable structures.
This has practical implications since many experimental techniques measure an expectation value over molecular distributions.
In order to achieve a meaningful comparison, sampling needs to be taken into account in theoretical simulations as well.
One of the most successful approaches to sample molecular distributions is MD simulation.
They model the time evolution of the atomic positions, sampling the PES by integrating Newton's equations of motion.
To this end, energy-conserving forces acting on each atom are required.
These forces are the negative derivative of the PES with respect to the atomic positions $F\in\reals{3\times \bm{a}}$.

This type of proxy for the prohibitively expensive ab initio quantum mechanical calculations is commonly used to enable long-timescale MD simulations that consist of millions of steps, each requiring the evaluation of the PES and FF for a new geometry.
Converged MD trajectories give unique insights into the dynamic behavior and structure-function relationships of physical systems at atomic scale. They are widely used in molecular biology research and play a crucial role in applications such as protein folding and drug discovery. \ML has the potential to profoundly advance this field, as it bears the promise of offering a unique cost-accuracy trade-off that is not achievable with traditional methods \citep{noe2020machine, von2020exploring, unke2020,keith2021combining}. However, some commonly deployed ML-based FFs rely on rather naive exhaustive sampling schemes to gather training data, which stands in the way of scaling to larger system sizes, both, from a data acquisition cost and training perspective. Here, we demonstrate how our method can be used to construct smaller, yet more effective training datasets.

\fig{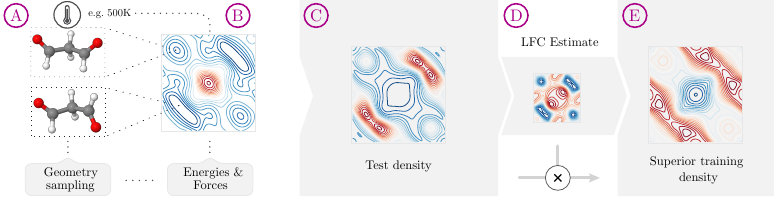}{1}{Reconstructing ML-based FFs using our \MoE approach:
(A-B) The inputs and outputs of the regression task are the geometries and energies (including forces, i.e., energy gradients) of malonaldehyde. As an example, we highlight the geometries of the two energetically stable states found in the local minima of the energy surface. (C) The density estimate of the true MD geometry distribution. (E) The \superiorTrainingDensity estimate \eqref{eq:GeneralizedOptimalSamplingInfAlpha} based on our approach.
All properties are evaluated at the relaxed malonaldehyde configurations and plotted with respect to the angles of the two aldehyde rotors of malonaldehyde (see \cite{chmiela2018,sauceda2020construction}).
}{0.6cm}{0.0cm}

In this experiment, we reconstruct a FF for the molecule malonaldehyde, which has $\bm{a} = 9$ atoms and the chemical formula $C_3H_4O_2$ (see \figref{fig:malonMD_overview} (A)). Formally, we try to infer the high-dimensional target function $\fctn{f}{\inputSpace}{\outputSpace}, R \mapsto [E,F]$, where $\inputSpace = \reals{3\bm{a}}$ and $\outputSpace = \reals{1+3\bm{a}}$.
For visualization purposes, we only show a two-dimensional subspace of the PES, which is characterized by the two main features of this molecule, its two rotors (aldehyde groups) \citep{chmiela2018,sauceda2020construction}. Their relative orientation is the dominant driver of the potential energy in this case and therefore most descriptive. Each point on the surface depicted in \figref{fig:malonMD_overview} (B) is generated by fixing the rotor pair at a particular angle and relaxing all remaining degrees of freedom to obtain a minimal energy configuration. We will refer to these geometries as the \emph{relaxed configurations} in the following.

To reconstruct the FF, we consider the broadly adopted \emph{symmetric gradient-domain machine learning} (\sGDML) method \citep{chmiela2018,chmiela2019}, which is a \GPR model that takes energy and force labels and also roto-translational and permutational invariances of the geometries into account (see \appref{subsec:sGDML} for details). We anticipate that \sGDML will benefit from our \MoE approach, where we deploy \sGDML as the expert model, since the transition paths along the PES vary in complexity, due to the interplay between distinct atom types with different characteristic interaction length scales.
Our \AL approach can only improve training efficiency if there are inhomogeneities in the data. Using our \LFC estimate, we therefore first verify our intuition that the PES of malonaldehyde varies in complexity. Based on this, we derive the \superiorTrainingDensity, which we finally input into our \AL framework to refine the training dataset in a superior way.

\myParagraph{Experimental setup}
All experiments use an extensive pre-computed reference trajectory (almost a million data points ($\poolInputs,\poolLabels$)) as ground truth, as opposed to generating new data points on demand. This test setup allows a post-hoc verification of the training distribution generated by our \AL approach, while still providing ample redundancy and therefore sampling freedom.

Recall from \secref{subsec:ALprocedure} that we require an unnormalized density estimate of the trajectory $\poolInputs \sim \poolDensity$ since we are dealing with a pool-based \AL scenario.
We estimate $\approxPoolDensity{}$ by standard \emph{kernel density estimation}, based on the
energy-to-energy entry of the \sGDML kernel $\widetilde{\bm{k}}$ from \eqref{eq:sGDMLKernel} at $\sigma = 0.03$.
\figref{fig:malonMD_overview} (C) shows the density estimate of the relaxed configurations, where we observe that $\poolDensity$ is very unbalanced, with a strong concentration of mass near the stable configurations.

We implement our \MoE approach, using the \sGDML kernel $\widetilde{\bm{k}}$ from \eqref{eq:sGDMLKernel} with a Gaussian base kernel function $k$.
While we sample the training data randomly (with appropriate weights) from the pool, we will draw sub-samples (i.e., for choosing the \inducingPoints of sparse expert and gate models) via symmetrized \emph{distributional clustering} (DC) with distributional \kMeansPlusPlus initialization (see \appref{sec:diverseIPs}).

Since this dataset comes with practically noise-free labels (we consider the first principle calculations as ground truth), we tune the experts (and \MoE model) with respect to \MSE rather than the \MLL objective. For stability, we will apply $\gpNoiseVariance(x) = 10^{-9}$ even though we assume no noise.


\figSideCaption{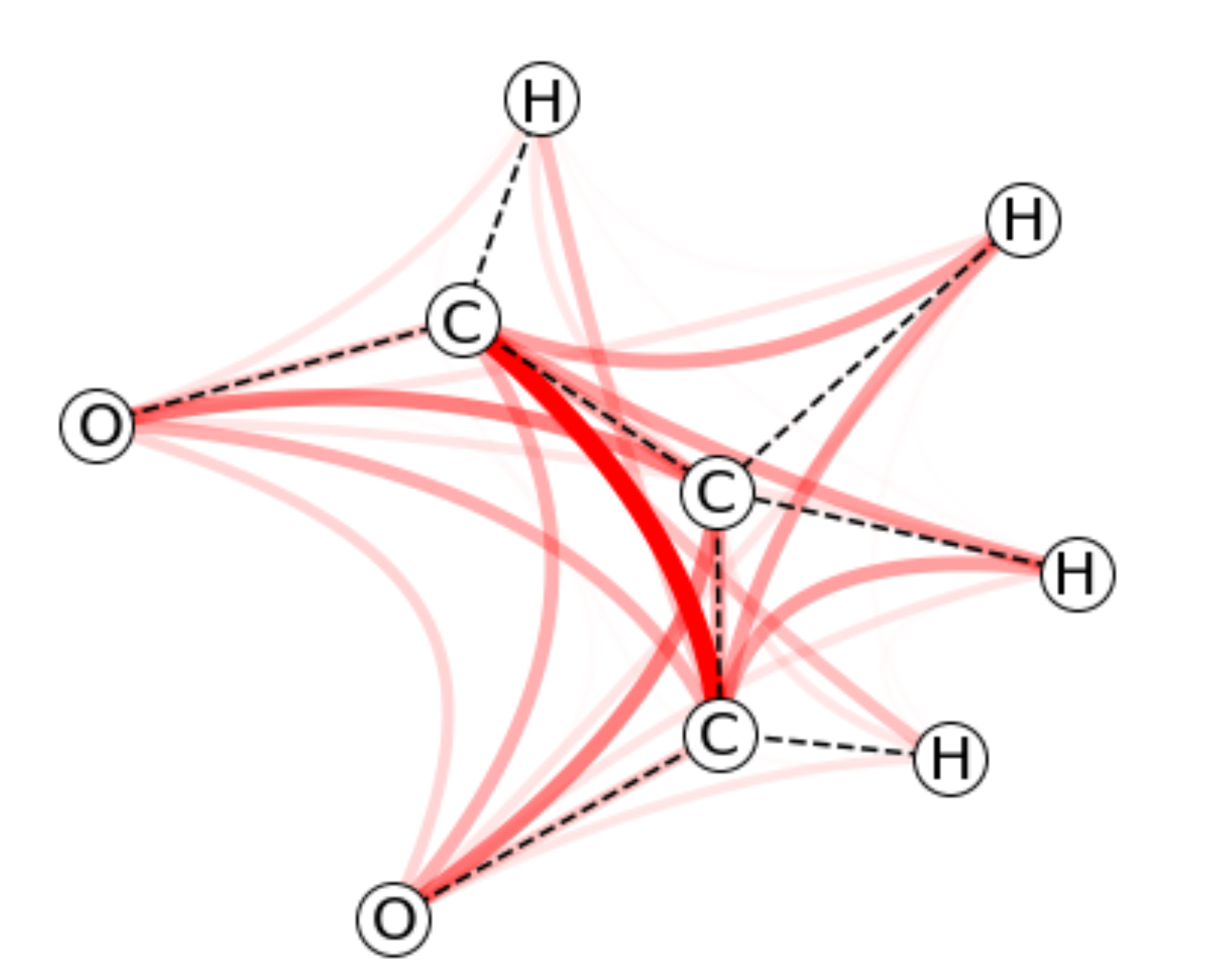}{1.1}{A visualization of the individual feature importance of malonaldehyde in the anisotropic case:
The structural formula of the molecule is plotted in black.
The importance of the individual interatomic distances is reciprocal to $\bandwidth_E$, which is the bandwidth estimate obtained by training the anisotropic \sGDML model.
Hence, we express the importance of each interatomic distance of the molecule in red, where the importance corresponds to the line saturation
$\gamma = \left[\frac{-\log(\bandwidth_E) - \min(-\log(\bandwidth_E))}{\max(-\log(\bandwidth_E)) - \min(-\log(\bandwidth_E))}\right]^2 \in [0,1]$.
}{7cm}{0.4cm}

\myParagraph{Anisotropic bandwidths}
\sGDML operates on $\bm{d} = \bm{a}(\bm{a}-1)/2 = 36$ features that are based on the interatomic distances of the molecule. In contrast to the work of Chmiela et al.~who restrict themselves to an isotropic bandwidth $\bandwidth_E = \sigma_E\idMatrix{\bm{d}}$, our implementation of \sGDML in GPyTorch naturally enables us to tune an anisotropic bandwidth $\bandwidth_E = \diag(\sigma_1,\ldots,\sigma_{\bm{d}})$ in the preprocessing step.

We partially offset the increased memory footprint of the model due to the tunable $\bandwidth_E$ by implementing the sparse \GPR model from \secref{subsubsec:sparseVarGP} under the \sGDML kernel $\widetilde{\bm{k}}$ from \eqref{eq:sGDMLKernel} and limiting the
number of \inducingPoints to $m=128$ configurations.
Since all our features are of the same type---pairwise interatomic distances---they are inherently calibrated in terms of scale. Hence, the reciprocal entries of $\bandwidth_E$ directly translate into the importance of the features, which we display in
\figref{fig:malonMDAnisotropicBandwidthsVisualization_cropped}.

We observe, that the importance assigned to some pairs of atoms agrees with chemical intuition, e.g., interactions with light hydrogen atoms are generally weaker. Furthermore, the important role of the opposing aldehyde groups in malonaldehyde emerges in the form of a heavily weighted path that connects the \mbox{O-C-C-C-O} backbone of the molecule.

In \figref{fig:malonMD_IsoVsAniVsMoEVsActivePerformance_pMD_rmse_withLogBox} we see that our anisotropic variant of \sGDML performs consistently better than the original isotropic \sGDML model.
Similar to the calculation of the relative sample size in \eqref{eq:relSampleSizeGPRApprox} we can compare two models of equal asymptotic \MSE law. When comparing anisotropic to isotropic \sGDML, both under \randomTestSampling, we can save about 10\% of samples.

\myParagraph{Setting up the \MoE model}
After having trained $\bandwidth_E$, we apply dense \sGDML experts with $\bandwidth_j = \sigma_j\bandwidth_E$, where $\sigma_j = 2^{-5/4 + j/2}, 1\leq j\leq 8$ as the individual expert bandwidths, $\lambda_E = 1$ as the initial expert regularization, and $\sigma_G = 0.1$ and $\lambda_G = 10^4$ for the sparse gate with $1024$ \inducingPoints.
For the training, we apply a batch size of $B = 1024$, a terminal expert sparsity $\sparsity = 8$, a penalty factor of $\vartheta_\sigma = 0.01$ for small bandwidth choices, gate noise parameters $\gateNoise_0 = 0.01$ and $\eta_{\gateNoise} = 1/\sqrt{2}$, and learning rate parameters $\eta = 0.005$, $\eta_H = 0.05$, $\eta_G = 0.1$.
As we discuss in \appref{sec:designChoices}, for tuning the \MoE with dense (\sGDML) experts, we either require an additional gate training set, which is independent of the training set for the experts, or we could provide leave-one-out (\LOO) responses of the experts for the training of the gate.
In our experiment, we use an additional gate training set $\bm{X}_{n_G}^{G}$ of fixed size $n_{G} = 2^{14}$.
The anisotropic \MoE model performs consistently better than anisotropic \sGDML, as can be seen in \figref{fig:malonMD_IsoVsAniVsMoEVsActivePerformance_pMD_rmse_withLogBox}.
When comparing the anisotropic \MoE model to isotropic \sGDML, both under \randomTestSampling, we can save about 21\% of samples.

\fig{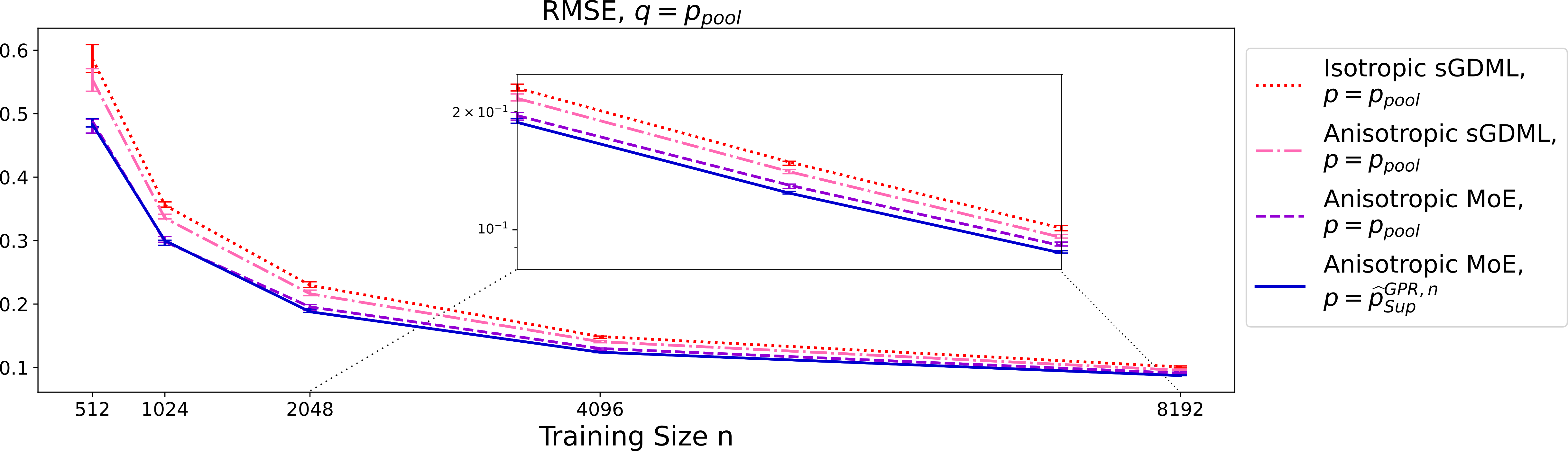}{1}{The \RMSE under the true MD trajectory test distribution for different variants of \sGDML and training distribution at varying training size: The performance is given for passive sampling, using the original isotropic \sGDML (dotted), anisotropic \sGDML (dash-dotted) and our \MoE model with anisotropic \sGDML experts (dashed), and for the \MoE model, applying the proposed \superiorSamplingScheme (solid). The results are averaged over $5$ repetitions.}{0.6cm}{0.0cm}

\myParagraph{Active learning}
We assume an intrinsic dimension of $\intrinsicDimension = 2$ (the two aldehyde rotor angles, the most salient features of malonaldehyde) and a smooth target function $f \in \diffableFunctions{\inputSpace,\outputSpace}{\infty}$.
The test distribution is given by the MD trajectory such that $q = \poolDensity$.
Prior to the \AL procedure, we separate the validation samples $\validationInputs$ and test samples $\trueTestInputs$ at random from the pool $\poolInputs$.
We apply an initial expert training size of $n_0 = 2^{9}$, doubling the sample size with each iteration of the \AL procedure.
The initial expert training set $\trainingInputs{n_0}$ and the gate training set $\bm{X}_{n_G}^{G}$ are drawn via importance sampling from the remaining pool with weights $\approxPoolDensity{-1/2}(\poolInputs \setminus(\validationInputs\cup \trueTestInputs))$. By this it is $\trainingInputs{n_0} \sim q^{1/2}$, which is more in alignment with the \superiorTrainingDensity \eqref{eq:GeneralizedOptimalSamplingInfAlpha} than sampling $\trainingInputs{n_0} \sim q$.

In \figref{fig:malonMD_overview} (D, E) we show the estimates of \LFC and the \superiorTrainingDensity under the pool test distribution, evaluated on the relaxed configurations of malonaldehyde.
The \LFC estimates confirm our expectation that the transition areas are more complex to model than the regions near the stable configurations. Subsequently, our active sampling scheme shifts sample mass away from the stable regimes in favor of the transition areas.

We have plotted the error curves of passive and active sampling schemes in \figref{fig:malonMD_IsoVsAniVsMoEVsActivePerformance_pMD_rmse_withLogBox}.
When estimating the relative sample size \eqref{eq:relSampleSizeGPRApprox} that we require to achieve the same \RMSE via active sampling compared to \randomTestSampling, we obtain $\textstyle\relSampleSize{\fMoE}{\approxPSup{\GPR}{n}} = 0.920\pm0.013$.
This means that we save about 8\% of samples under the \MoE model with our active sampling scheme compared to \randomTestSampling.
In total, when comparing our actively trained \MoE approach to the passively trained, original \sGDML model, we can save about 31\% of samples.
Notably, DFT level calculations \citep{perdew1996generalized,blum2009ab,tkatchenko2009accurate} for the studied system require minutes to hours of computation \emph{per sample}, CCSD(T) level computations even require days of computation per sample. So in the field of quantum chemistry saving roughly a third of computing power is of practical importance.

\section{Discussion}
\label{sec:discussion}


\myParagraph{Active learning}
Recall that in this work we have restricted ourselves to the scenario of model-agnostic \AL with persistent performance at large training size. This scenario is complementary to the more common small sample size regime. And while both cases are important, a lot of \AL related work (as discussed in \secref{sec:relatedWork}) does not apply to our \AL scenario.
We also discussed and demonstrated in our experiments that input space geometric arguments which are model-free asymptotically come with no benefit over \randomTestSampling.
Since our proposed model is \GPR-based, we also analyzed uncertainty sampling (\MoGPU) for our \MoE model to show that our proposed \superiorSamplingScheme differs from uncertainty sampling. Moreover, our \superiorSamplingScheme showed to be superior to uncertainty sampling.
Finally, in the regime of our \AL scenario, the approach by \cite{panknin2021optimal} recently has demonstrated \SOTA performance to recent, sophisticated, model-agnostic \AL approaches.
This was done by training different models on the actively constructed training sets of their and other \AL approaches and assessing their performance.
In particular, they compared favorably to \cite{goetz2018active}---a random tree-based \AL approach---in a heteroscedastic setting, using a regression forest model and to \cite{bull2013spatially}---a wavelet-based \AL approach---in a setting of inhomogeneous complexity, using an \RBF-network. This demonstrates the flexibility of this \AL approach in terms of learning problem specifications as well as model choices.
Now that our work builds on the previous work of \cite{panknin2021optimal}, \SOTA performance of our work is implied.

\myParagraph{Interpretability of \LFC}
Due to model-agnosticity, we consider \LFC to be an intrinsic, \emph{interpretable} property of the regression problem, which can be used as an analysis tool by domain experts:
\vspace*{-1mm}
\par\noindent
When looking at a single point in a high-dimensional input space, a visual assessment of the local structural complexity (e.g., a human can visually detect more complexity to the left of the Doppler function) is challenging. Here, the scalar \LFC value gives a human assessable, quantitative description.
The \LFC (as a scalar-valued function) then even allows for an easy visualization of the local structural complexity in high-dimensional input spaces, if the
input space features a reasonable low-dimensional projection.
This benefit was demonstrated in the high-dimensional FF reconstruction experiment, where the two-dimensional visualization of \LFC provides new insights into the regression problem.
Note that the \LFC function cannot be visualized in the absence of a low-dimensional projection.

\myParagraph{Parsimonious modeling}
We proposed a novel, model-agnostic approach to select the \inducingPoints of \GPR, sampling them in a diverse way from a distribution that is representative for the training data and respects the \LFC. In the experiments, we have seen that for problems of inhomogeneous complexity, our approach sustains the expressive power of the model at a considerably smaller number of \inducingPoints, compared to the \GFF \inducingPoint selection method of \cite{seeger2003fast}.

\myParagraph{Heteroscedasticity}
While inhomogeneities in noise are not the focus of our work, note that both, our model as well as the original \AL framework upon which we built our approach naturally deal with heteroscedasticity.
It, therefore, suffices to complement our work with an estimate of the noise variance function $v$ as, e.g., given in \cite{Kersting07mostlikely,cawley06}.
The only aspect left open is to elaborate on the impact of $v$ on the \LOB of \GPR and, thus, the adequate adjustment with respect to $v$ in the derivation of the \LFC.
As we argued, \GPR already treats heteroscedasticity through local adaptions of the regularization. Hence, we assume the influence of $v$ on the \LOB to be negligible to not existent. This is opposed to the \LOB of \LPS whose only way to deal with heteroscedasticity is through adaption of its \LOB.

\myParagraph{Intrinsic dimension and smoothness of the problem}
In our derivation of \LFC and the \superiorTrainingDensity, we assumed the intrinsic dimension $\intrinsicDimension \leq d$ and the smoothness $\alpha\in (0, \infty]$ of $f \in \diffableFunctions{\inputSpace,\R}{\alpha}$ to be given through domain knowledge.

If \intrinsicDimension is unknown, we can estimate it from unlabeled input instances, such as \poolInputs in a pre-processing step.
However, this is beyond the scope of our work and we refer to the approach of \cite{facco2017estimating}\footnote{For an implementation in Python see \url{https://scikit-dimension.readthedocs.io/en/latest/index.html}} for an estimate of the \intrinsicDimension for unbalanced input distributions in high-dimensional input spaces \inputSpace.

If we have no ground truth knowledge about the smoothness $\alpha$ of the target function $f \in \diffableFunctions{\inputSpace,\R}{\alpha}$, we resort to $\widehat{\alpha}=\infty$ as default in practice. This assumption is justified, as long as $f$ happens to be rougher in at most finitely many locations of the input space.
Since, asymptotically, the violation of $\alpha = \infty$ affects only a set of measure zero, the influence on our \AL setting that addresses large training sizes is marginal.
We consider the restriction to target functions that are at most rough on a finite set of input space locations a weak assumption which, thus, comes with no practical limitations.

One way to deal with an unknown smoothness $\alpha$ is to deploy the \emph{Mat\'ern} kernel 
\[
k_{\nu}(x,x') := \frac{2^{1-\nu}}{\Gamma(\nu)}\left(\pnorm{x-x'}{}/\sigma\right)^{\nu}\widetilde{k}_\nu\left(\pnorm{x-x'}{}/\sigma\right),
\]
where $\Gamma$ is the \emph{gamma function} and $\widetilde{k}_\nu$ is the modified Bessel function of the second kind or order $\nu$. After finding the best fitting $\nu^{*}$, we obtain by $\widehat{\alpha} := \lceil\nu^{*}\rceil - 1$ a reasonable estimate to $\alpha$. We defer this idea to future work as it is beyond the scope of this work.

\myParagraph{Dimensional scaling}
 As opposed to nonstationary GP approaches (e.g., the tree-based or local GP by \cite{gramacy2008bayesian,gramacy2015local} that suffer from the curse of dimensionality through input space localization, we segment the input space into a fixed number $L$ of patches, given by the $L$ experts of our \MoE. Thus, if we were to instantiate our \MoE with dense \GPR models, our approach scales well concerning the input space dimension $d$.
 However, in real-world applications, we typically deal with training sizes that are too large for dense modeling. In this regime, sparse \GPR representations scale poorly in $d$ as their \inducingPoints must be space-filling \citep{binois2022survey}. Likewise, our density-based \AL approach encounters decaying power for large $d$.
 A low intrinsic dimension ($\intrinsicDimension < d$) of the regression problem is therefore crucial for our work to apply.

\myParagraph{\LFC and the \superiorSamplingScheme as a \ML concept}
Recall that we consider \LFC to be a problem intrinsic property. Here, the problem is characterized by features $x \in \inputSpace$ with labels $f(x)\in \R$, a hypothesis space of \locallyAdaptiveModels, and the \MISE as the loss function (or rather the pointwise \MSE from \eqref{eq:conditionalMSE} due to the localization of \LFC). In this sense, \LFC is formally a property of the combination of model and loss according to the characterization of (\cite{jung2022basics},~Chapter 2).
Similarly, the \superiorTrainingDensity is a property of the combination of model and loss with respect to the same hypothesis space and \MISE instead of the pointwise \MSE as loss function.  

\myParagraph{On a realistic implementation in ab initio FF reconstruction}
In our FF reconstruction experiment, we assumed a large unlabeled reference trajectory \poolInputs to be given that already follows the true molecular distribution. This will not be given in practice, since building the input trajectory already requires the computationally expensive estimation of the respective labels. At this point, the actual task behind the regression problem would already be solved. We outline a realistic ab initio FF reconstruction scenario in \appref{sec:MDrealistic}.

\section{Conclusion}
\label{sec:conclusion}
Standard \ML tasks implicitly assume a certain homogeneity in the data scales. However, in practice this structural property of the learning problem may not be fulfilled, e.g., in multiscale problems from the sciences such as turbulence \citep{brunton2020machine} or quantum chemistry \citep{noe2020machine,von2020exploring,unke2020,keith2021combining}.

In this work, we aimed to identify local inhomogeneities in regression tasks, which can be used to construct better models and training datasets and for domain interpretation.
To this end, we combined recent results on model-agnostic \LFC estimates and asymptotically optimal sampling, which are founded in the domain of \LPS, with estimates of \LOB, which are derived in the \GPR domain. By this, we benefit from both sides, having a theoretically sound superior sampling scheme on the one hand, and having access to the required estimates from a model that naturally can cope with high input space dimensions on the other hand.
Furthermore, we have shown how respecting \LFC in the selection of \inducingPoints contributes to parsimonious modeling.

On synthetic data, we showcased and validated our approach, where we analyzed similarities with the \LPS-based analog but also compared to the most related \GPU concepts for \AL.
To show the full potential of our approach, we studied a real-world, high-dimensional force field reconstruction task.
Our approach not only gave access to an interpretable visualization of the inhomogeneous structural complexity but also guided the sampling process in a way that takes the structural changes into account, enhancing the quality of the training data.
Here, we additionally identified the multi-scale structure of the individual atomic interactions, whose treatment also results in a substantial performance gain of the broadly adopted method \sGDML.

\myParagraph{Future work}
In \secref{subsec:GPRlfcEstimate} we conjecture that the \LOB of heteroscedastic \GPR is invariant or scales at most weakly with respect to the local noise level $v(x)$. This claim should be supported by further theoretical investigation.
While we deployed our estimates of \LFC and the superior training density, using $\widehat{\alpha} = \infty$, if the smoothness of the target function $f \in \diffableFunctions{\inputSpace,\R}{\alpha}$ is unknown,
it is possible to (re-)estimate $\widehat{\alpha}$, e.g., by tuning the regularity of the  Mat\'ern kernel of a \GPR model after the acquisition of each new training data batch.
While we have compared to baseline \inducingPoint selection methods, a thorough comparison to more sophisticated approaches remains open. A promising idea is also to combine our \LFC estimate with the \inducingPoint selection approach by \cite{moss2023inducing} to obtain informative and diverse \inducingPoints in \GPR.
Finally, we will focus on the application of our approach to real-world problems from chemistry, physics, and further domains also applying techniques from \emph{explainable AI} (e.g. \cite{samek2021explaining, letzgus2022toward}). In particular, recent advances on \sGDML regarding the scalability by \cite{chmiela2022} will enable the application of our approach to large molecular systems.







\subsubsection*{Acknowledgments}
\DP, \SC, \SN, and \KRM were funded by the German Ministry for Education and Research as BIFOLD - Berlin Institute for the Foundations of Learning and Data (ref.~BIFOLD23B).
\DP was also supported by the BMBF project ALICE III, Autonomous Learning in Complex Environments (01IS18049B).
\KRM was also supported by the BMBF Grants 01GQ1115 and 01GQ0850,
under the Grants 01IS14013A-E, 031L0207A-D; DFG under Grant Math+, EXC 2046/1, Project ID 390685689 and by the Institute of Information \& Communications Technology Planning \& Evaluation (IITP) grants funded by the Korea Government (No.~2017-0-00451, Development of BCI based Brain and Cognitive Computing Technology for Recognizing User’s Intentions using Deep Learning) and funded by the Korea Government (No.~2019-0-00079, Artificial Intelligence Graduate School Program, Korea University).

All funding sources were not involved in the process of writing and submitting this work.

\bibliography{tmlr}
\bibliographystyle{tmlr}

\appendix

\section{Asymptotic results for local polynomial smoothing}
\label{sec:optimalSamplingSupplement}
In this section, we will review the theory of \cite{panknin2021optimal}.

The prediction of the \LPS model of order \LPSorder under the bandwidth $\bandwidth\in\posDefSet{\inputDimension}$ in $x\in\inputSpace$ can be understood as follows: First, the regression problem is localized around $x$ according to weights $k^{\bandwidth}(\cdot,x)$ that decrease with growing distance to $x$.
Then we search for the polynomial up to order \LPSorder that fits the localized regression problem best. Finally, the evaluation of this polynomial in $x$ is returned as the prediction.
Formally, it is
\begin{align}
 \label{eq:lpsPredictionGeneral}
 \predictorLPS{\LPSorder}{\bandwidth}(x) &= \mathfrak{p}_{\LPSorder,\bandwidth,x}^*(0),
 \mbox{ where }\\
 \notag
 \mathfrak{p}_{\LPSorder,\bandwidth,x}^* &= \argmin\limits_{\mathfrak{p} \in \mathcal{P}_{\LPSorder}(\reals{\inputDimension})} \mySum{i=1}{n}k^{\bandwidth}(x_i^{},x)\left(y_i^{}-\mathfrak{p}(x_i^{}-x)\right)^2,
\end{align}
and $\mathcal{P}_{\LPSorder}(\reals{\inputDimension})$ is the space of the real polynomial mappings $\fctn{\mathfrak{p}}{\reals{\inputDimension}}{\R}$ up to order \LPSorder.

The localization is controlled by $\bandwidth$ through the kernel weights $\kernelMatrix{\bandwidth}(x, x_i)$ for $x_i\in \trainingInputs{n}$:
For an \RBF-kernel, $k^\bandwidth(x,x')$ decays monotonically with growing distance of $x'$ to $x$.
This decay is dampened or amplified as \bandwidth increases or decreases, respectively (in the sense of the Loewner order).

For readability, since $\bandwidth$ will be replaced by terms with more involved notation, we redefine \eqref{eq:conditionalMSE} by
\begin{align}
\label{eq:conditionalMSEsigma}
\MSE\left(x, \widehat{f}, \bandwidth | \trainingInputs{n}\right) := \MSE\left(x, \widehat{f}^\bandwidth | \trainingInputs{n}\right).
\end{align}

For a bandwidth space $\SigmaSpace \subseteq \posDefSet{\inputDimension}$, \cite{panknin2021optimal} proposed to minimize the \AL objective
\begin{align}
\label{eq:conditionalMISEsigma}
\MISE\left(q, \widehat{f} | \trainingInputs{n}\right) = \textstyle \mathop{\mathlarger{\int}}_{\hspace*{-5pt}\inputSpace} \inf_{\bandwidth \in \SigmaSpace}\MSE\left(x, \widehat{f}, \bandwidth | \trainingInputs{n}\right) q(x) dx,
\end{align}
which is the optimal \MISE, obtained by predictions that are based on locally optimal chosen bandwidths.
If these locally optimal bandwidth choices are well-defined, that is, if for all $x\in\inputSpace$ there exists a unique $\bandwidth' \in \SigmaSpace$ such that 
\[
\textstyle\MSE\left(x, \widehat{f}, \bandwidth' | \trainingInputs{n}\right) = \inf_{\bandwidth \in \SigmaSpace}\MSE\left(x, \widehat{f}, \bandwidth | \trainingInputs{n}\right),
\]
we are able to define the \LOB function
\begin{align*}
 \LOBfunctionOfLPS{}{n}(x)
  = \textstyle \argmin_{\bandwidth \in \SigmaSpace}\MSE\left(x, \widehat{f}, \bandwidth | \trainingInputs{n}\right).
\end{align*}
This function exists, for example, in the \emph{isotropic} case $\SigmaSpace = \condset{\sigma\idMatrix{\inputDimension}}{\sigma > 0}$ for \LPS under mild conditions, where we denote $\LOBfunctionOfLPS{}{n}(x) = \lobfunctionOfLPS{}{n}(x)\idMatrix{\inputDimension}$ (see, e.g.,~\cite{masry1996multivariate,masry1997multivariate,fan1997local} or \cite{panknin2021optimal} for an overview).

Assuming the isotropic bandwidths candidate space $\SigmaSpace = \condset{\sigma\idMatrix{\inputDimension}}{\sigma > 0}$, the \LOB as in Eq.~\eqref{eq:LOBDefinition} is an asymptotically well-defined function under mild assumptions\footnote{We require non-vanishing leading bias- and variance-terms of $\predictorLPS{\LPSorder}{}(x)$, which is guaranteed if $\forall x\in\inputSpace$ it holds that $\biasLPS{\idMatrix{\inputDimension}}{x}{\LPSorder} \neq 0$ from Eq.~\eqref{eq:asymptoticBiasofLPS} and $v(x) > 0$.}:
Denoting the \LOB of \LPS of order \LPSorder by $\LOBfunctionOfLPS{\LPSorder}{n}(x) = \lobfunctionOfLPS{\LPSorder}{n}(x)\idMatrix{\inputDimension}$ such that
\begin{align}
\label{eq:lpsLOBDefinition}
 \lobfunctionOfLPS{\LPSorder}{n}(x)
  = \textstyle \argmin_{\sigma > 0}\MSE\left(x, \predictorLPS{\LPSorder}{}, \sigma\idMatrix{\inputDimension} | \trainingInputs{n}\right),    
\end{align}
asymptotically it holds
\begin{align}
 \label{eq:asymptoticIsotropicLOBofLPS}
 &\lobfunctionOfLPS{\LPSorder}{n}(x) =\textstyle C_{\LPSorder}^{}\left[\frac{v(x)}{p(x)n}\right]^\frac{1}{2(\LPSorder+1)+\inputDimension}\biasLPS{\idMatrix{\inputDimension}}{x}{\LPSorder}^{-\frac{2}{2(\LPSorder+1)+\inputDimension}} + \convergenceInProbability{}{n^{-\frac{1}{2(\LPSorder+1)+\inputDimension}}},
\end{align}
where $C_{\LPSorder}^{}$ is a constant, and $\biasLPS{\idMatrix{\inputDimension}}{x}{\LPSorder}$ is a function of $x$ taken from the asymptotic \emph{conditional bias} $f(x) - \E\left[\predictorLPS{\LPSorder}{h_n\idMatrix{\inputDimension}}(x)\middle|\trainingInputs{n}\right]$ of \LPS \citep{masry1996multivariate,masry1997multivariate}. That is, for a sequence $h_n\rightarrow 0$ as $n\rightarrow\infty$ we can write the conditional bias, which is of order $\LPSorder+1$, as
\begin{align}
\label{eq:asymptoticBiasofLPS}
\textstyle f(x) - \E\left[\predictorLPS{\LPSorder}{h_n\idMatrix{\inputDimension}}(x)\middle|\trainingInputs{n}\right] = h_n^{\LPSorder+1}\biasLPS{\idMatrix{\inputDimension}}{x}{\LPSorder} + \convergenceInProbability{\noexpand\big}{h_n^{\LPSorder+1}}.
\end{align}

Eq.~\eqref{eq:asymptoticIsotropicLOBofLPS} shows how \LOB scales asymptotically with respect to the training size $n$, the local noise level function $v(x)$ and the training density $p(x)$.
The remaining bias component depends on the local structural complexity, which can be characterized by the derivatives of $f$ in a non-trivial way. Therefore it encodes the local structural complexity of $f$.
Given all other properties and \LOB itself, we are able to formulate \LFC in a closed form.
\begin{restatable}[\cite{panknin2021optimal}]{mydef}{isotropicComplexityForLPS}
\label{def:isotropicComplexityForLPS}
For \LPS of order \LPSorder, the \LFC of $f$ in $x\in\inputSpace$ is asymptotically given by
\[
 \fctnComplexityLPS{\LPSorder}{n}(x) = \left[\frac{v(x)}{p(x)n}\right]^{\frac{\inputDimension}{2(\LPSorder+1)+\inputDimension}}\abs{\LOBfunctionOfLPS{\LPSorder}{n}(x)}^{-1} = \left[\frac{v(x)}{p(x)n}\right]^{\frac{\inputDimension}{2(\LPSorder+1)+\inputDimension}}\lobfunctionOfLPS{\LPSorder}{n}(x)^{-d}.
\]
\end{restatable}

As already mentioned in Eq.~\eqref{eq:conditionalMISEsigma}, given a test density $q$, the \AL task is to minimize $\MISE\left(q, \predictorLPS{\LPSorder}{} | \trainingInputs{n}\right)$.
Now, if \LOB is well-defined, we can rewrite
\begin{align*}
\MISE\left(q, \predictorLPS{\LPSorder}{} | \trainingInputs{n}\right) &= \textstyle \mathop{\mathlarger{\int}}_{\hspace*{-5pt}\inputSpace} \inf_{\bandwidth \in \SigmaSpace}\MSE\left(x, \predictorLPS{\LPSorder}{}, \bandwidth | \trainingInputs{n}\right) q(x) dx\\
&= \textstyle \mathop{\mathlarger{\int}}_{\hspace*{-5pt}\inputSpace} \MSE\left(x, \predictorLPS{\LPSorder}{}, \LOBfunctionOfLPS{\LPSorder}{n}(x) | \trainingInputs{n}\right) q(x) dx.
\end{align*}
Finally, when solving for the optimal training dataset
\begin{align*}
 \optTrainingInputs{n} \approx \textstyle \argmin_{\trainingInputs{n} \in \inputSpace^n_{} }\MISE\left(q, \predictorLPS{\LPSorder}{} | \trainingInputs{n}\right),
\end{align*}
as in Eq.~\eqref{eq:ALtask},
the optimal training inputs $\optTrainingInputs{n}$ can be written asymptotically as an independent and identically distributed sample from the optimal training distribution, whose density $\pOptLPS{\LPSorder}{n}$ possesses an asymptotic closed form.
\begin{mythm}[\cite{panknin2021optimal}]
 \label{thm:OptimalSampling}
 Let $v, q \in \diffableFunctions{\inputSpace, \nonnegativeReal{}}{0}$ for a compact input space \inputSpace, where $q$ is a test probability density.
 Additionally, assume that $v$ and $q$ are bounded away from zero. I.e., $v,q \geq \epsilon$ for some $\epsilon > 0$.
 Let $k$ be a \RBF-kernel with bandwidth parameter space $\SigmaSpace = \condset{\sigma\idMatrix{\inputDimension}}{\sigma > 0}$. Let $\LPSorder \in \N$ be odd and $f \in \diffableFunctions{\inputSpace}{\LPSorder+1}$ such that the bias of order $\LPSorder+1$ does not vanish almost everywhere.
 Then the \optimalTrainingDensity for \LPS of order \LPSorder is asymptotically given by
\[
 \pOptLPS{\LPSorder}{n}(x) \propto \textstyle \left[\fctnComplexityLPS{\LPSorder}{n}(x) q(x)\right]^{\frac{2(\LPSorder+1)+\inputDimension}{4(\LPSorder+1)+\inputDimension}}v(x)^{\frac{2(\LPSorder+1)}{4(\LPSorder+1)+\inputDimension}}(1 + o(1)).
\]
\end{mythm}
We will use this optimal distribution to sample $\optTrainingInputs{n} \sim \pOptLPS{\LPSorder}{n}$ with a proposed estimator for $\fctnComplexityLPS{\LPSorder}{n}$ that is scalable with respect to the input space dimension.

For \LPS with $\bm{X}_{n}' \sim p$ and $\trainingInputs{n} \sim q$, we can asymptotically calculate the relative required sample size from \defref{def:relSampleSize} in \secref{subsec:GPRlfcEstimate} by
\begin{align}
 \label{eq:relSampleSizeLPSApprox}
 \textstyle \relSampleSize{\predictorLPS{\LPSorder}{}}{p} = \left[\frac{\MISE\left(q, \predictorLPS{\LPSorder}{} | \bm{X}_{n}'\right)}{\MISE\left(q, \predictorLPS{\LPSorder}{} | \trainingInputs{n}\right)}\right]^\frac{2(Q+1)+\inputDimension}{2(Q+1)}.
\end{align}

\section{Analytic \GPR formulations}
\label{sec:analyticGPR}

\subsection{Classical Gaussian process regression}
\label{subsec:analyticGP}
The \GPR model $\widehat{y} \sim \GP{\theta}$ (see, e.g.~\cite{williams1996gaussian}) is defined as follows:
The \GaussianProcess is described by the hyperparameters $\theta = (\mu, \lambda, \gpNoiseVariance, \bandwidth)$, which are the global constant prior mean $\mu$, the regularization parameter $\lambda$, the label noise variance function $\gpNoiseVariance$ and the bandwidth matrix $\bandwidth$ of the kernel. If we can assume homoscedastic noise, we let $\gpNoiseVariance(x) \equiv \sigma_\varepsilon^2$.

The \GaussianProcess prior then assumes the labels $\trainingLabels{n}$ of $\trainingInputs{n}$ to be distributed according to 
$\trainingLabels{n} = \widehat{y}(\trainingInputs{n}) \sim \Gauss{\cdot}{\bm{\mu}(\trainingInputs{n})}{\bm{C}(\trainingInputs{n}) | \theta}$, for the
constant mean function
$\bm{\mu}(\trainingInputs{n}) = \mu\ones{n}$,
and the covariance function
\[\bm{C}(\trainingInputs{n}) = \lambda\bm{K}_n + \diag(\gpNoiseVariance(\trainingInputs{n})),
\]
where $\bm{K}_n = \kernelMatrix{\bandwidth}(\trainingInputs{n})$ is the kernel matrix of $\trainingInputs{n}$.

For test inputs $\evalInputs$, the posterior predictive distribution of $\evalLabels$ is then given by
\[\widehat{y}(\evalInputs) \sim \Gauss{\cdot}{\bm{\mu}^*(\evalInputs)}{\bm{C}^*(\evalInputs) | \theta},\]
where the predictive mean and covariance are given by
\begin{align}
\bm{\mu}^*(\evalInputs) &= \bm{\mu}(\evalInputs) + \bm{C}_{*n}\bm{C}_{n}^{-1}(\trainingLabels{n} - \bm{\mu}(\trainingInputs{n})),\\
\bm{C}^*(\evalInputs) &= \bm{C}_{*} - \bm{C}_{*n}\bm{C}_{n}^{-1}\bm{C}_{*n}^\top,
\end{align}
and we have defined
\[\bm{C}(\trainingInputs{n} \cup \evalInputs) = \begin{bmatrix}
\bm{C}_n & \bm{C}_{*n}^\top \\ \bm{C}_{*n} & \bm{C}_*
\end{bmatrix}.
\]

\subsection{Analytic sparse Gaussian processes}
\label{subsec:analyticSparseGP}
We define the sparse \GPR model $\widehat{y} \sim \SGP{\theta}$ as follows, following \cite{snelson2005sparse}:
The sparse \GaussianProcess is described by the (hyper-) parameters $\theta = (\mu, \lambda, \gpNoiseVariance, \bandwidth, \expertInducingPoints{})$, which are the global constant prior mean $\mu$, the regularization parameter $\lambda$, the label noise variance function $\gpNoiseVariance$, the bandwidth matrix $\bandwidth$ of the kernel and the prior distribution, given by the \inducingPoint locations $\expertInducingPoints{} \in \inputSpace^m$.

Here, the degree of sparsity is described by $m$ \inducingPoints: This number can be fixed in advance or gradually increased with training size $n$, where the increase $m_n = o[n]$ is typically much slower than $n$.
If we can assume homoscedastic noise, we let $\gpNoiseVariance(x) \equiv \sigma_\varepsilon^2$.

The sparse \GaussianProcess then outputs
\[\widehat{y}(\evalInputs) \sim \Gauss{\cdot}{\bm{\mu}^*(\evalInputs)}{\bm{C}^*(\evalInputs) | \theta_e}\] for the
mean function
\[\bm{\mu}^*(\evalInputs) = \bm{K}_{*\ipSymbol}\bm{Q}_{\ipSymbol}^{-1}\bm{K}_{n\ipSymbol}^\top(\Lambda + \diag(\gpNoiseVariance(\trainingInputs{n})))^{-1}(\trainingLabels{n} - \bm{\mu}(\trainingInputs{n}))\]
and the covariance function
\[\bm{C}^*(\evalInputs) = \bm{K}_{*} - \bm{K}_{*\ipSymbol}(\bm{K}_{\ipSymbol}^{-1} - \bm{Q}_{\ipSymbol}^{-1})\bm{K}_{*\ipSymbol}^\top + \diag(\gpNoiseVariance(\evalInputs))\]
where we have defined $\bm{K}_\ipSymbol = \kernelMatrix{\bandwidth}(\ipInputs)$, $\bm{K}_n = \kernelMatrix{\bandwidth}(\trainingInputs{n})$, $\bm{K}_{*\ipSymbol} = \kernelMatrix{\bandwidth}(\evalInputs, \expertInducingPoints{})$, $\bm{K}_{n\ipSymbol} = \kernelMatrix{\bandwidth}(\trainingInputs{n},\expertInducingPoints{})$,
\mbox{$\bm{Q}_{\ipSymbol} = \bm{K}_{\ipSymbol} + \bm{K}_{n\ipSymbol}^\top(\Lambda + \diag(\gpNoiseVariance(\trainingInputs{n})))^{-1}\bm{K}_{n\ipSymbol},$}
and $\Lambda = \diag(\bm{\lambda})$ with
\(\bm{\lambda} = \diag(\bm{K}_n + \bm{K}_{n\ipSymbol} \bm{K}_{\ipSymbol}^{-1} \bm{K}_{n\ipSymbol}^\top).\)

We choose $\bm{\mu}$ to be the constant mean function, i.e., $\bm{\mu}(X) = \mu\ones{n}$ for $X \in \inputSpace^n$, noting that other mean functions are possible.

\section{\LFC of \GPR}
\label{sec:proofLFCGPR}
\isotropicComplexityForGPR*
\begin{proof}
Let $\inputSpace = \sideset{}{_{i=1}^k}\biguplus\inputSpace^k_i$ be a segmentation of the input space with non-empty interiors $(\inputSpace^k_1)^\circ,\ldots,(\inputSpace^k_k)^\circ \neq \emptyset$, over which we can define the restricted bandwidth function search space
\begin{align*}
    \SigmaStep{k} = \condset{\bandwidth(x) = \mySum{i=1}{k} \indicatorFunction{\inputSpace^k_i}{x}\bandwidth_i}{\bandwidth_1,\ldots,\bandwidth_k\in\SigmaSpace}.
\end{align*}
Here, $\indicatorFunction{A}{z}$ is the indicator function, returning $1$ for $z\in A$ and $0$, else.
Furthermore let $\bandwidth^{k,n} \in \SigmaStep{k}$ be the minimizer of the \MISE over $\SigmaStep{k}$ with $\bandwidth^{k,n}(x) = \mySum{i=1}{k} \indicatorFunction{\inputSpace^k_i}{x}\bandwidth^{k,n}_i$ such that
\[\textstyle \myInt{\inputSpace} \MSE\left(x, \widehat{f}^{\bandwidth^{k,n}(x)} | \trainingInputs{n}\right) q(x) dx = \min_{\bandwidth \in \SigmaStep{k}} \myInt{\inputSpace} \MSE\left(x, \widehat{f}^{\bandwidth(x)} | \trainingInputs{n}\right) q(x) dx.\]
Recall from \eqref{eq:GPRbandwidthSamplesizeProportionality} that $\LOBfunctionOfLPS{\GPR}{n} \propto n^{-\frac{1}{2\alpha + d}}$ generally holds for arbitrary input spaces.
Due to this, asymptotically, $\widehat{f}_{|_{\inputSpace^k_i}}$ does not depend on training samples outside $\inputSpace^k_i$.
Hence, letting $\trainingInputs{i,n} := \condset{x\in\trainingInputs{n}}{x_i\in\inputSpace^k_i}$, the individual $\bandwidth^{k,n}_i$ are asymptotically found by solving the isolated segments of the objective
\[\textstyle \myInt{\inputSpace^k_i} \MSE\left(x, \widehat{f}^{\bandwidth^{k,n}_i} | \trainingInputs{i,n}\right) q(x) dx = \min_{\bandwidth \in \SigmaSpace} \myInt{\inputSpace^k_i} \MSE\left(x, \widehat{f}^{\bandwidth} | \trainingInputs{i,n}\right) q(x) dx.\]
First of all, it is $\E\trainingInputs{i,n} = p(\inputSpace^k_i)n$, where $p(A) := \myInt{A}p(x)dx$ is the probability for a training sample to fall into $A \subset \inputSpace$.
In addition, we need to account for the expanse of $\inputSpace^{k_n}_i$, which we measure by $\vol(\inputSpace^{k_n}_i)$. Here, $\vol(A) := \textstyle\myInt{A}dx$ is the volume of $A \subset \inputSpace$.
Again with \eqref{eq:GPRbandwidthSamplesizeProportionality}, it is therefore
\[\bandwidth^{k,n}_i \propto \left[p(\inputSpace^{k_n}_i) / \vol(\inputSpace^{k_n}_i)n\right]^{-\frac{1}{2\alpha + d}}.\]
Subsequently, we can slowly refine the segmentation $\inputSpace = \sideset{}{_{i=1}^{k_n}}\biguplus\inputSpace^{k_n}_i$, where $\max_{1\leq i \leq k_n}\vol(\inputSpace^{k_n}_i) \rightarrow 0$ for $k_n \rightarrow \infty$ slow enough (with $k_n = o(n)$).
Then, for almost every $x\in\inputSpace$, there exists a sequence $(i_{k,x})_{k\in\N}$ with $x \in \inputSpace^{k}_{i_{k,x}}$ for all $k\in\N$ such that 
\[\bandwidth^{k_n,n}_{i_{k_n,x}} = p(x)n(1 + \convergenceInProbability{}{1}).
\]
By construction, it is $\bandwidth^{k_n,n}_{i_{k_n,x}} = \LOBfunctionOfLPS{\GPR}{n}(x) (1 + \convergenceInProbability{}{1})$.
It follows $\LOBfunctionOfLPS{\GPR}{n}(x) = p(x)n(1 + \convergenceInProbability{}{1})$ such that $\abs{\LOBfunctionOfLPS{\GPR}{n}(x)} = \left[p(x)n\right]^{\frac{\inputDimension}{2\alpha+\inputDimension}}(1 + \convergenceInProbability{}{1})$.
Therefore, asymptotically, $\fctnComplexityLPS{\GPR}{n}(x) := \left[\frac{1}{p(x)n}\right]^{\frac{\inputDimension}{2\alpha+\inputDimension}}\abs{\LOBfunctionOfLPS{\GPR}{n}(x)}^{-1}$
does not depend on $n$ and $p$. Under homoscedasticity, asymptotically, $\fctnComplexityLPS{\GPR}{n}$ is necessarily a function that only depends on $f$, which justifies its use as a measure of \LFC.
\end{proof}

\section{Algorithmic summary of the proposed \AL framework}
\label{sec:algo}

\begin{algorithm}[H]
  \caption{$(\Theta_H, \bandwidth_E) \leftarrow \text{hyper\_init}(\trainingInputs{n_0}, \trainingLabels{n_0},p_0, \bm{X}_\text{val}, \bm{Y}_\text{val})$}
  \label{alg:hyperparameterInit}
  \renewcommand\thealgorithm{}
  \color{black}
  \relsize{-1}
\vspace{1.5mm}
  \begin{algorithmic}[1]
  \Algphase{Input}
  \State Initial training data $(\trainingInputs{n_0}, \trainingLabels{n_0})$
  \State Training data density $p_0$
  \State A labeled validation set $(\bm{X}_\text{val}, \bm{Y}_\text{val})$
  \Algphase{Output}
  \State Initial hyperparameters $\Theta_H = (B, \sparsity, \{\sigma_l\}_{l=1}^L, \sigma_G, \lambda_G, \expertInducingPoints{E}, \gateInducingPoints, \gateNoise_0, \eta_{\gateNoise}, \vartheta_\sigma, \eta_0, \eta_H, \eta_G)$
  \State Global (anisotropic) expert bandwidth $\bandwidth_E \in \Theta_T$
  \Algphase{Procedure}
  \LineComment{Initialize secondary hyperparameters related to computational complexity}
  \State Identify $\sparsity \equiv L$
  \State Set $m_E \leftarrow n_0$ and $m_G \leftarrow \frac{n_0}{4}$ \Comment{Recall $m_E = \abs{\expertInducingPoints{E}}$, $m_G = \abs{\gateInducingPoints}$ are the number of \inducingPoints}
  \State Draw \inducingPoint locations $\expertInducingPoints{E}, \gateInducingPoints \sim p_0$ as described in \appref{sec:diverseIPs}
  \LineComment{Tune expert-related hyperparameters}
  \State Choose ($B, \eta_0, \eta_H$) as described in \secref{subsubsec:hyperpars} according to the validation performance of $\SVGP{\theta}$, where $\mu, \lambda, \gpNoiseVariance, \bandwidth_E, \expertInducingValueMean{} \in \theta$ are learned with respect to $(\trainingInputs{n_0}, \trainingLabels{n_0})$ and ($B, \eta_0, \eta_H$)
  \State Set $\bandwidth_E \in \Theta_T$, where we choose $\bandwidth_E \in \theta$ from the best performing $\SVGP{\theta}$ of the previous step
  \LineComment{Initialize secondary hyperparameters related to fine-tuning}
  \State Set $L \leftarrow 7$ and $\sigma_l \leftarrow 2^\frac{l-4}{\intrinsicDimension}$ for $1\leq l \leq L$ as described in \secref{subsubsec:hyperpars}
  \State Set $\gateNoise_0 \leftarrow 0.1, \eta_{\gateNoise} \leftarrow 1/\sqrt{2}$ and $\vartheta_\sigma \leftarrow 0.01$ as described in \secref{subsubsec:hyperpars}
  \LineComment{Tune \MoE related hyperparameters}
  \State Choose ($\sigma_G, \lambda_G$, $\eta_G$) as described in \secref{subsubsec:hyperpars} according to the validation performance of $\fMoE$ from \eqref{eq:fMoE}, where the model parameters
  $\Theta_T \setminus \{\bandwidth_E\}$ from \eqref{eq:MoEpars}
  are learned with respect to $(\trainingInputs{n_0}, \trainingLabels{n_0})$ and the model hyperparameters
  $\Theta_H \setminus \{\sigma_G, \lambda_G, \eta_G\}$
  from \eqref{eq:hyperPars} are fixed
  \State Choose ($\vartheta_\sigma$, $\gateNoise_0$, $L$, $\{\sigma_l\}_{l=1}^L$) as described in \secref{subsubsec:hyperpars} according to the validation performance of $\fMoE$, where the model parameters
  $\Theta_T \setminus \{\bandwidth_E\}$ from \eqref{eq:MoEpars}
  are learned with respect to $(\trainingInputs{n_0}, \trainingLabels{n_0})$ and the model hyperparameters
  $\Theta_H \setminus \{\vartheta_\sigma, \gateNoise_0, \{\sigma_l\}_{l=1}^L\}$
  from \eqref{eq:hyperPars} are fixed
  \State Decrease $\sparsity \in \Theta_H$ (beginning from $\sparsity = L$) as long as the validation performance of $\fMoE$ does not degrade
  \end{algorithmic}
\end{algorithm}

\section{Finding diverse \inducingPoint locations}
\label{sec:diverseIPs}
In order to obtain diverse \inducingPoint locations with a certain distribution, we consider two approaches, \emph{Stein variational gradient descent} (SVGD) \citep{liu2016NIPSstein,han2018stein} and \emph{distributional clustering} (DC) \citep{krishna2019distributional}.

\myParagraph{Stein Variational Gradient Descent}
SVGD takes a particle swarm and tries to align the empirical distribution of the particles with a target distribution, of which we require the density, as well as its derivative \citep{liu2016NIPSstein}. In addition, the individual particles repel each other, such that we have both diversity and representativeness.
In our scenario we have no access to this derivative, such that we resort to the work of \cite{han2018stein} that is solely based on the density.
Since the particles move freely in the input space and we have to evaluate the target density a considerable number of times, we suggest applying SVGD, when we deal with well-behaved input spaces and target densities that are easy to evaluate.
If the input space is only given through high-dimensional features from a finite set of samples, SVGD might move particles into regions far apart from the data manifold.

\myParagraph{Distributional Clustering}
DC is similar to the known \emph{\kMeans clustering} \citep{gan2020data} but solves a different \emph{inertia} objective, that is modified such that asymptotically, as the number of cluster centers $\abs{\expertInducingPoints{}}\rightarrow\infty$, the distribution of the training data is preserved \citep{krishna2019distributional}. Under the standard \kMeans clustering objective, we would observe $\expertInducingPoints{} \sim p^\frac{\inputDimension}{2+\inputDimension}$ \citep{graf2007foundations}, where it was $\trainingInputs{n} \sim p$.
Since we intend to use clustering for sub-sampling rather than identifying a fixed number of true cluster centers, we deal with a comparably large number of cluster centers, here. Thus, we will use DC so as to obtain a representative set of \inducingPoints.
Due to very mild assumptions on the problem, DC is specifically easy to perform in higher dimensions.

\myParagraph{Dealing with local optima of DC}
The inertia objective of DC is given by
\begin{align}
\label{eq:dcInertia}
\textstyle\text{inertia}_\text{DC}[\bm{c}|\bm{X}_n] = \mySum{c\in\bm{c}}{}\mySum{x\in I_c}{} \indicator_{x \neq c} \log\pnorm{x-c}{},
\end{align}
where
\begin{align}
\label{eq:dcClusterAssignment}
\textstyle I_c = \condset{x\in\bm{X}_n}{\pnorm{x-c}{} \leq \pnorm{x-c'}{}, \forall c'\in\bm{c}}
\end{align}
are those elements in $\bm{X}_n$ that are closest to the center $c$.

In the classical Lloyd-step the centers are updated so as to minimize the intra-cluster inertia, which is given in the case of DC by
\begin{align}
 \label{eq:dcClusterUpdate}
 \textstyle c^* = \argmin_{z\in I_c} \mySum{x\in I_c}{} \indicator_{x \neq z} \log\pnorm{x-z}{}.
\end{align}

It is a known problem that \kMeans-related inertia objectives suffer from local optima \citep{arthur2006k}: The converged solution of cluster centers will typically lie close to their initialization. One way to tackle this issue in practice is to run multiple repetitions of the procedure, followed by choosing the solution with minimal inertia.
Unfortunately, the amount of local optima increases with the number of cluster centers.
In our case, where we use DC for sub-sampling rather than clustering in its usual sense, we deal with a large number of clusters such that this strategy becomes computationally tedious.

Complementary to running multiple repetitions of \kMeans, we will extend the \SOTA method \emph{\kMeansPlusPlus} for choosing the initial set of clusters in a more sophisticated way, where we additionally account for the training distribution.
Given the inertia objective
\[
\textstyle\text{inertia}[\bm{c}|\bm{X}_n] = \mySum{i=1}{n} \min_{c\in\bm{c}} \pnorm{x_i-c}{}^2
\]
of the cluster centers $\bm{c}$, the \emph{\kMeansPlusPlus} procedure builds the set of initial cluster centers as follows:
Draw the first center $c_1$ randomly from $\bm{X}_n$. Then keep track of the current closest squared distance
\begin{align}
 \label{eq:distancesKMeansPlusPlus}
 d_{i}^{m} = \min_{j\in\{1,\ldots,m\}}\pnorm{x_i-c_j}{}^2
\end{align}
of each element $x_i\in\bm{X}_n$ to the so far drawn centers $c_1,\ldots,c_m$ and sample the next center $c_{m+1}$ with probability $\propto (d_{i}^{m})_{i=1}^{n}$ from $\bm{X}_n$. This procedure is repeated until the desired number of cluster centers is reached.

The advantage of \kMeansPlusPlus is that the initial centers are more diverse than if they were sampled at random from $\bm{X}_n$.
However, in its standard form, the centers initialized by \kMeansPlusPlus are themselves distributed flatter than $\bm{X}_n$. And so, in the case of DC, we propose the following adjustment for a \emph{distributional \kMeansPlusPlus}:

We sample with probability $\propto \left(d_{i}^{m}p(x_i)^{2/d}\right)_{i=1}^{n}$ from $\bm{X}_n$, where $\bm{X}_n \sim p$.

\myParagraph{Symmetrized DC for molecules}
Since any symmetric molecule has multiple equivalent representations, care must be taken when measuring distances in DC. 
The key idea is to always compare the two configurations in its closest representation. Using the notation from \appref{subsec:sGDML}, let
\begin{align*}
 \textstyle d(z,z') = \min_{1\leq s\leq\bm{s}} \pnorm{\Phi(z)-\Phi(\pi_s z')}{}
\end{align*}
be the symmetrized distance between two molecule representations.
The symmetrized DC algorithm is then obtained by replacing all occurrences of $\pnorm{z-z'}{}$ with $d(z,z')$ in the cluster assignments $I_c$, the objective $\text{inertia}_\text{DC}[\bm{c}|\bm{X}_n]$, the cluster updates $c^*$ and closest distances $d_{i}^{m}$ from Equations~\ref{eq:dcClusterAssignment},~\ref{eq:dcInertia},~\ref{eq:dcClusterUpdate}~and~\ref{eq:distancesKMeansPlusPlus}.

\section{Design choices of the sparse \MoE model}
\label{sec:designChoices}
In \secref{subsec:modelArchitecture} we have made several design choices with computational feasibility in mind. We will discuss these summarized in this section.

\myParagraph{The gate model}
While in \secref{subsec:modelArchitecture} we have chosen the gate $g_l \sim \SVGP{\theta_{g_l}}$ to be a \GPR model, note that any choice of model with sufficient flexibility would have been possible. \GPR features universal approximation properties, which makes it a favorable choice.

Furthermore, the gate should come with a small degree of freedom to prevent compared to the experts to prevent those from overfitting during the training of the \MoE. For this reason, and the fact that we have no ground truth labels for the training of the gate anyhow, we choose our \GPR-based gate to be sparse.

Finally, note that we share the set of gate \inducingPoint locations \gateInducingPoints across all gate channels. While this is not necessary, it simplifies our method without costs as the \MoE is rather insensitive concerning the gate \inducingPoint locations, as long as these are well-spread.

\myParagraph{The expert models}
While we made clear why we use \GPR experts in our work, we left open in \secref{subsec:modelArchitecture}, whether these experts should be sparse or dense. Here, the deciding factor is the amount of $n$ training samples that we have to deal with: When $n$ goes beyond a few thousand, we suggest switching to sparse \GPR experts for computational reasons.
Note that after training of the \MoE, it is also possible to switch back to full \GPR experts, if one aims for a high accuracy predictor. For the purpose of \AL, this is not necessary.

Similar to the gate, we share the \inducingPoint locations \expertInducingPoints{E} across all experts, which simplifies our model. In contrast to the gate situation, the \MoE is sensitive to the choice of expert \inducingPoint locations. Now, if we were to allow individual \inducingPoint locations for each expert, an elsewise locally underperforming expert might work better than the remaining experts due to a lucky choice of its individual \inducingPoints. Subsequently, this would result in a sub-optimal gate and, hence, ultimately in a wrong \superiorTrainingDensity estimate.

For better generalization, if our \MoE model comprises dense \GPR experts, we will either have to rely on individual training sets for the experts and the gate, or we use \emph{leave-one-out} expert responses on a shared training set.

In the sparse expert case, it is necessary to learn reasonable inducing values $\expertInducingValueMean{}$ prior to the actual learning procedure of the \MoE to not get stuck in a spurious solution.
Therefore, there should be a short pre-training phase for each individual expert.

In addition---whether or not the experts are sparse---the shared expert parameters $\mu_E, \lambda_E, \gpNoiseVariance, \bandwidth_E$ should be initialized reasonably.
In this regard, we suggest training a single, global expert model before the (pre-)training of the actual experts to obtain those initial parameter estimates for which we have no prior knowledge. If we assume isotropic bandwidths to be sufficient, we can simply set $\bandwidth_E = \sigma_E\idMatrix{\inputDimension}$ and learn the scalar $\sigma_E>0$ instead.
Note that, from practice, the training of the \MoE suffers tremendously from online changes of the expert bandwidths. Thus, we suggest to keep $\bandwidth_E$ fixed after initialization.

Finally, note that, if we stick with sparse experts after training of the \MoE, it can be beneficial for the prediction accuracy to re-train the \MoE, where we keep the gate fixed.
In this post-processing step, we would like to apply larger learning rates on the experts to escape local optima. However, larger learning rates also lead to underperforming intermediate steps, in which an actively trained gate might reject the best fitting expert at random---therefore pushing the gate towards a local optimum. Keeping the pre-trained gate fixed at this point prevents this undesired behavior.

\myParagraph{The \inducingPoints}
Recall that we have set the covariance $\expertInducingValueCov{} = 0$ of the inducing value distribution to zero, whereas it could have also been a diagonal or positive definite matrix. Playing around with this parameter, we have seen no significant improvement that would justify the considerable amount of additional model parameters from a computational point-of-view. 

In our approach we suggest keeping the \inducingPoint locations fixed, which is also for reasons of computational feasibility, but, more importantly, adaptive \inducingPoint locations come along with heavy prediction instabilities during the training.

We found it necessary and sufficient to initialize the \inducingPoint locations by \SOTA methods, as described in \appref{sec:diverseIPs}. 

\myParagraph{The \MoE objective}
For the training of our \MoE in \secref{subsubsec:objective}, we added a penalty on small bandwidth choices.
As described in \citep{lepski1991problem,lepski1997optimal}, the optimal bandwidth choice is the largest one that is capable of modeling the function.
Now that we are able to model a comparably flat function by small bandwidths, as long as we have got enough training support, it can occur that, with no regularization, we choose a too-small bandwidth for such a flat region. A too-small bandwidth choice might cause overfitting. But even worse, in the subsequent \AL loop the flat region is falsely identified as complex, leading to more training queries in this location, which then allow for even smaller bandwidths to model this flat region. We will demonstrate this pathological behavior for the unregularized case on toy-data in \secref{subsec:doppler}.

\myParagraph{The gate noise \gateNoise}
Like already mentioned in \secref{subsubsec:objective}, 
it is possible to tune \gateNoise in the training process:
\begin{myRemark}
\cite{shazeer2017outrageously} proposed to learn the \gateNoise parameter by adding a penalty term to the main objective that penalizes the imbalance of how likely training inputs are assigned to each expert: Let $\pi_b\in[0,1]^{L}$ be the expert assignment probabilities of $x_b$ and define $\pi_{\mathcal{B}} = \mySum{b\in\mathcal{B}}{}\pi_b$. Then they add a penalty $\Var[\pi_{\mathcal{B}}]\big/ [\E\pi_{\mathcal{B}}]^2$ to the objective,
which is the \emph{squared coefficient of variation}---a coefficient that accounts for the non-uniformity of a set of positive variables.
\end{myRemark}
We justify our simple heuristic to shrink \gateNoise in a static way as follows:
Recall from \secref{subsec:modelArchitecture} that \gateNoise prevents premature commitment to a spurious solution.
When treating \gateNoise as a trainable parameter, it does not decay towards zero. Maintaining the noise then prevents the locally best-performing experts from converging by randomly withholding training samples.
For this reason, we find that \gateNoise behaves best when decaying towards zero as the training progresses.

\section{Supplemental results on the Doppler experiment}
\label{sec:dopplerSupplement}

\subsection{The single-scale \GPR model}
\label{subsec:dopplerGlobalGPR}
When training a single-scale \GPR model on the Doppler dataset, the tuned bandwidth parameter will typically take an intermediate value, trying to compromise between more complex and simpler regions. This is reflected in the predictions in \figref{fig:dopplerExperiment_dataset_globalPredictions}, where the single-scale \GPR model suffers from the inhomogeneous structure, underfitting the complex region to the left while simultaneously overfitting the simple region to the right.

\fig{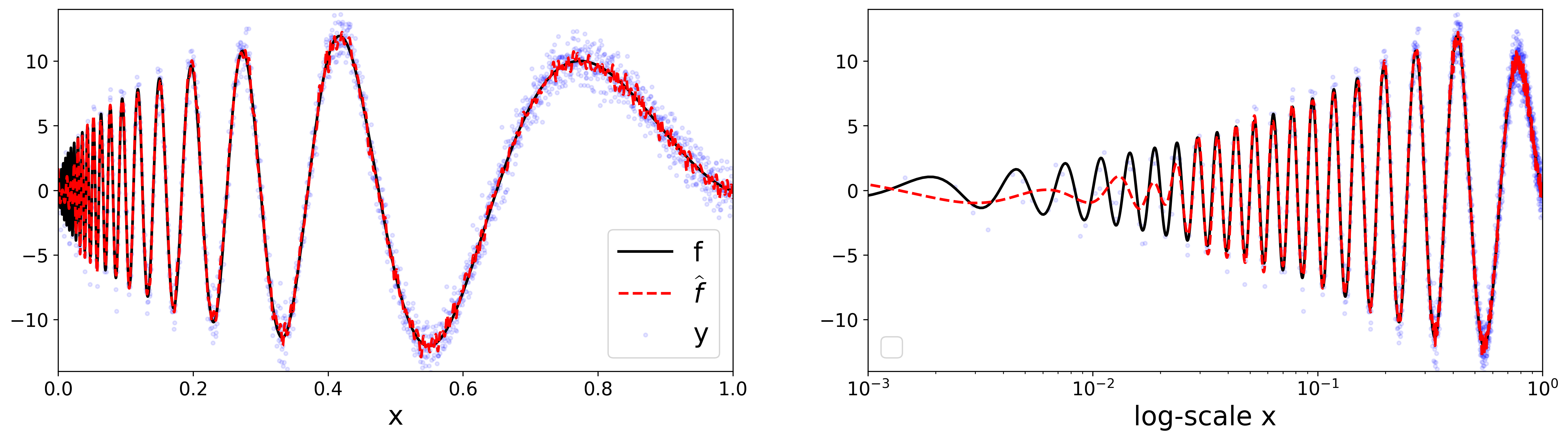}{1}{The Doppler experiment: An exemplary dataset and the predictions of a global \GPR model, shown on natural x-scale (left) and on logarithmic x-scale (right).}{0.8cm}{0.0cm}

In \figref{fig:dopplerExperiment_MoE_VS_globalGPR} we compare the performance of our multi-scale \MoE approach to the single-scale \GPR model. The consistently inferior performance of the single-scale \GPR model shows that the issue above persists even for large training sizes.

\fig{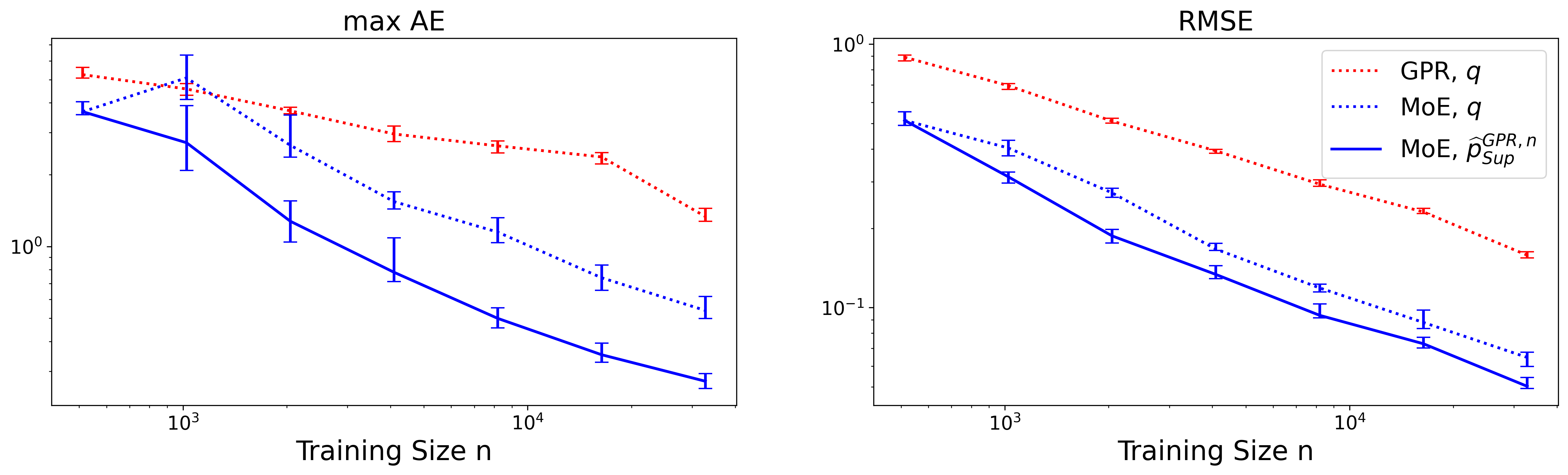}{1}{The Doppler experiment: The \maxAE (left) and the \RMSE (right) of our proposed \MoE model in comparison to a single-scale \GPR model. The results are averaged over $20$ repetitions.}{0.8cm}{0.0cm}

\subsection{Necessity of the small bandwidth penalty}
\label{subsec:dopplerUnregularizedCase}
In \appref{sec:designChoices} we discuss overfitting issues with too small local bandwidth estimates as a consequence of inadequate regularization of \LOB.
To address this issue, we have proposed to penalize such small bandwidth choices by $\vartheta_\sigma \text{pen}_\sigma(\trainingInputs{n}, \trainingLabels{n},\mathcal{B}, w, \thetaMoE)$ with the penalty term $\text{pen}_\sigma$ from \eqref{eq:smallBWpenalty} and a scaling factor $\vartheta_\sigma \geq 0$.

Now, while the \LOB estimate with $\vartheta_\sigma = 0.5$ (see \figref{fig:dopplerExperiment_gate_LFC_LOB_pOpt}) consistently behaves as expected, we show for comparison a typical \LOB estimate in \figref{fig:dopplerExperiment_noBWPen_dataset_LOB} that results from applying no regularization ($\vartheta_\sigma = 0$).
By chance---here, the flat region of the Doppler function to the right---the trained model suffers from massive overfitting by too small \LOB estimates. These falsely obtained small \LOB estimates then lead to overestimation of \LFC, which subsequently results in a detrimental oversampling of these locations by the \AL procedure.

\fig{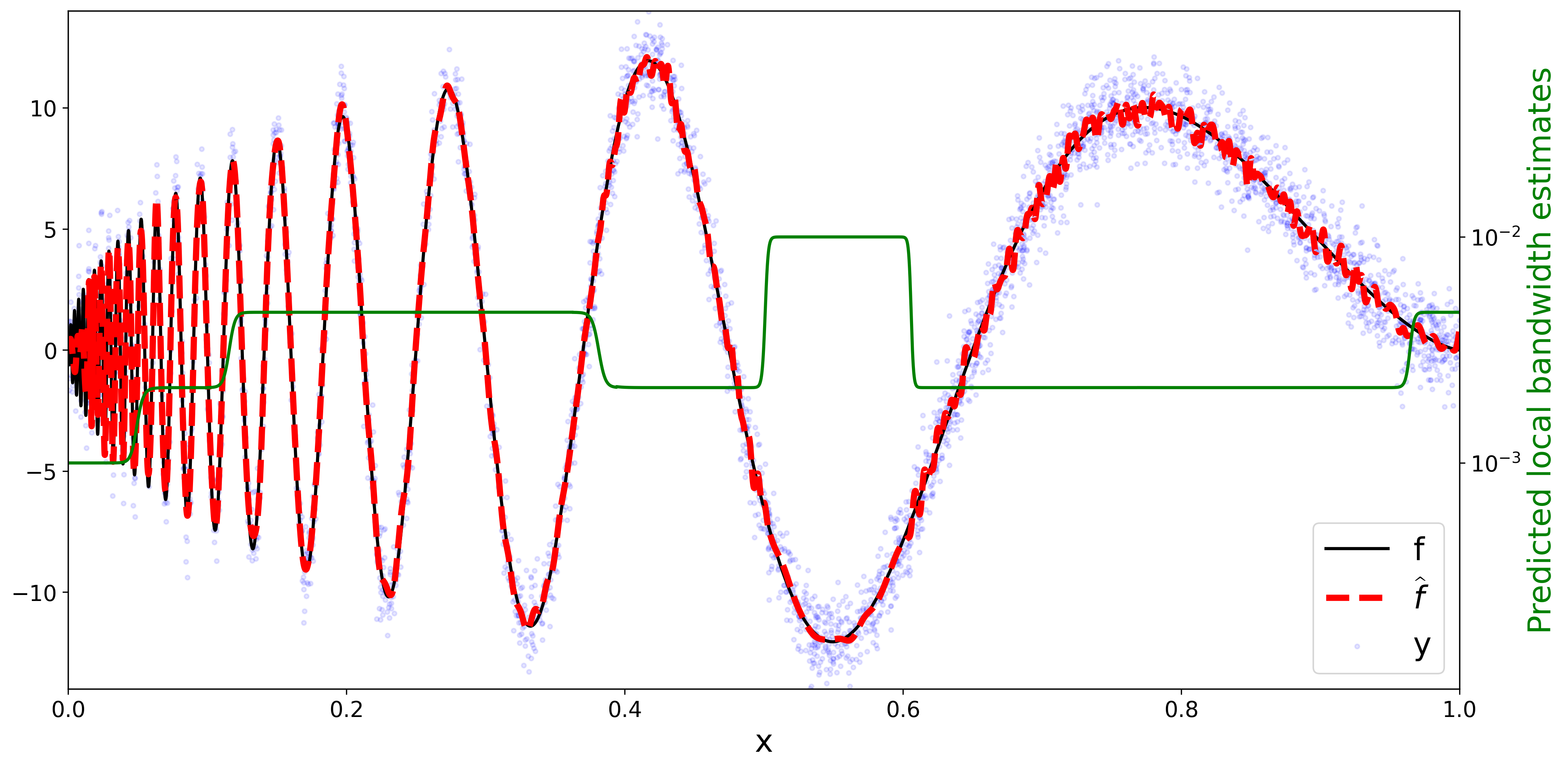}{0.8}{The Doppler experiment: An actively sampled dataset (top) with our \MoE fit at $n = 2^{12}$ training samples without small bandwidth penalty ($\vartheta_\sigma = 0$), and the associated \LOB estimate (bottom).}{0.4cm}{0.0cm}


\section{Supplemental on the malonaldehyde MD simulation experiment}
\label{sec:MDsupplemental}
\subsection{The \sGDML model}
\label{subsec:sGDML}
The GDML model by \cite{chmiela2017} represents the geometry $x = \left[R_1,\ldots, R_{\bm{a}}\right]\in\reals{3\times \bm{a}}$ of each molecule in terms of the reciprocal distances $\Phi(x)_{kl} = \pnorm{R_k - R_l}{}^{-1}$ of all atom-pairings to achieve roto-translational invariance of the input. This representation gives us a total $\bm{d} = \bm{a}(\bm{a}-1)/2$ input features.
The similarity of a pair of configurations $(z,E,F)$ and $(z',E',F')$ is then given by the extended covariance function
\begin{align*}
    &\bm{Cov}(E,E') = k(\Phi(z), \Phi(z')),\\
    &\bm{Cov}(E,F') = \frac{dk(\Phi(z),\Phi(z'))}{d\Phi'}\frac{d\Phi(z')}{dx},\\
    &\bm{Cov}(F,F') = \left[\frac{d\Phi(z)}{dx}\right]^\top\frac{dk(\Phi(z),\Phi(z'))}{d\Phi d\Phi'}\frac{d\Phi(z')}{dx}.
\end{align*}
Hence, we denote the overall kernel function of two configurations by
\begin{align*}
 \textstyle\bm{k}(z, z') = \bm{Cov}((E,F),(E',F')) \in \reals{(3\bm{a}+1) \times (3\bm{a}+1)}.
\end{align*}

Atoms of the same type are physically identical and therefore exchangeable, albeit only a small subset of such symmetries is exercised at a given (low) MD simulation temperature. Full permutational invariance is only needed when enough energy is put into the system for all bonds to break and all atoms to disassociate.

The symmetric extension \sGDML \citep{chmiela2018,chmiela2019} automatically identifies all accessed atom permutations from the training set and adds this symmetric prior to the covariance function.
Formally, let $(\pi_s)_{s=1}^{\bm{s}}$ be atomic permutations that lead to an equivalent molecular representation. Then, the extended symmetric kernel of \sGDML is given by
\begin{align}
 \label{eq:sGDMLKernel}
 \textstyle\widetilde{\bm{k}}(z, z') = \mySum{s=1}{\bm{s}}\mySum{t=1}{\bm{s}} \bm{k}(\pi_s z, \pi_t z').
\end{align}
Malonaldehyde possesses $\bm{s} = 4$ such permutations.

\begin{myRemark}
The identified set of permutations is transitively closed to form a group. Under isotropy, it suffices to permute only one of the two configurations given to the kernel: Permuting both entries (as in \eqref{eq:sGDMLKernel}) equals permuting one entry and multiplying by the constant $\bm{s}$. However, if the applied bandwidth is not of the form $\bandwidth = \sigma\idMatrix{\bm{d}}$, this property does not hold.
\end{myRemark}

\subsection{The malonaldehyde MD simulation experiment under a uniform test distribution}
\label{sec:MDuniformTest}

In this scenario, we assume a uniform test density $q = \uniformDist{\inputSpace}$. Accordingly, we weight the validation and test \MSE by the importance weights $1/\approxPoolDensity{}(\validationInputs)$ and $1/\approxPoolDensity{}(\trueTestInputs)$.
We draw the initial expert training set $\trainingInputs{n}$ of size $n = 2^9$ and the gate training set $\bm{X}_{n_G}^{G}$ via importance sampling from the remaining pool with weights $1/\approxPoolDensity{}(\poolInputs \setminus(\validationInputs\cup \trueTestInputs))$. By this it is $\trainingInputs{n} \sim \uniformDist{\inputSpace}$.

\fig{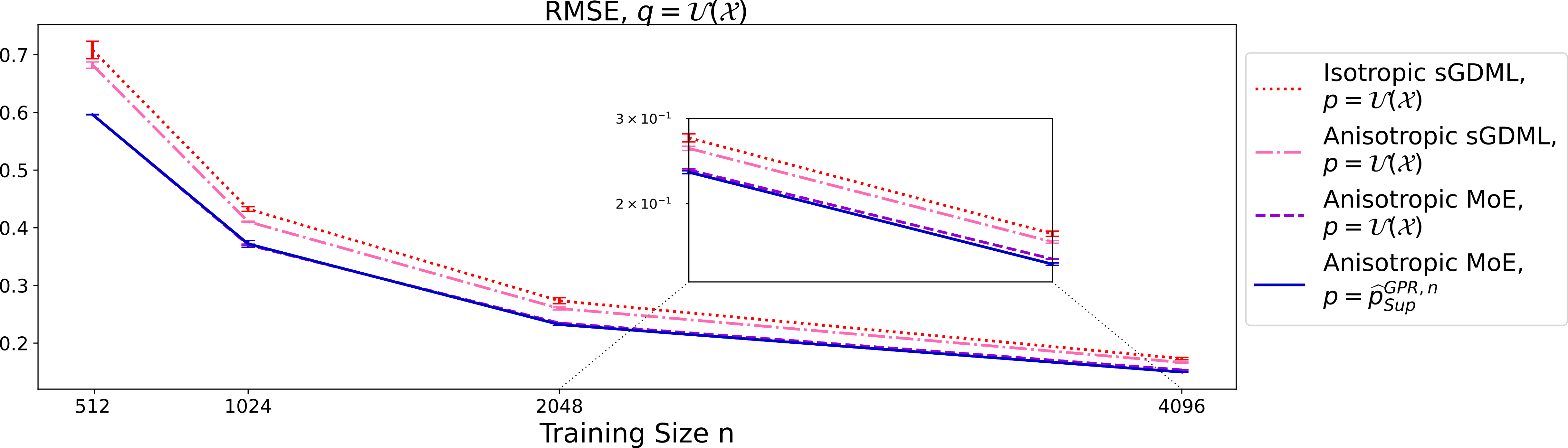}{1}{The \RMSE under the uniform test distribution for different variants of \sGDML and training distribution at varying training size: The performance is given for passive sampling, using the original isotropic \sGDML (dotted), anisotropic \sGDML (dash-dotted) and our \MoE model with anisotropic \sGDML experts (dashed), and for the \MoE model, applying the proposed \superiorSamplingScheme (solid). The results are averaged over $2$ repetitions.}{0.6cm}{0.0cm}

\fig{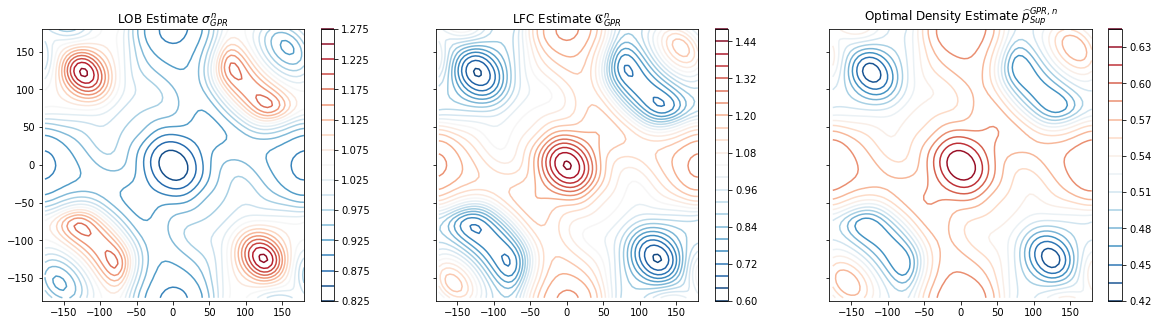}{1}{Estimates of \LOB (left), \LFC (middle) and the \superiorTrainingDensity (right) under the pool test distribution $q = \uniformDist{\inputSpace}$, evaluated at the relaxed malonaldehyde configurations, plotted with respect to the angles of the two aldehyde rotors of malonaldehyde.}{0.6cm}{0.0cm}

In \figref{fig:malonMDestimates_testUnif_lob_lfc_pOpt_contour} we show the estimates of \LOB, \LFC, and the \superiorTrainingDensity under the pool test distribution, evaluated on the relaxed configurations of malonaldehyde.
The \LFC estimates in \figref{fig:malonMDestimates_testUnif_lob_lfc_pOpt_contour} confirm our expectation that the transition areas are more complex to model than the regions near the stable configurations. Subsequently, our active sampling scheme shifts sample mass away from the stable regimes in favor of the transition areas.

We have plotted the error curves of passive and active sampling schemes in \figref{fig:malonMD_IsoVsAniVsMoEVsActivePerformance_pUnif_rmse_withLogBox}.
When estimating the relative sample size \eqref{eq:relSampleSizeGPRApprox} that we require to achieve the same \RMSE via active sampling compared to \randomTestSampling, we obtain $\textstyle\relSampleSize{\fMoE}{\approxPSup{\GPR}{n}} = 0.965\pm0.009$.
This means that we save about 3.5\% of samples with our active sampling scheme.
With similar calculations, we save about 27\%, when comparing the original \sGDML approach with passive sampling to our \MoE model with active sampling.

\subsection{A realistic MD simulation \AL scenario}
\label{sec:MDrealistic}
In the realistic ab initio FF reconstruction \AL scenario, we begin by sampling the initial training set ($\trainingInputs{n_0}, \trainingLabels{n_0}$) as well as the validation set ($\validationInputs, \validationLabels$) by simulating the true MD trajectory solving the computationally expensive Schr{\"o}dinger equation.
Estimate the initial MD density $p_{\inputSpace,0}^{}$ based on ($\trainingInputs{n_0} \cup \validationInputs$).

For $k \in \Nzero:$
\begin{itemize}
 \item Set $q^{k} \leftarrow p_{\inputSpace,k}^{1-\frac{1}{k+1}}$ to encourage exploration in early iterations and exploitation in later iterations
\item Estimate the model $\widehat{f}_k := \fMoE$ based on ($\trainingInputs{n_k}, \trainingLabels{n_k}$)
\item Estimate $\pSup{\GPR}{n_k}$ based on $q^k$ and $\widehat{f}_k$
\item Sample a large pool ($\bm{X}_{pool, k+1}^{}, \widehat{\bm{Y}}_{pool,k+1}^{}$) of, e.g., 100,000 candidates by simulating the approximate MD trajectory using the computationally cheap model $\widehat{f}_k$. While simulation, avoid unreliable out-of-distribution predictions, e.g., by resetting the trajectory in some $x$ whenever $p_{\inputSpace,k}^{}(x) < \epsilon$ drops below a reasonable threshold.
\item Estimate the trajectory density $p_{\inputSpace,k+1}^{}$ of $\bm{X}_{pool, k+1}^{}$
\item Update the training set ($\trainingInputs{n_{k+1}}, \trainingLabels{n_{k+1}}$) by selecting input candidates from the pool $\bm{X}_{pool, k+1}^{}$ with distribution $\pSup{\GPR}{n_k}$ and estimating the respective labels solving the Schr{\"o}dinger equation
\end{itemize}

\end{document}